\documentclass[10pt,twocolumn,letterpaper]{article}

\usepackage{wacv}
 \usepackage{subcaption}
\usepackage{algorithm}
\usepackage{algorithmic}
\usepackage{amsmath}
\usepackage{amssymb}
\usepackage{adjustbox}
\usepackage{cite}
\usepackage[table,xcdraw]{xcolor}

\wacvfinalcopy 

 \def\wacvPaperID{107} 

\ifwacvfinal\fi
\begin{document}

\title{Task Specific Visual Saliency Prediction with Memory Augmented Conditional Generative Adversarial Networks}


\author{Tharindu~Fernando \hspace{1cm} Simon Denman \hspace{1cm} Sridha Sridharan \hspace{1cm} Clinton Fookes\\
\\Image and Video Research Laboratory, Queensland University of Technology (QUT), Australia\\
{\tt\small  \{t.warnakulasuriya, s.denman, s.sridharan, c.fookes\}@qut.edu.au}
}

\maketitle

\begin{abstract}
Visual saliency patterns are the result of a variety of factors aside from the image being parsed, however existing approaches have ignored these. To address this limitation, we propose a novel saliency estimation model which leverages the semantic modelling power of conditional generative adversarial networks together with memory architectures which capture the subject's behavioural patterns and task dependent factors. We make contributions aiming to bridge the gap between bottom-up feature learning capabilities in modern deep learning architectures and traditional top-down hand-crafted features based methods for task specific saliency modelling. The conditional nature of the proposed framework enables us to learn contextual semantics and relationships among different tasks together, instead of learning them separately for each task. Our studies not only shed light on a novel application area for generative adversarial networks, but also emphasise the importance of task specific saliency modelling and demonstrate the plausibility of fully capturing this context via an augmented memory architecture. 
\end{abstract}

\section{Introduction}
Visual saliency patterns are the result of a number of factors, including the task being performed, user preferences and domain knowledge. However existing approaches to predict saliency patterns  \cite{walther2002attentional,hadizadeh2014saliency,wang2015saliency,achanta2009saliency,sharma2012discriminative,yubing2011spatiotemporal} ignore these factors, and instead learn a model specific to a single task while disregarding factors such as user preferences. \par

Based on empirical results, the human visual system is driven by both bottom-up and top-down factors \cite{connor2004visual}. The first category (bottom-up) is entirely driven by the visual scene where humans deploy their attention towards salient, informative regions such as bright colours, unique textures or sudden movements. The bottom-up factors are typically exhibited during the free viewing mechanism. In contrast, the top-down attention component, where the observer is performing a task specific search, is modulated by the task at hand \cite{modellin-search} and the subject's prior knowledge \cite{deep-fix}. For example, when the observer is searching for people in the scene, they can selectively attend to the scene regions which are most likely to contain the targets \cite{ einha2008task, zelinsky2008theory}. Furthermore, a subject's prior knowledge, such as the scene layout, scene categories and statistical regularities will influence the search mechanisms and the fixations \cite{castelhano2007initial, chaumon2008unconscious,neider2006scene}, rendering task specific visual saliency prediction a highly subjective, situational and challenging task \cite{modellin-search, action-in-the-eye,deep-fix}, which motivates the need for a working memory \cite{deep-fix, fernando2017tree,Fernando_2017_ICCV}. Even within groups of subjects completing the same task, due to the differences in a subject's behavioural goals, expectations and domain knowledge, unique saliency patterns are generated. Ideally, this user related context information should be captured via a working memory. \par

\begin{figure}[t]
    \centering
    \begin{subfigure}{.45\linewidth}
       \centering
        \includegraphics[width=.95\linewidth]{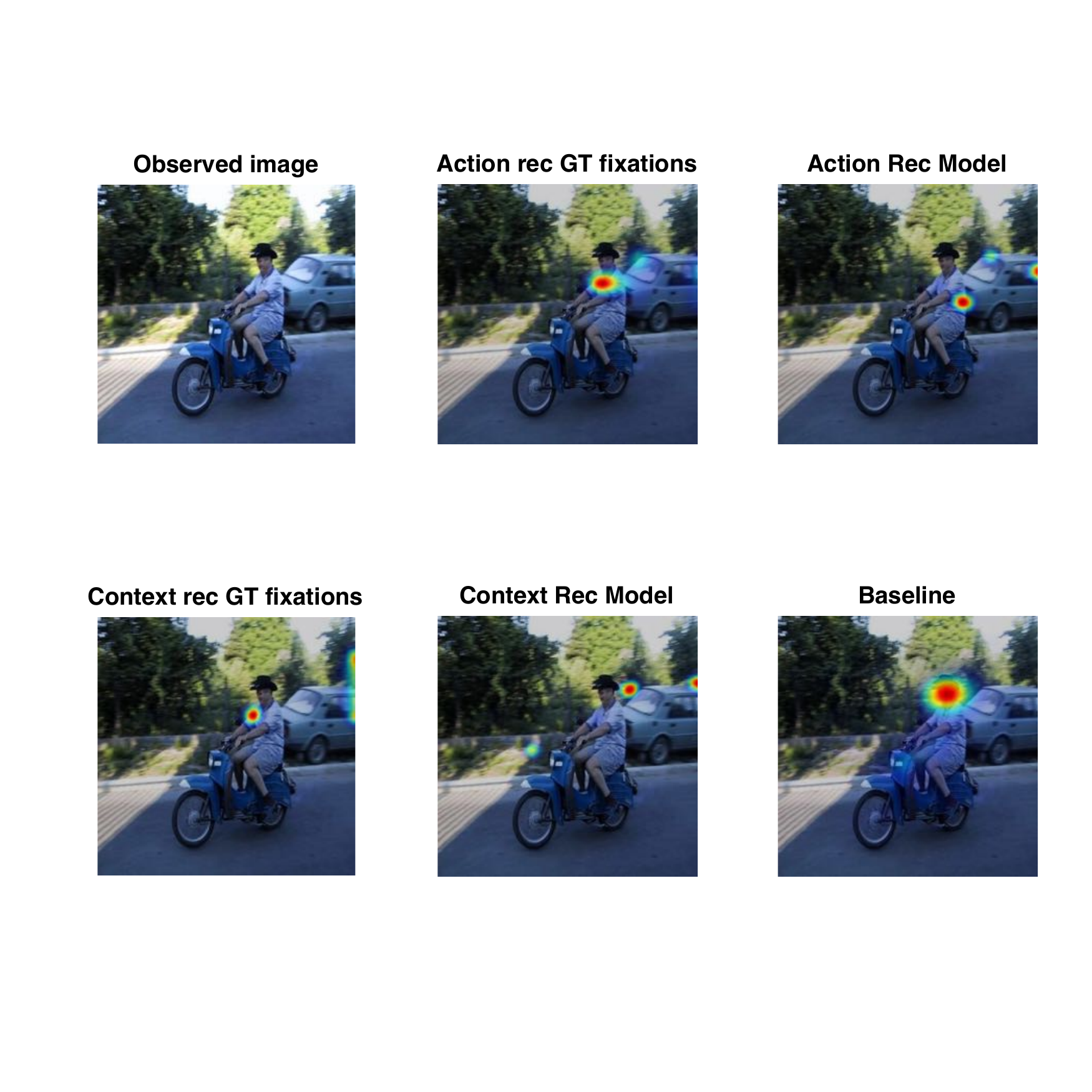} %
        \caption{Human action recognition task}
    \end{subfigure}
     \begin{subfigure}{.45\linewidth}
       \centering
        \includegraphics[width=.95\linewidth]{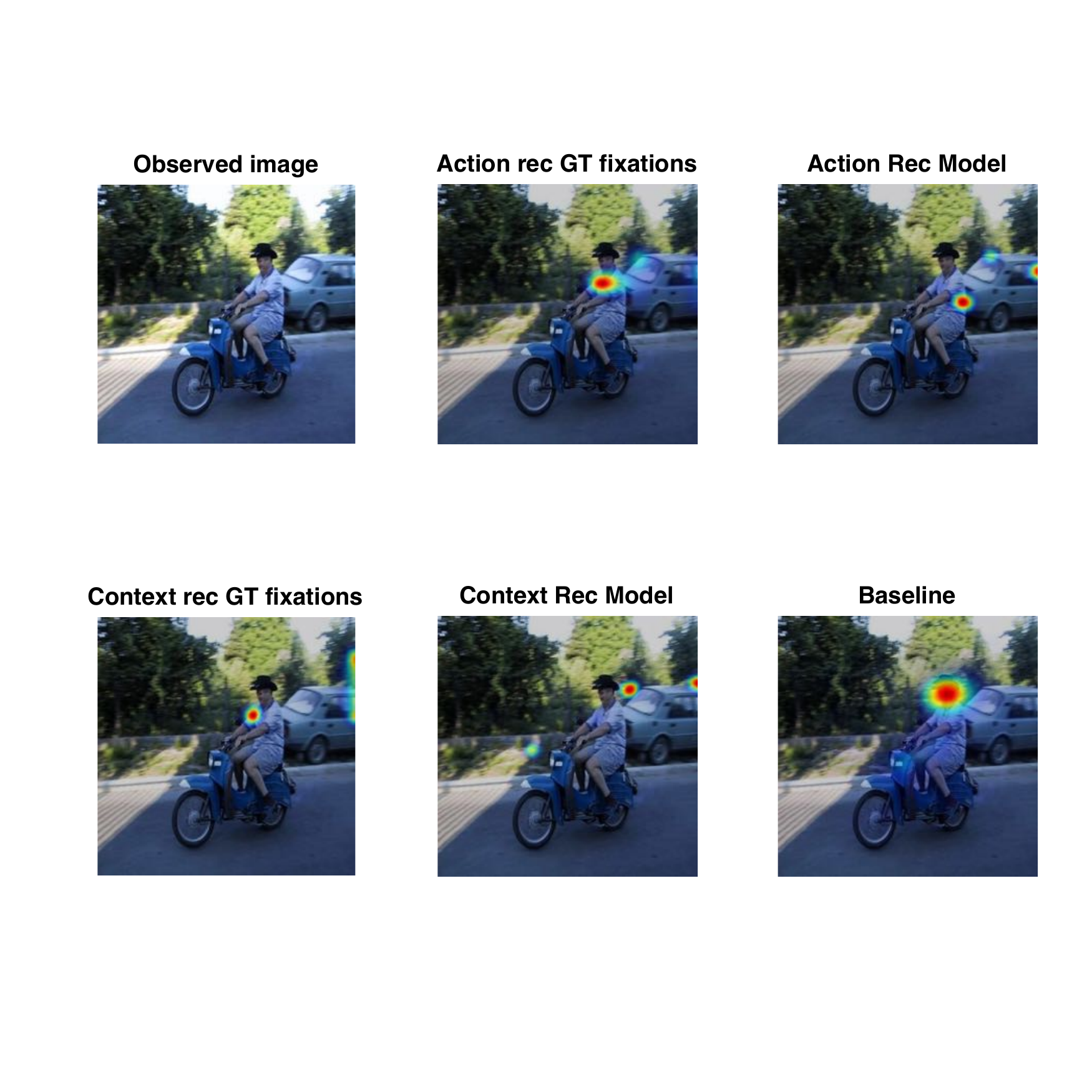} %
        \caption{Searching for cars/ trees}
    \end{subfigure}
     \caption{Variability of the saliency maps when observers are performing different tasks}
			\vspace{-3mm}
        \label{fig:fig_1}
\end{figure}

Fig. \ref{fig:fig_1} shows the variability of the saliency maps when observers are performing action recognition and context recognition on the same image. In Fig. \ref{fig:fig_1} (a) the observer is asked to recognise the action performed by the human in the scene; while in Fig. \ref{fig:fig_1} (b) the saliency map is generated when the observer is searching for cars/ trees in the scene. It is evident that there exists variability in the resultant saliency patterns, yet accurate modelling of human fixations in the application areas specified above requires task specific models. For example, semantic video search, content aware image resizing, video surveillance, and video/ scene classification may require a search for pedestrians, for different objects, or recognising human actions depending on the task and video context. \par

In recent years, motivated by the success of deep learning techniques \cite{gammulle2017two,fernando2017soft+}, there have been several attempts to model visual saliency of the human free viewing mechanism with the aid of deep convolutional networks \cite{deep-fix,deep-ml, vig2014large, kummerer2014deep}. Yet, the usual approach when modelling visual saliency for  task specific viewing is to hand-craft the visual features. For instance, in \cite{modellin-search} the authors utilise the features from person detectors \cite{dalal2006human} when estimating the search for humans in the scene; while in \cite{action-in-the-eye} the authors use the features from HoG descriptors \cite{dalal2005histograms}  when searching for the objects in the scene. Therefore, these approaches are application dependent and fail to capture the top-down attentional mechanism of humans which is driven by factors such as a subject's prior knowledge and expectation. \par 

In this work we propose a deep learning architecture for task specific visual saliency estimation. We draw our inspiration from the recent success of Generative Adversarial Networks (GAN) \cite{goodfellow2014generative, denton2015deep, radford2015unsupervised, salimans2016improved, zhao2016energy} for pixel to pixel translation tasks. We exploit the capability of the conditional GAN framework \cite{pix2pix}  for automatic learning of task specific features in contrast to hand-crafted features that are tailored for specific applications \cite{modellin-search, action-in-the-eye}. This results in a unified, simpler architecture enabling direct application to a variety of tasks. The conditional nature of the proposed architecture enables us to learn one network for all the tasks of interest, instead of learning separate networks for each of the tasks. Apart from the advantage of a simpler learning process, this enables the capability of learning semantic correspondences among different tasks and propagating these contextual relationships from one task to another. \par
\begin{figure}[t]
    \centering
    \begin{subfigure}{.49\linewidth}
       \centering
        \includegraphics[width=.99\linewidth]{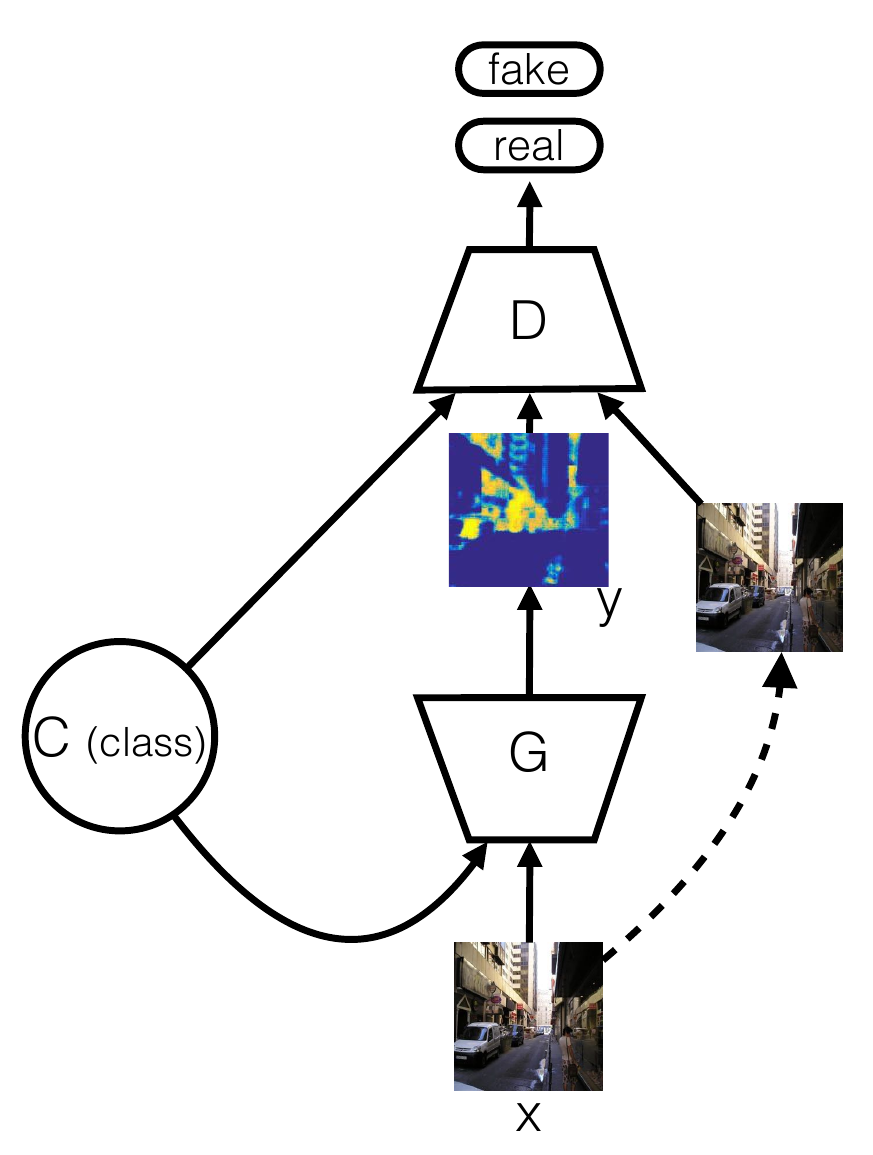} %
        \caption{Conditional GAN}
    \end{subfigure}
     \begin{subfigure}{.49\linewidth}
       \centering
        \includegraphics[width=.99\linewidth]{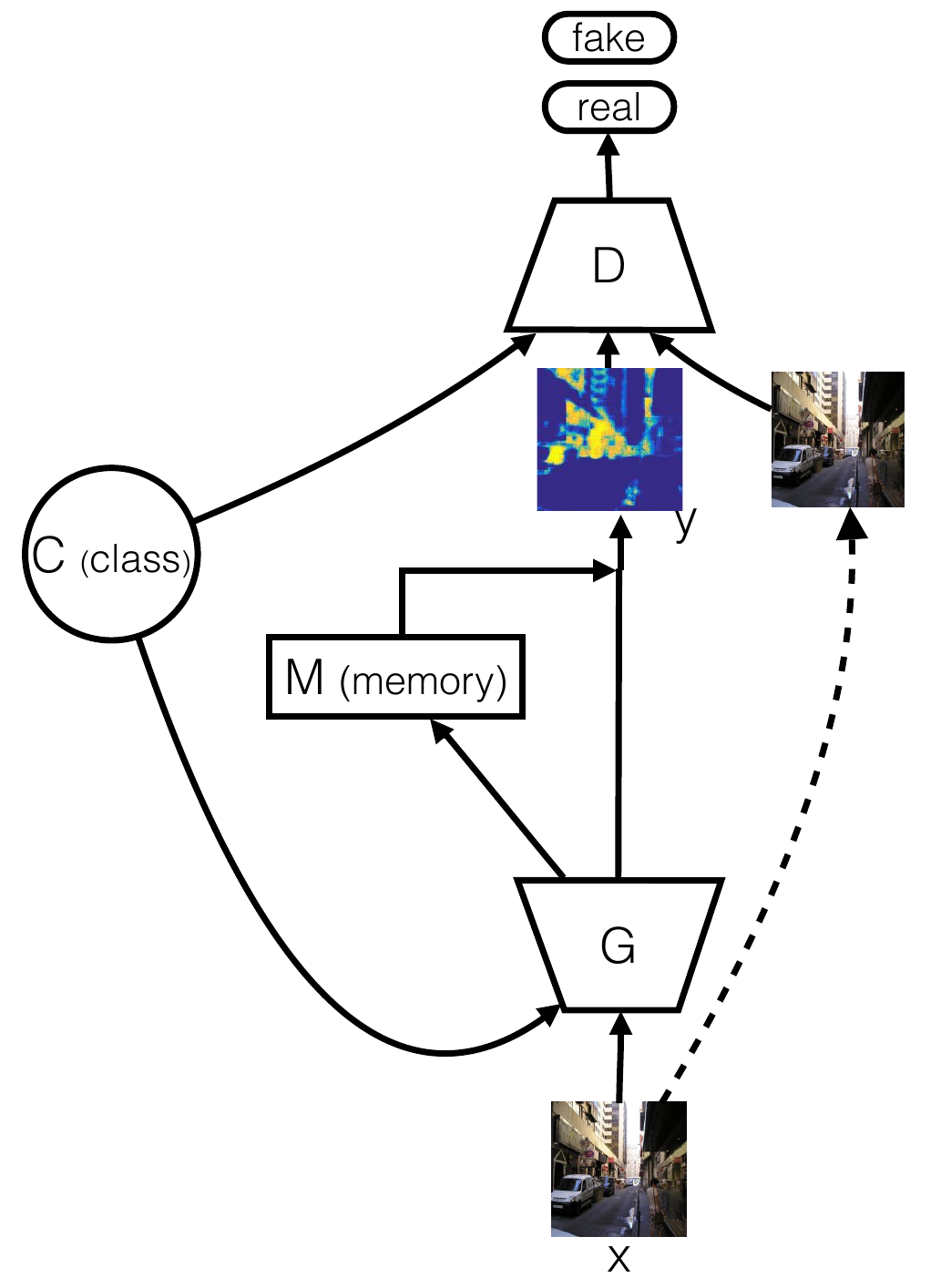} %
        \caption{MC-GAN (proposed model)}
    \end{subfigure}
     \caption{A comparison of conditional GAN architecture with the proposed model}
			\vspace{-3mm}
        \label{fig:networks}
\end{figure}

Fig. \ref{fig:networks} (a) shows the conditional GAN architecture where the discriminator $D$ learns to classify between real and synthesised pairs of saliency maps $y$, given the observed image $x$ and task specific class label $c$. The generator $G$ tries to fool the discriminator. It also observes the observed image $x$ and task specific class label $c$. We compare this model to the proposed model given in Fig. \ref{fig:networks} (b). The differences arise in the utilisation of memory $M$, where we capture subject specific behavioural patterns. This incorporates a subject's prior knowledge, behavioural goals and expectations.

\vspace{-2mm}
\section{Related Work}
Literature related to this work can be broadly categorised into ``Visual Saliency Prediction'' and ``Generative Adversarial Networks'', and these two areas are addressed in Sections 2.1 and 2.2 respectively.

\subsection{Visual Saliency Prediction}
Since the first attempts to model human saliency through feature integration \cite{treisman1980feature}, the area of saliency prediction has been widely explored. Building upon this bottom-up approach, Koch and Ullman \cite{koch1987shifts} and Itti et al. \cite{itti1998model} proposed approaches based on extracting image features such as colour, intensity and orientation. These methods generate centre-biased acceptable saliency predictions for free viewing but are highly inaccurate in complex real world scenes \cite{deep-fix}. Recent studies such as \cite{valenti2009image,liu2013saliency, lang2012depth, erdem2013visual,porikli2006covariance} have looked into the development of more complex features for saliency estimation.  \par
In contrast, motivated by information theory, authors in \cite{bruce2006saliency,modellin-search, action-in-the-eye} have taken the top-down approach where task dependent features comes into play. They incorporate local information from regions of interest for the task at hand, such as features from person detectors and HoG descriptors. These models \cite{bruce2006saliency,modellin-search, action-in-the-eye} are completely task specific, rendering adaptation from one task to another nearly impossible. Furthermore, they neglect the fact that different subjects may exhibit different behavioural patterns when achieving the same goal which generates unique strategies or sub goals that we term user preferences. \par

In order to exploit the representative power of deep architectures, more recent studies have been driven towards the utilisation of convolution neural networks. In contrast to the above approaches, which use hand crafted features, deep learning based approaches offer automatic feature learning. In \cite{kummerer2014deep} the authors propose the usage of feature representations from a pre-trained model that has been trained for object classification. This work was followed by \cite{deep-fix,deep-ml} where authors train end-to-end saliency prediction models from scratch and their experimental evaluations suggest that deep models trained for saliency prediction itself can outperform off-the-shelf CNN models. \par
Liu et al. \cite{liu2015predicting} proposed a mulitresolution-CNN for predicting saliency, which has been trained on multiple scales of the observed image. The motivation behind this approach is to capture low and high level features. Yet this design has an inherit deficiency due to the use of isolated image patches which fail to capture the global context, composed of the context of the observed image, the task at hand (i.e free viewing, recognising actions, searching for objects) and user preferences. Even though the context of the observed image is well represented in deep single scale architectures such as \cite{deep-fix,deep-ml} they ignore the rest of the global context, the task description and user behavioural patterns, which are often crucial for saliency estimation. \par
The proposed work bridges the research gap between deep architectures that capture bottom-up saliency features; and top-down methods \cite{bruce2006saliency,modellin-search, action-in-the-eye} that are purely driven by the hand crafted features. We investigate the plausibility of the complete automatic learning of global context, which has been ill represented in literature thus far,  through a memory augmented conditional generative adversarial model. 

\subsection{Generative Adversarial Networks}
Generative adversarial networks (GAN), which belong to the family of generative models, have achieved promising results for pixel-to-pixel synthesis \cite{arici2016associative}. Several works have looked into numerous architectural augmentations when synthesising natural images. For instance, in \cite{gregor2015draw} the authors utilise a recurrent network approach where as in \cite{dosovitskiy2015learning} a de-convolution network approach is used to generate higher quality images. Most recently authors in \cite{pan2017salgan} have utilised the GAN architecture for visual saliency prediction and proposed a novel loss function which is proven to be effective for both initialising the generator, and stabilising adversarial training. Yet their work fails to delineate the ways of achieving task specific saliency estimation and of incorporating task specific dependencies and the subject behavioural patterns for saliency estimation. \par
The proposed work draws inspiration from conditional GANs \cite{li2016precomputed,mathieu2015deep,mirza2014conditional,pathak2016context,reed2016generative,yoo2016pixel,zhou2016learning,wang2016generative}. This architecture is extensively applied for image based prediction problems such as image prediction from normal maps \cite{wang2016generative}, future frame prediction in videos \cite{mathieu2015deep}, image style transfer \cite{li2016precomputed}, image manipulation guided by user preferences \cite{zhou2016learning}, etc. In \cite{pix2pix} the authors proposed a novel U-Net \cite{ronneberger2015u} based architecture for conditional GANs. Their evaluations suggested that this network is capable of capturing local semantics with applications to a wide range of problems. We investigate the possibility of merging the discriminative learning power of conditional GANs together with a local memory to fully capture the global context, contributing a novel application area and structural argumentation for conditional GANs. 

\vspace{-2mm}
\section{Visual Saliency Model}

\subsection{Objectives}

Generative adversarial networks learn a mapping from a random noise vector $z$ to an output image $y, G: z \rightarrow y$ \cite{goodfellow2014generative}; where as conditional GANs learn a mapping from an observed image $x$ and random noise vector $z$, to output $y$, given auxiliary information $c$, where $c$ can be class labels or data from other modalities. $G: \{x,z | c \} \rightarrow y$. When we incorporate the notion of time into the system, then the observed image at time instance $t$ will be $x_t$, the respective noise vector $z_t$ and the relevant class label will be $c_t$. Then the objective function of a conditional GAN can be written as, 
\begin{equation}
\begin{split}
L_{cGAN}(G,D)  =E_{x_t, y_t \sim  p_{data}(x_t, y_t)}[log(D (x_t, y_t | c_t))] + \\
E_{x_t \sim p_{data(x_t), z_t \sim p_z(z_t)}}[log(1-D(x_t, G(x_t,z_t | c_t)))].
\end{split}
\end{equation}

\begin{figure*}[t]
    \centering
    \includegraphics[width=.90\linewidth]{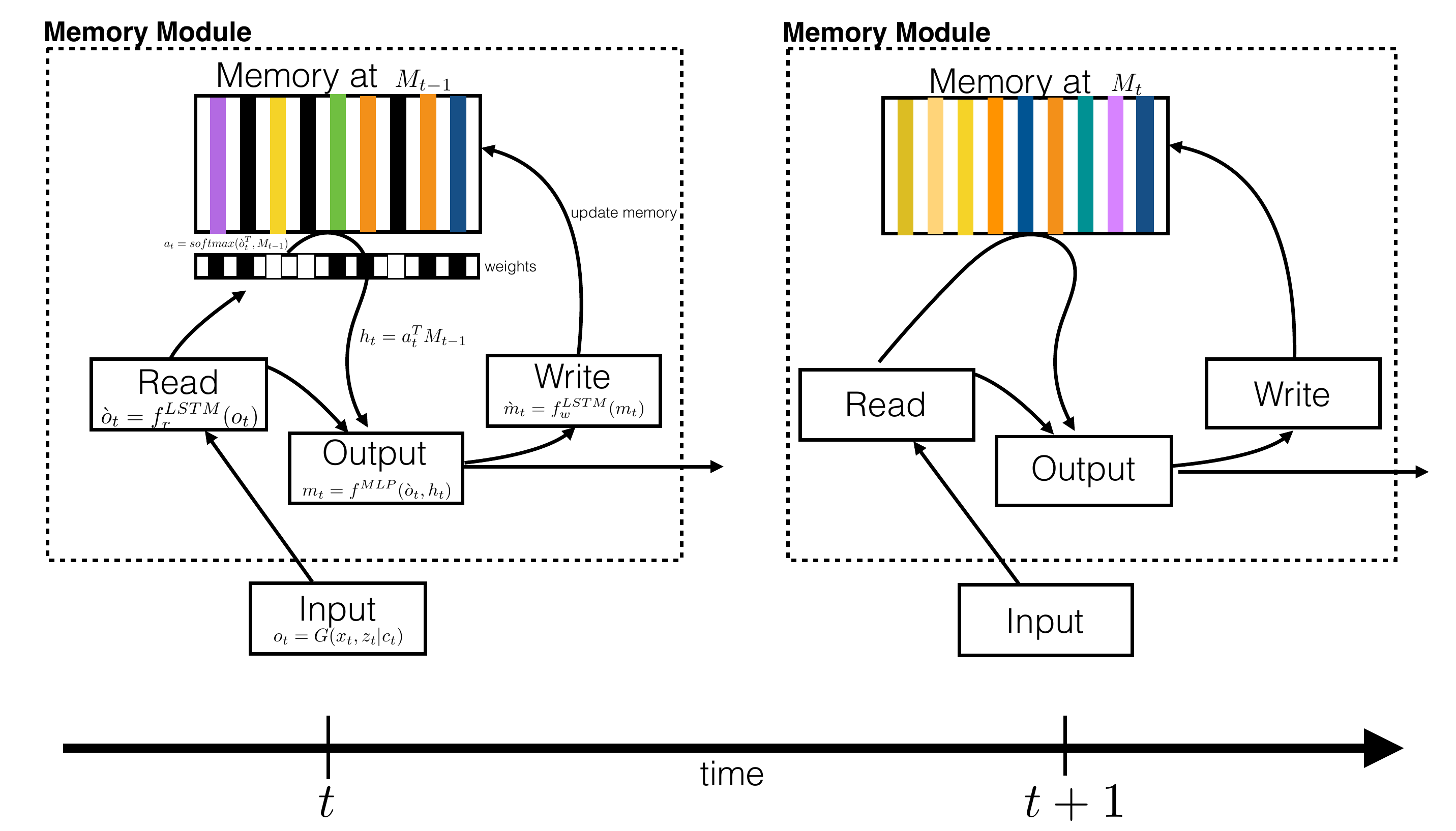} %
        \caption{Memory architecture: The model receives an input $o_t$ at time instance $t$. A function $f_r^{LSTM}$ is used to embed this input and retrieve the content of the memory $M_{t-1}$. When reading, we use a $softmax$ function to weight the association between each memory slot and the input $o_t$, deriving a weighted retrieval $h_{t}$. The final output $m_t$ is derived using both $\grave{o}$ and $h_{t}$. Finally we update the memory using memory write function $f_w^{LSTM}$. This generates the memory representation $M_{t}$ at time instance $t+1$, shown to the right.}
			\vspace{-3mm}
         \label{fig:memory_architecture}
 \end{figure*}
 
Let $ M \in \mathbb{R}^{k*l} $, shown in Fig. \ref{fig:memory_architecture}, be the working memory with $k$ memory slots and $l$ is the embedding dimension of the generator output,
\begin{equation}
o_t= G(x_t,z_t| c_t).
\end{equation}
If the representation of memory at time instance $t-1$ is given by $M_{t-1}$ and $f_r^{LSTM}$ is a read function, then we can generate a key vector $a_t$, representing the similarity between the current memory content and the current generator embedding via attending over the memory slots such that,
\begin{equation}
\grave{o}_t=f_r^{LSTM}(o_t),
\end{equation}
\begin{equation}
a_t= softmax(\grave{o}_t^T,M_{t-1}),
\end{equation}
and
\begin{equation}
h_t= a_t^TM_{t-1}.
\end{equation}
Then we retrieve the current memory state by, 
\begin{equation}
m_t= f^{MLP}(\grave{o}_t,h_t), 
\end{equation}
where  $f^{MLP}$ is a neural network composed of multi-layer perceptrons (MPL) trained jointly with other networks. 
Then we generate the vector for the memory update $\grave{m}_t$ via passing it through a write function $ f_w^{LSTM}$
\begin{equation}
\grave{m}_t= f_w^{LSTM}(m_t),
\end{equation}
and finally we completely update the memory using,
\begin{equation}
M_{t}=M_{t-1}(1- (a_t \otimes e_k )^T) + (\grave{m}_t \otimes e_l)(a_t \otimes e_k )^T.
\end{equation}
where 1 is a matrix of ones, $e_l \in \mathbb{R}^l $ and $ e_k \in \mathbb{R}^k$ be vectors of ones and $ \otimes$ denotes the outer product which duplicates its left vector $l$ or $k$ times to form a matrix.
Now the  objective of the proposed memory augmented cGAN can be written as, 
\begin{equation}
\begin{split}
L^*_{cGAN}(G,D)  =E_{x_t, y_t \sim  p_{data}(x_t, y_t)}[log(D (x_t, y_t | c_t))] + \\
E_{x_t \sim p_{data(x_t), z_t \sim p_z(z_t)}}[ log(1-D(x_t, o_t \otimes tanh(m_t)))].
\end{split}
\end{equation}
We would like to emphasise that we are learning a single network for all the tasks at hand, rendering a simpler but informative framework, which can be directly applied to a variety of tasks without any fine tuning.
\subsection{Network Architecture}

\begin{figure}[t]
    \centering
    \includegraphics[width=.4\linewidth]{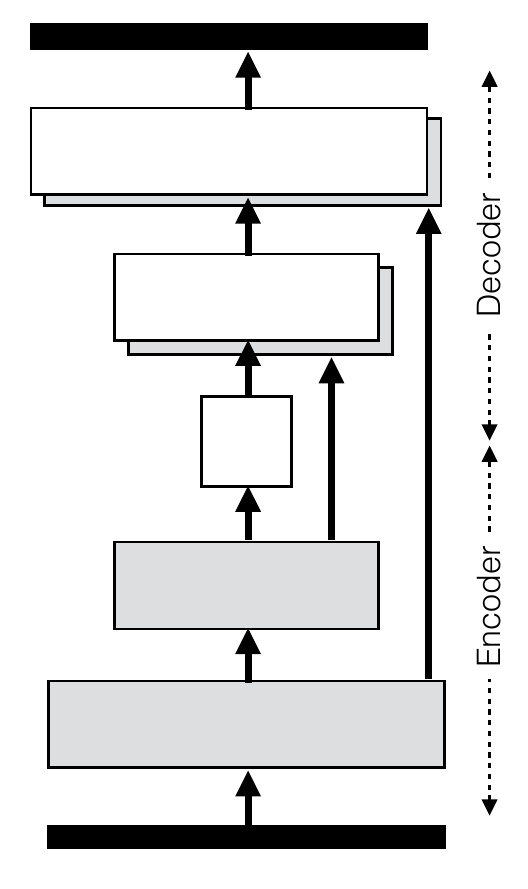} %
        \caption{U-Net architecture}
         \label{fig:U_Net_architecture}
			\vspace{-3mm}
    \end{figure}

For the generator we adapt the U-Net architecture of \cite{pix2pix} (see Fig. \ref{fig:U_Net_architecture}). Let $Ck$ denote a Convolution-BatchNorm-ReLU layer group with $k$ filters. $CDk$ denotes a Convolution-BatchNorm-Dropout-ReLU layer with a dropout rate of 50\%. Then the generator architecture can be written as,
Encoder: C64-C128-C256-C512-C512-C512-C512-C512 followed by a U-Net decoder: CD512-CD1024-CD1024-C1024-C1024-C512-C256-C128 where there are skip connections between each $i^{th}$ layer in the encoder and the ${n-i}^{th}$ layer of the decoder, and there are $n$ total layers in the generator (see \cite{pix2pix} for details). 
The discriminator architecture is: C64-C128-C256-C512-C512-C512. \par
All convolutions are 4 x 4 spatial filters applied with stride 2. Convolutions in the encoder and in the discriminator down sample by a factor of 2, whereas in the decoder they up sample by a factor of 2. A description of our memory architecture is as follows. For functions $f_r^{LSTM}$ and $f_w^{LSTM}$ we utilise two, one-layer LSTM networks \cite{hochreiter1997long} with 100 hidden units and for $f^{MLP}$ we use a neural network with a single hidden layer and 1024 units with ReLU activations. This memory module is fully differentiable, and we learn it jointly with other networks. We trained the proposed model with the Adam \cite{kingma2014adam} optimiser, with a batch size of 32 and an initial learning rate to 1e-5, for 10 epochs.

\vspace{-2mm}
\section{Experimental Evaluations}

\subsection{Datasets}
We evaluate our proposed approach on 2 publicly available datasets, VOCA-2012 \cite{action-in-the-eye} and MIT person detection (MIT-PD) \cite{modellin-search}.  \par
The VOCA-2012 dataset consists of 1,085,381 human eye fixation from 12 volunteers (5 male and 7 female) divided into 2 groups based on the given task. It contains 8 subjects performing action recognition in the given image where as the rest of the subjects are performing context dependent visual search. The subjects in this group are searching for furniture, paintings/ wallpapers, bodies of water, buildings, cars/ trucks, mountain/ hills, road/ trees in the given scene. The MIT-PD dataset consist of 12,768 fixations from 14 subjects (between 18-40 years), where the subjects search for people in 912 urban scenes. MIT-PD contains only a single task, and we use this dataset to demonstrate the effectiveness of the memory network.

\subsection{Evaluation metrics}
Let $N$ denote the number of examples in the testing set, $y$ denotes the ground truth saliency map and $\hat{y}$ is the predicted saliency map. Following this notation we define the following metrics:
\begin{itemize}
\item \textbf{Area Under Curve (AUC): }
AUC is a widely used metrics for evaluating saliency models. We use the formulation of this metric defined in \cite{borji2012exploiting}. 
\item \textbf{Normalised Scan path Saliency (NSS): }
NSS \cite{peters2005components} is calculated by taking the mean scores assigned by the unit normalised saliency map $\hat{y}^{norm}$ at human eye fixations. 
\begin{equation}
NSS=\frac{1}{N}\sum_{i=1}^{N}\hat{y}^{norm}_{i}
\end{equation}
\item \textbf{Linear Correlation Coefficient (CC): }
In order to measure the linear relationship between the ground truth and predicted map we utilise the linear correlation coefficient, 
\begin{equation}
CC=\frac{cov(y,\hat{y})}{\sigma_{y}*\sigma_{\hat{y}}},
\end{equation}
where $cov(y,\hat{y})$ is referred to as the covariance between distributions $y$ and $\hat{y}$, $\sigma_{y}$ is the standard deviation of distributions $y$ and $\sigma_{\hat{y}}$ is the standard deviation of distributions $\hat{y}$. As the name implies $CC=1$ denotes a perfect linear relationship between distributions $y$ and $\hat{y}$ where as a value of $0$ implies that there is no linear relationship. 
\item \textbf{KL divergence (KL): }
To measure the non-symmetric difference between two distributions we utilise the KL divergence measure given by,
\begin{equation}
KL=\sum_{i=1}^{N}\hat{y}_{i}log(\frac{\hat{y}_i}{y_i}).
\end{equation}
As ground truth and predicted saliency maps can be seen as 2D distributions, we can use KL divergence to measure the difference between them.
\item \textbf{Similarity metric (SM): }
Computes the sum of the minimum values at each pixel location between $\hat{y}^{norm}$ and $y^{norm}$ distributions. 
\begin{equation}
SM=\sum_{i=1}^{P}min(\hat{y}^{norm}_{i},y^{norm}_{i}), 
\end{equation}
where
$\sum_{i=1}^{P}\hat{y}^{norm}_{i}=1 $
and
$\sum_{i=1}^{P}y^{norm}_{i}=1 $
are the normalised probability distributions and $P$ denotes all the pixel location in the 2D maps.
\end{itemize}

\subsection{Results}
Quantitative evaluations on the VOCA-2012 dataset is presented in Tab. \ref{tab:tab_1}. In the proposed model, in order to retain the user dependent factors such as user preference in memory, we feed the examples in order such that examples from each specific user go through in sequence. We compare our model with 8 state-of-the-art methods.  The row `human' stands for the human saliency predictor, which computes the saliency map derived from the fixations made by half of the human subjects performing the same task. This predictor is evaluated with respect to the rest of the subjects, as opposed to the entire group \cite{action-in-the-eye}. \par

The evaluations suggest that the bottom-up model of Itti et. al \cite{itti2000saliency} generates poor results as it does not incorporate task specific information.  Even with high level object detectors, the models of Judd et al. \cite{judd2009learning} and HOG detector \cite{action-in-the-eye} fail to render accurate predictions. \par
Deep learning models PDP \cite{jetley2016end} and ML-net \cite{deep-ml} are able to out perform the techniques stated above but they lack the ability to learn task dependent information. We note the accuracy gain of cGAN model over PDP, ML-net and SalGAN, where the model incorporates a conditional variable to discriminate between the `action recognition' and `context recognition' tasks instead of learning two separate networks or fine-tuning on them individually. Our proposed approach builds upon this by incorporating an augmented memory architecture with conditional learning. We learn different user patterns and retain the dependencies among different tasks, and outperform all baselines considered (MC-GAN (proposed), Tab. \ref{tab:tab_1}). \par 
As further study, in Tab. \ref{tab:tab_1}, row M-GAN (separate), we show the evaluations for training two separate memory augmented GAN networks for the tasks in the VOCA-2012 test set without using the conditional learning process. The results emphasise the importance of learning a single network for all the tasks, leveraging semantic relationships between different tasks. The accuracy of the networks learned for separate tasks are lower than the combined MC-GAN and cGAN approaches (rows  MC-GAN (proposed) and cGAN, Tab. \ref{tab:tab_1}), highlighting the importance of learning the different tasks together and allowing the model to discriminate between the tasks and learn the complimentary information, rather than keeping the model completely blind regarding the existence of another task category.\par
\begin{table}[t]
  \centering
  \begin{adjustbox}{width=.98\linewidth,center}
  \begin{tabular}{|c|c|c|c|c|}
 \hline
    & \multicolumn{4}{|c|}{\textbf{Task}} \\
        \cline{2-5}
     & \multicolumn{2}{|c|}{\textbf{Action Rec}}  & \multicolumn{2}{|c|}{\textbf{Context Rec}}\\
           \cline{2-5}
   \textbf{Saliency Models} & \textbf{AUC} & \textbf{KL} & \textbf{AUC} & \textbf{KL} \\
    \hline
     \hline
   HOG detector \cite{action-in-the-eye} & 0.736 & 8.54 & 0.646 & 8.10 \\
    \hline
   Judd et al. \cite{judd2009learning} & 0.715 & 11.00  & 0.636 & 9.66  \\
    \hline
   Itti et. al \cite{itti2000saliency} & 0.533 & 16.53  & 0.512 & 15.04 \\
    \hline
  central bias \cite{action-in-the-eye} & 0.780 & 9.59 & 0.685 & 8.82\\
   \hline
 PDP \cite{jetley2016end} & 0.875 & 8.23 & 0.690 & 7.98 \\
 \hline
  ML-net \cite{deep-ml}& 0.847 & 8.51 & 0.684 & 8.02 \\
   \hline
     SalGAN \cite{pan2017salgan}& 0.848 & 8.47 & 0.679 & 8.00 \\
  \hline
   cGAN \cite{pix2pix} &0.852 &8.24  &0.701 &7.95  \\
    \hline
    M-GAN (separate) & 0.848 & 8.54 & 0.704 & 8.00\\
     \hline
   \textbf{MC-GAN (proposed)} &\textbf{0.901} & \textbf{8.07 } &\textbf{0.734} & \textbf{7.65 } \\
    \hline
     \hline
  Human \cite{action-in-the-eye} & 0.922 & 6.14 & 0.813 & 5.90\\
  \hline
  \end{tabular}
  \end{adjustbox}
  \caption{Experimental evaluation on VOCA-2012 test set. We augment the current state-of-the-art GAN method (SalGAN \cite{pan2017salgan}) by adding a conditional variable (cGAN \cite{pix2pix}) to mimic the joint learning process instead of learning two separate networks. To capture user and task specific behavioural patterns we add a memory module to cGAN, MC-GAN (proposed), and outperform all the baselines. We also compare training $2$ separate memory augmented GAN networks, M-GAN (separate) without the conditional learning process.}
			\vspace{-3mm}
  \label{tab:tab_1}
\end{table}

%

\begin{table}[t]
  \centering
  \begin{adjustbox}{width=.98\linewidth,center}
  \begin{tabular}{|c|c|c|c|c|c|c|}
 \hline
    & \multicolumn{6}{|c|}{\textbf{Task}} \\
        \cline{2-7}
     & \multicolumn{3}{|c|}{\textbf{Action Rec}}  & \multicolumn{3}{|c|}{\textbf{Context Rec}}\\
           \cline{2-7}
   \textbf{Saliency Models} & \textbf{NSS} & \textbf{CC} & \textbf{SM} & \textbf{NSS} & \textbf{CC} & \textbf{SM}  \\
    \hline
     \hline
     ML-net \cite{deep-ml}& 2.05 & 0.71 & 0.51 & 2.03 & 0.64 & 0.42 \\
     \hline
     SalGAN \cite{pan2017salgan}& 2.10 & 0.73 & 0.51 & 2.10 & 0.68 & 0.44 \\
     \hline
   cGAN \cite{pix2pix} & 2.23 & 0.76 & 0.55 & 2.14 & 0.71 & 0.57   \\
    \hline
   \textbf{MC-GAN (proposed)} &\textbf{2.23} & \textbf{0.79 } &\textbf{0.60} & \textbf{2.20 } &\textbf{0.77} & \textbf{0.69 }  \\
    \hline
  \end{tabular}
  \end{adjustbox}
  \caption{Comparison between ML-Net, SalGAN, cGAN and  MC-GAN (proposed) on VOCA-2012}
			\vspace{-3mm}
  \label{tab:tab_3}
\end{table}

\begin{figure*}[!htb]
%
%
%
%
%
%
   
   \centering
    \begin{subfigure}{.160\textwidth}
       \centering
        \includegraphics[width=.95\linewidth]{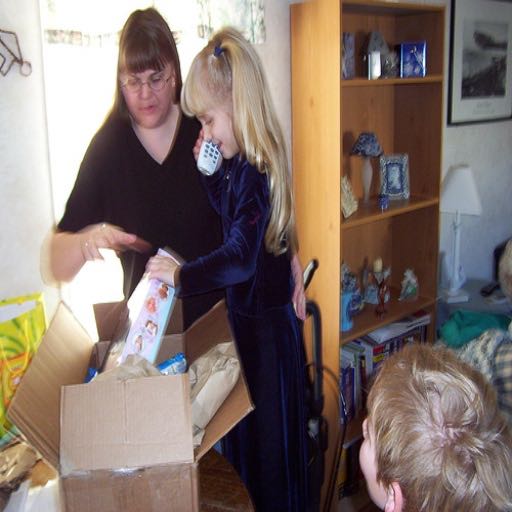} %
        
    \end{subfigure}
     \begin{subfigure}{.160\textwidth}
       \centering
        \includegraphics[width=.95\linewidth]{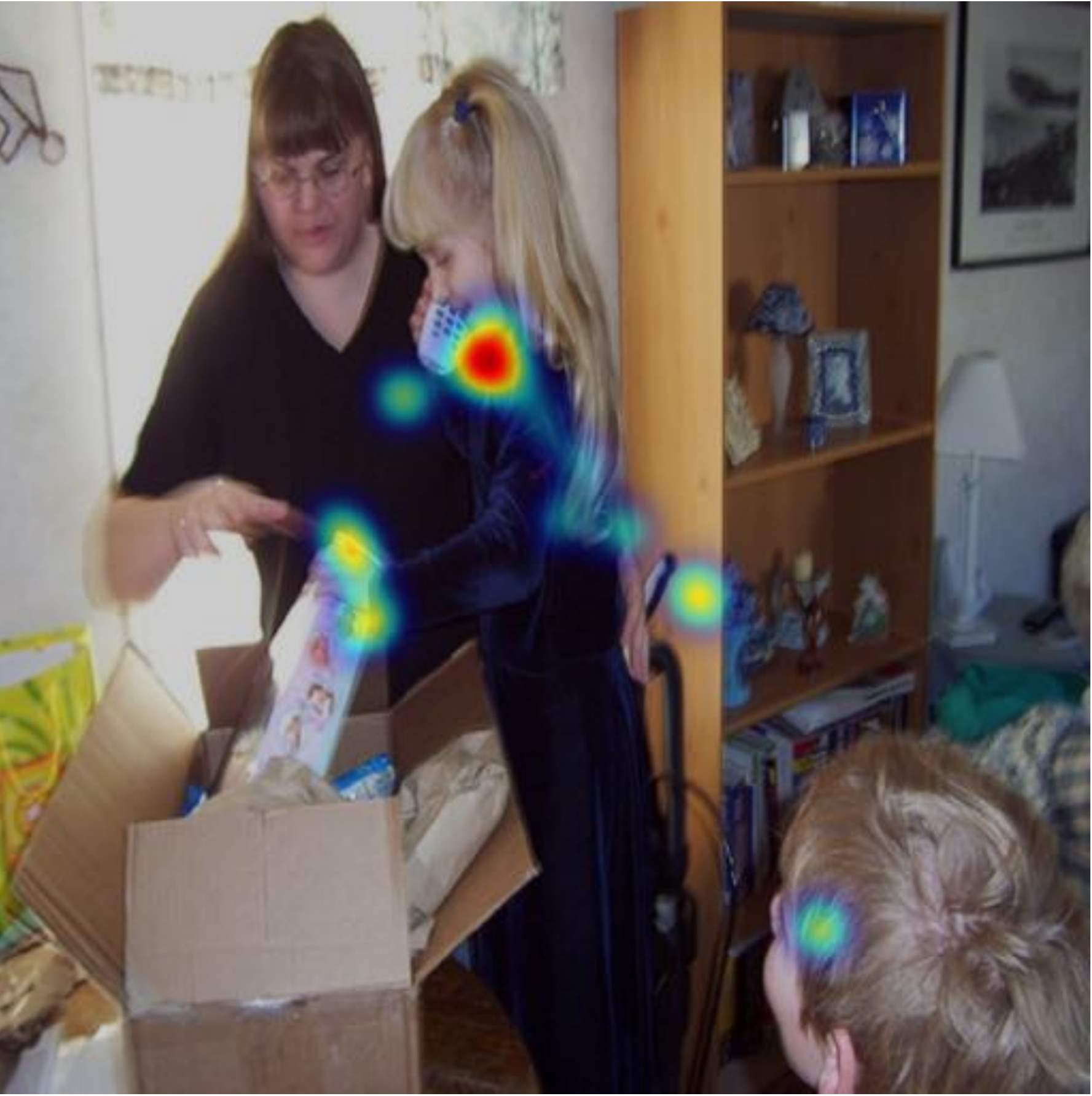} %
       
    \end{subfigure}
     \centering
    \begin{subfigure}{.160\textwidth}
       \centering
        \includegraphics[width=.95\linewidth]{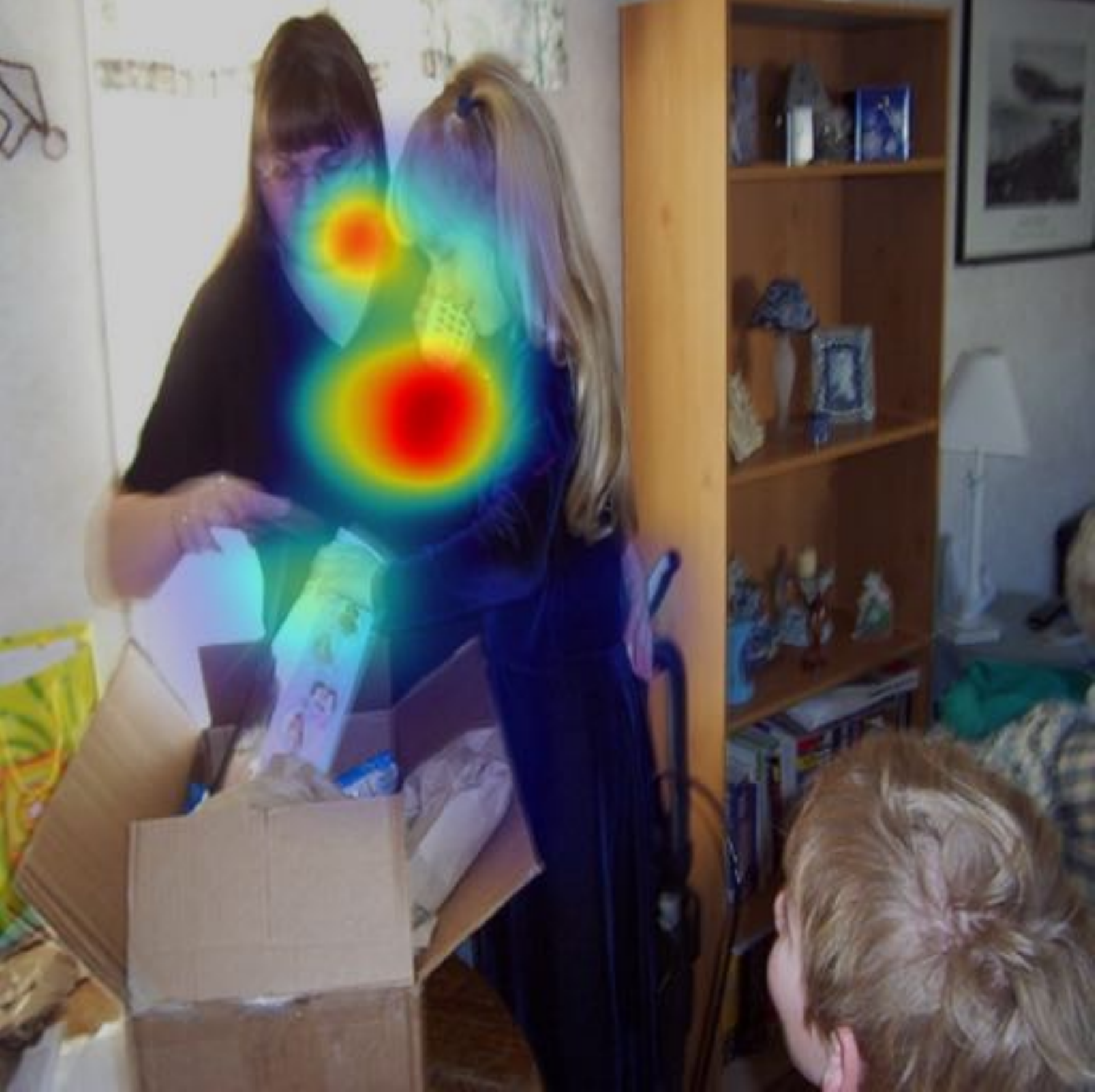} %
        
    \end{subfigure}
 \centering
    \begin{subfigure}{.160\textwidth}
       \centering
       \includegraphics[width=.95\linewidth]{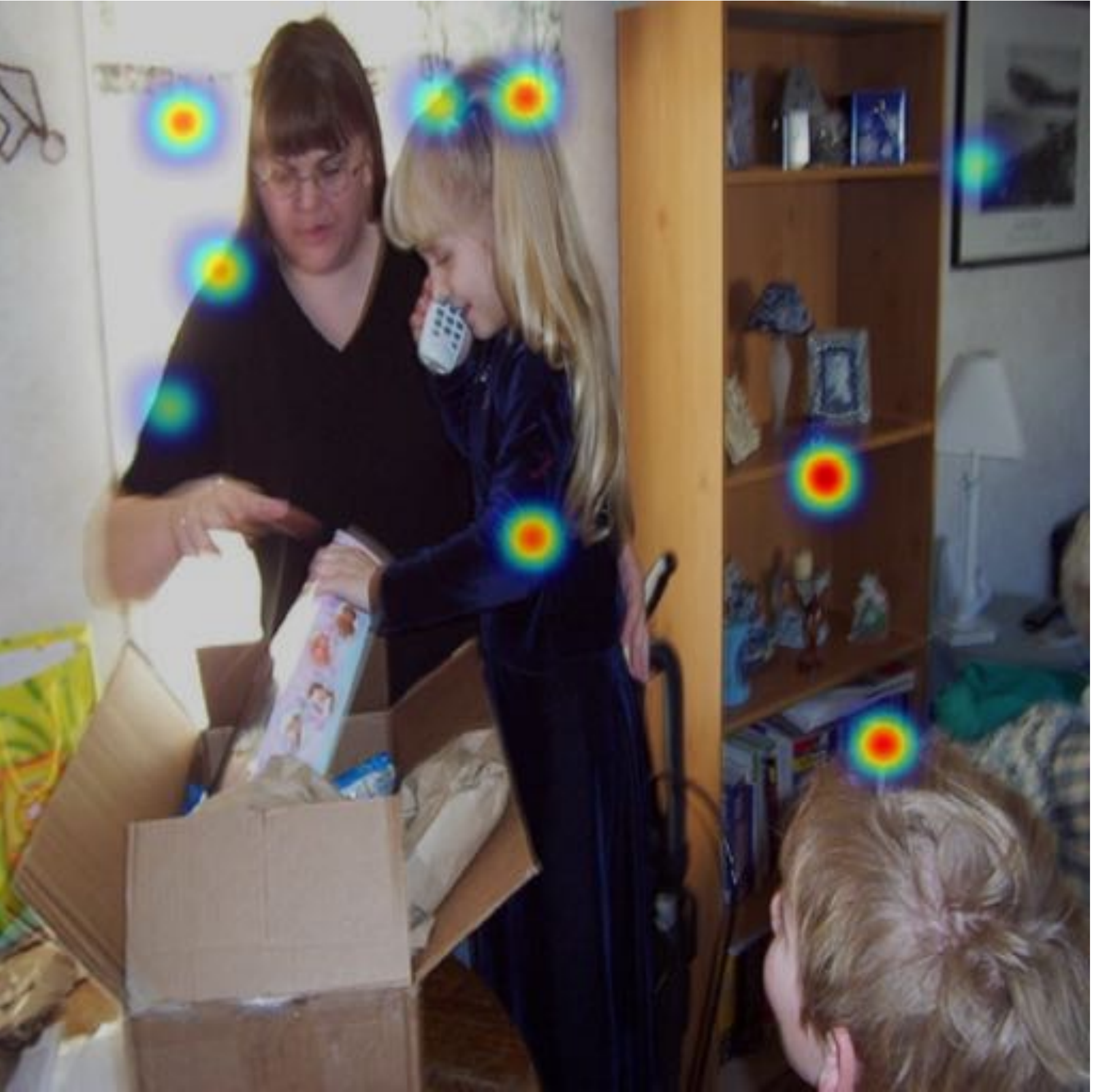} %
    
    \end{subfigure}
 \centering
    \begin{subfigure}{.160\textwidth}
       \centering
         \includegraphics[width=.95\linewidth]{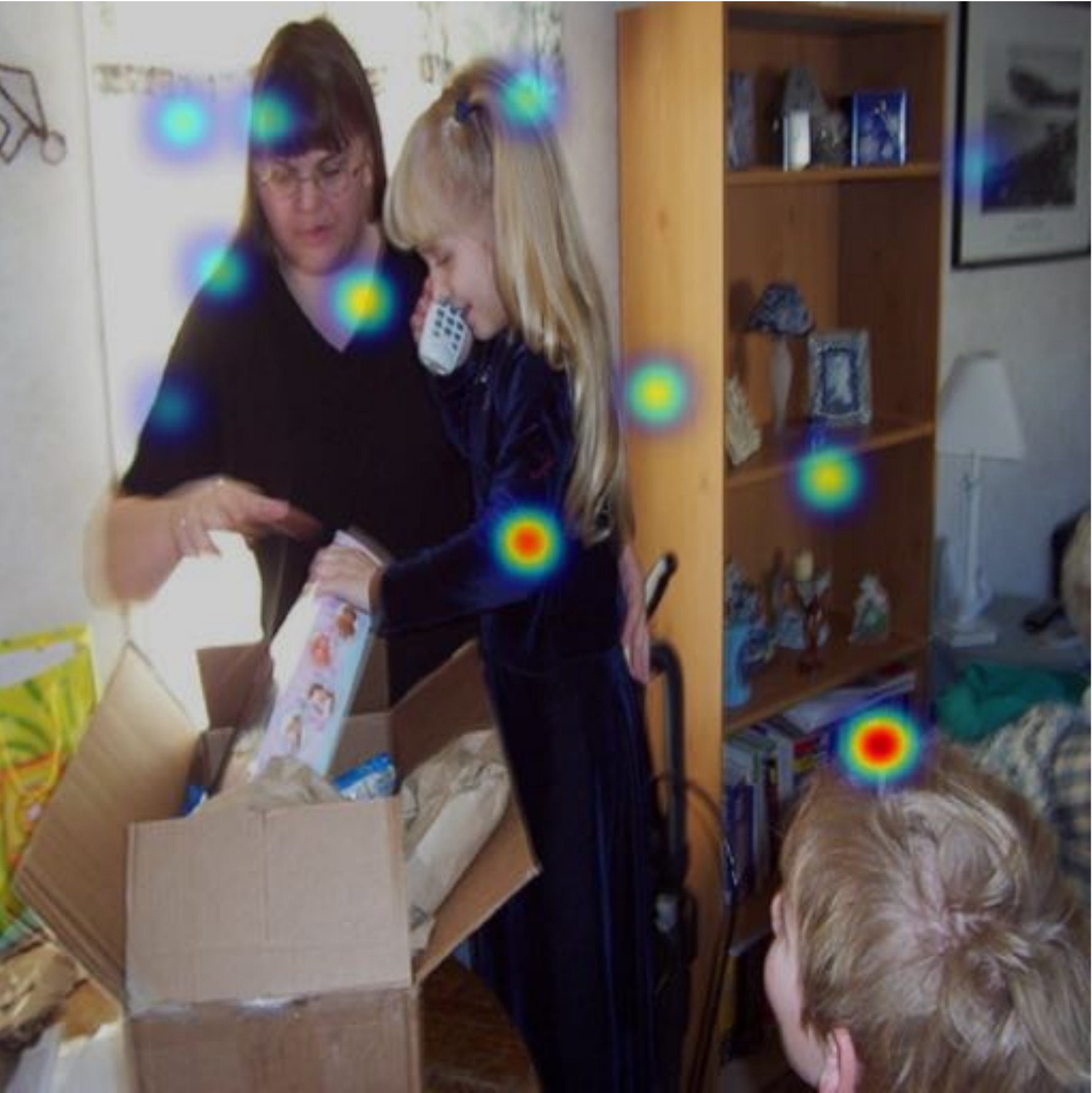} %
        
    \end{subfigure}
 \centering
    \begin{subfigure}{.160\textwidth}
       \centering
        \includegraphics[width=.95\linewidth]{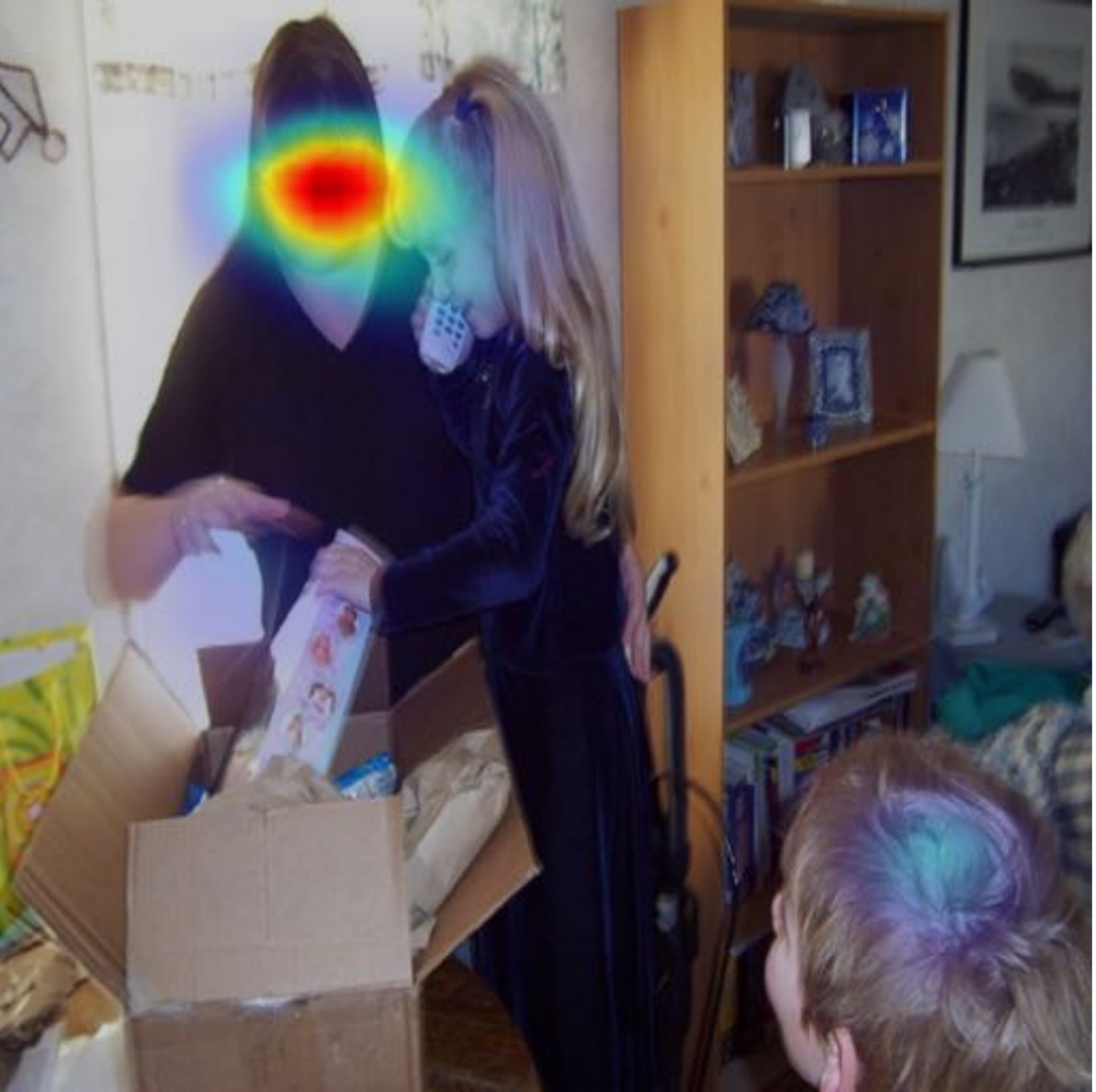} %
        
    \end{subfigure}
    
    \centering
    \begin{subfigure}{.160\textwidth}
       \centering
        \includegraphics[width=.95\linewidth]{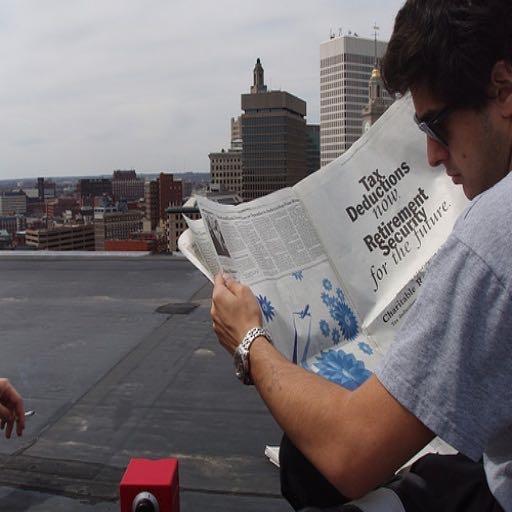} %
        
    \end{subfigure}
     \begin{subfigure}{.160\textwidth}
       \centering
        \includegraphics[width=.95\linewidth]{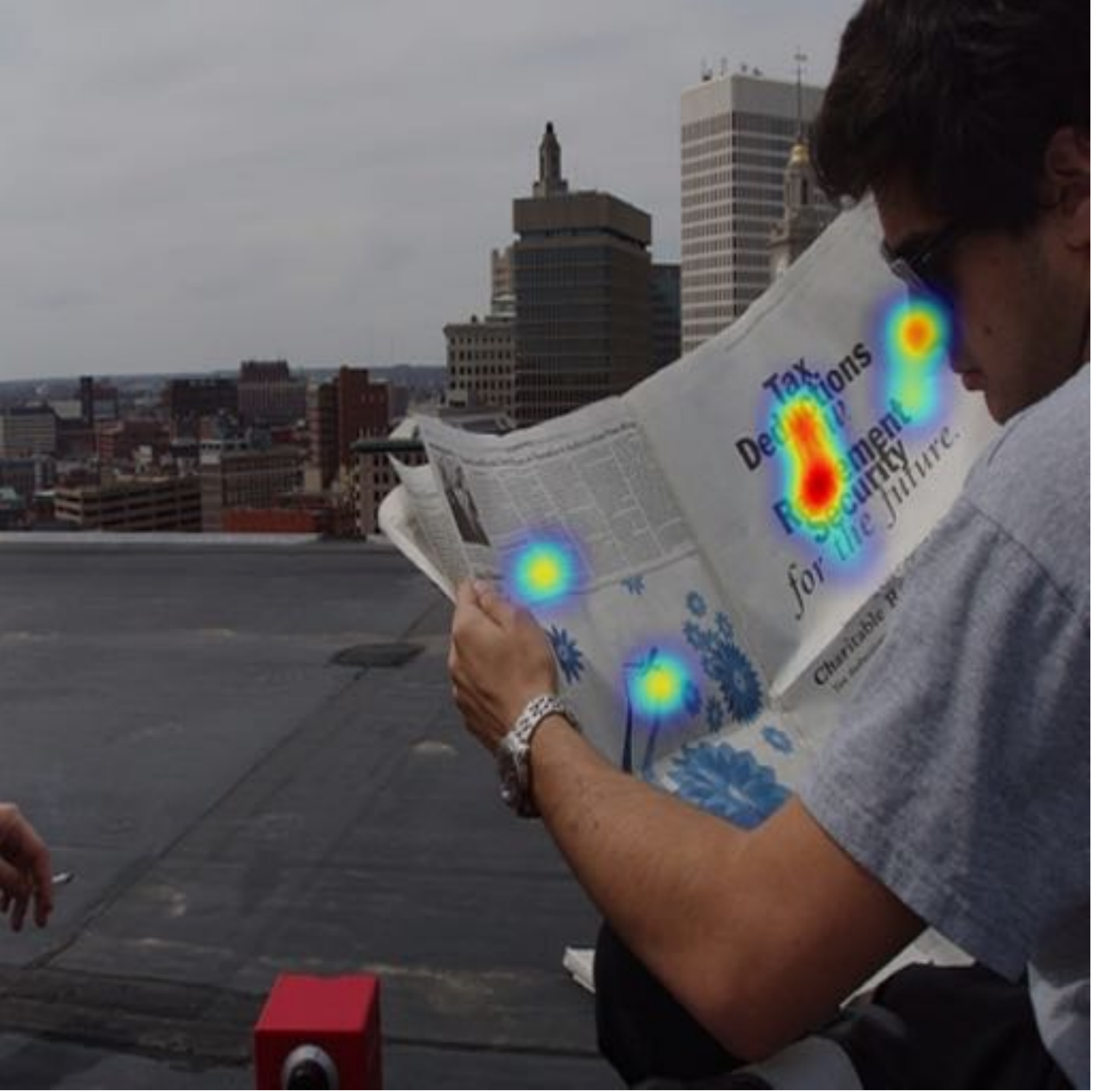} %
       
    \end{subfigure}
     \centering
    \begin{subfigure}{.160\textwidth}
       \centering
        \includegraphics[width=.95\linewidth]{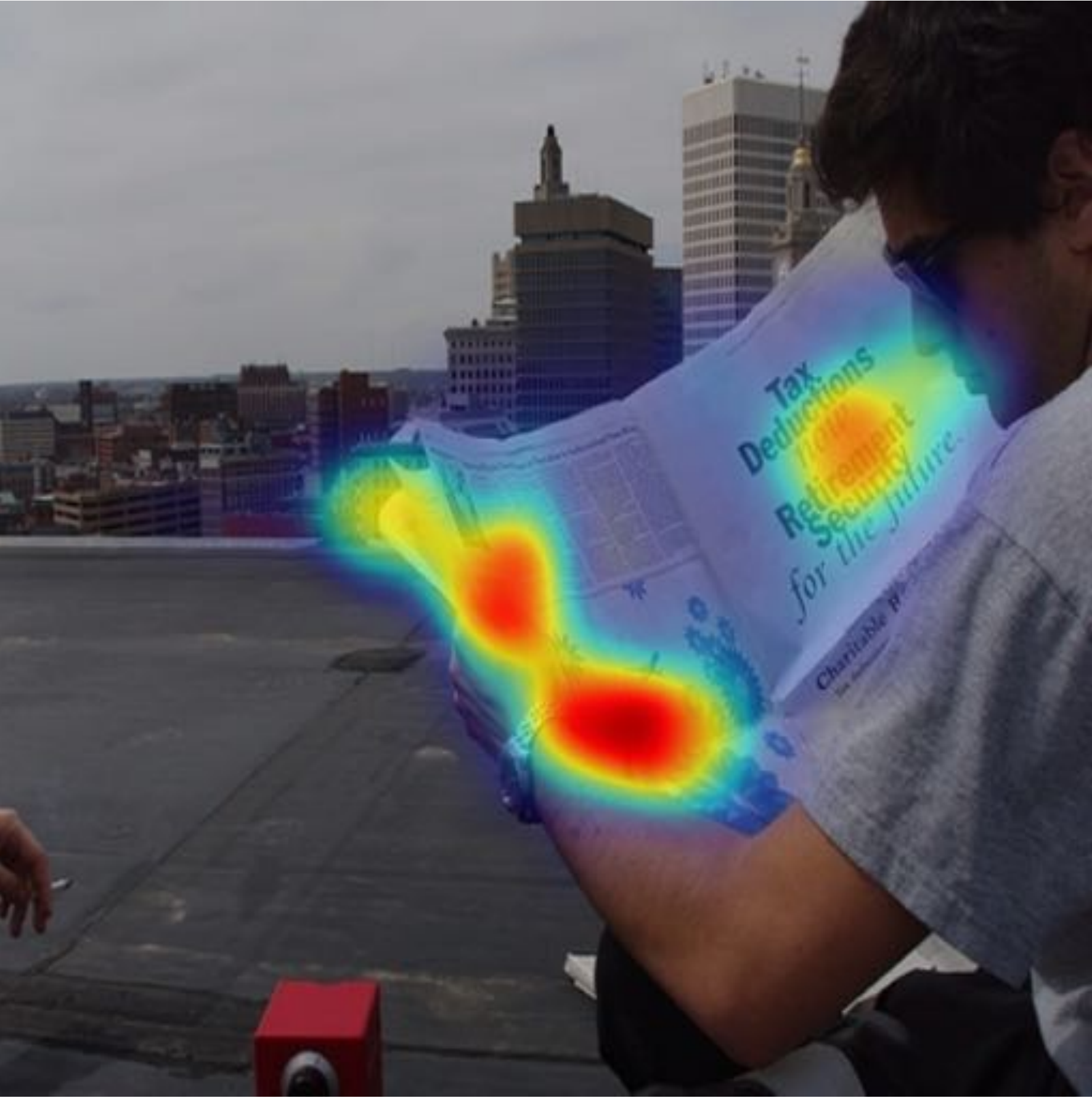} %
        
    \end{subfigure}
 \centering
    \begin{subfigure}{.160\textwidth}
       \centering
       \includegraphics[width=.95\linewidth]{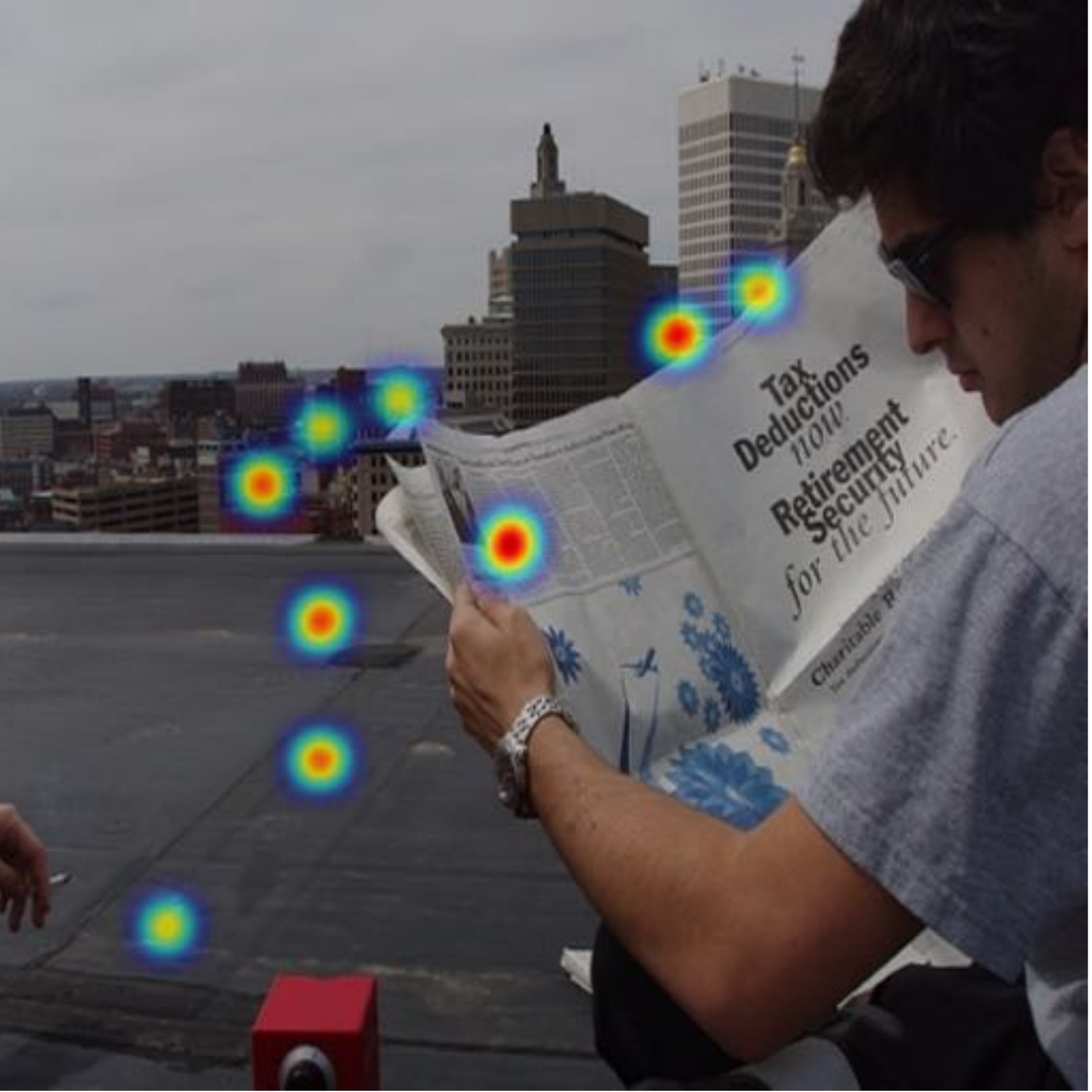} %
    
    \end{subfigure}
 \centering
    \begin{subfigure}{.160\textwidth}
       \centering
         \includegraphics[width=.95\linewidth]{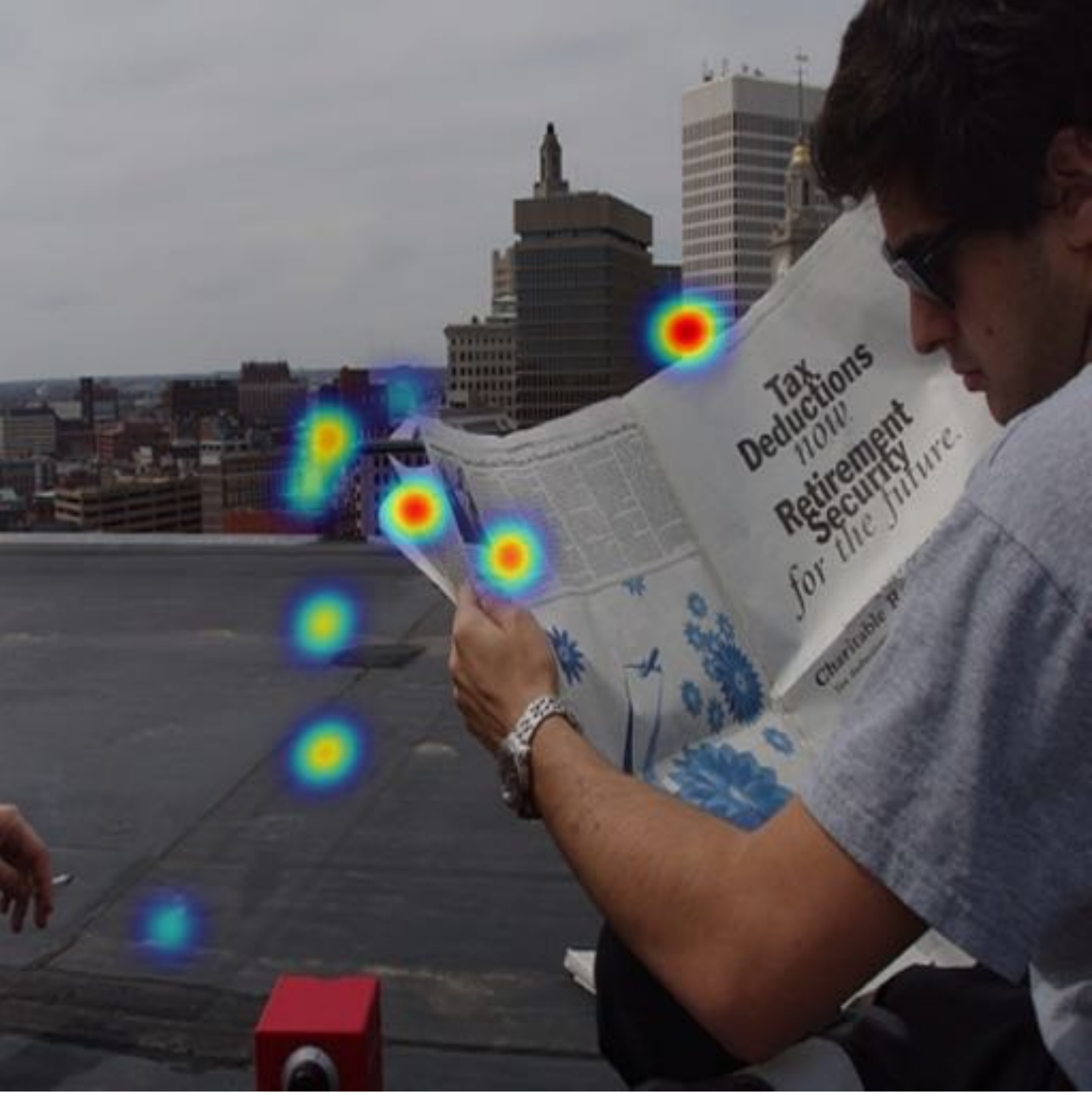} %
        
    \end{subfigure}
 \centering
    \begin{subfigure}{.160\textwidth}
       \centering
        \includegraphics[width=.95\linewidth]{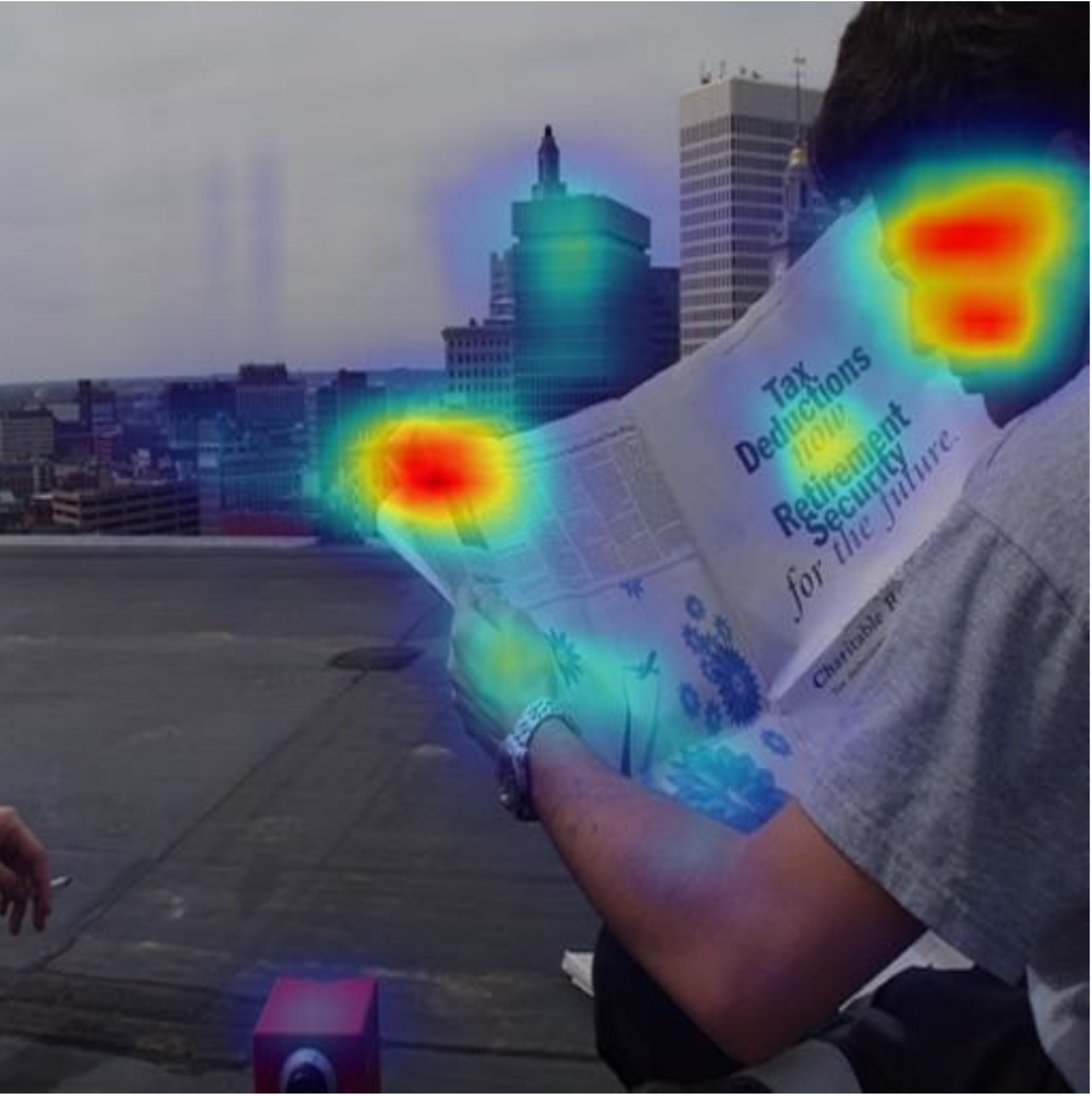} %
        
    \end{subfigure}

    \centering
    \begin{subfigure}{.160\textwidth}
       \centering
        \includegraphics[width=.95\linewidth]{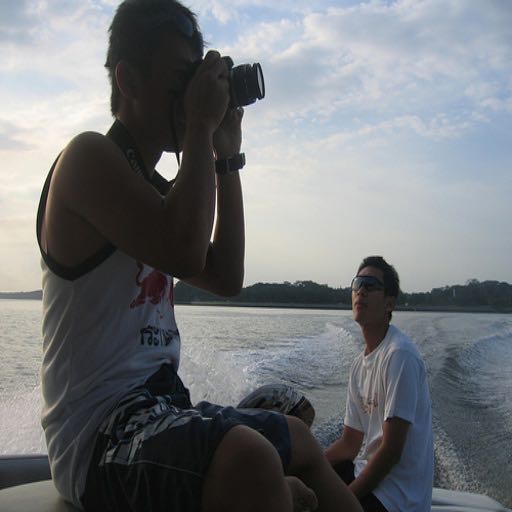} %
        
    \end{subfigure}
     \begin{subfigure}{.160\textwidth}
       \centering
        \includegraphics[width=.95\linewidth]{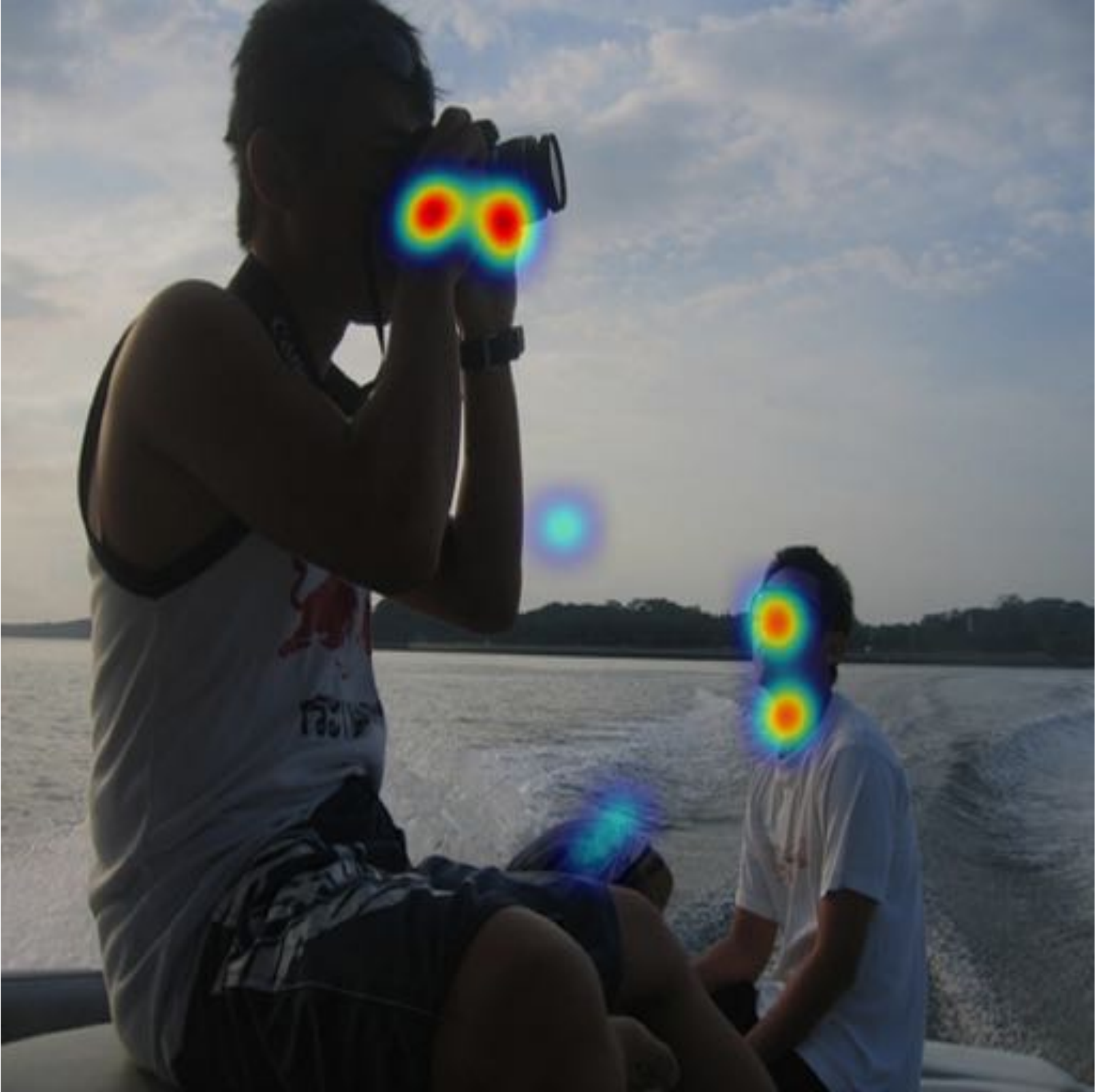} %
       
    \end{subfigure}
     \centering
    \begin{subfigure}{.160\textwidth}
       \centering
        \includegraphics[width=.95\linewidth]{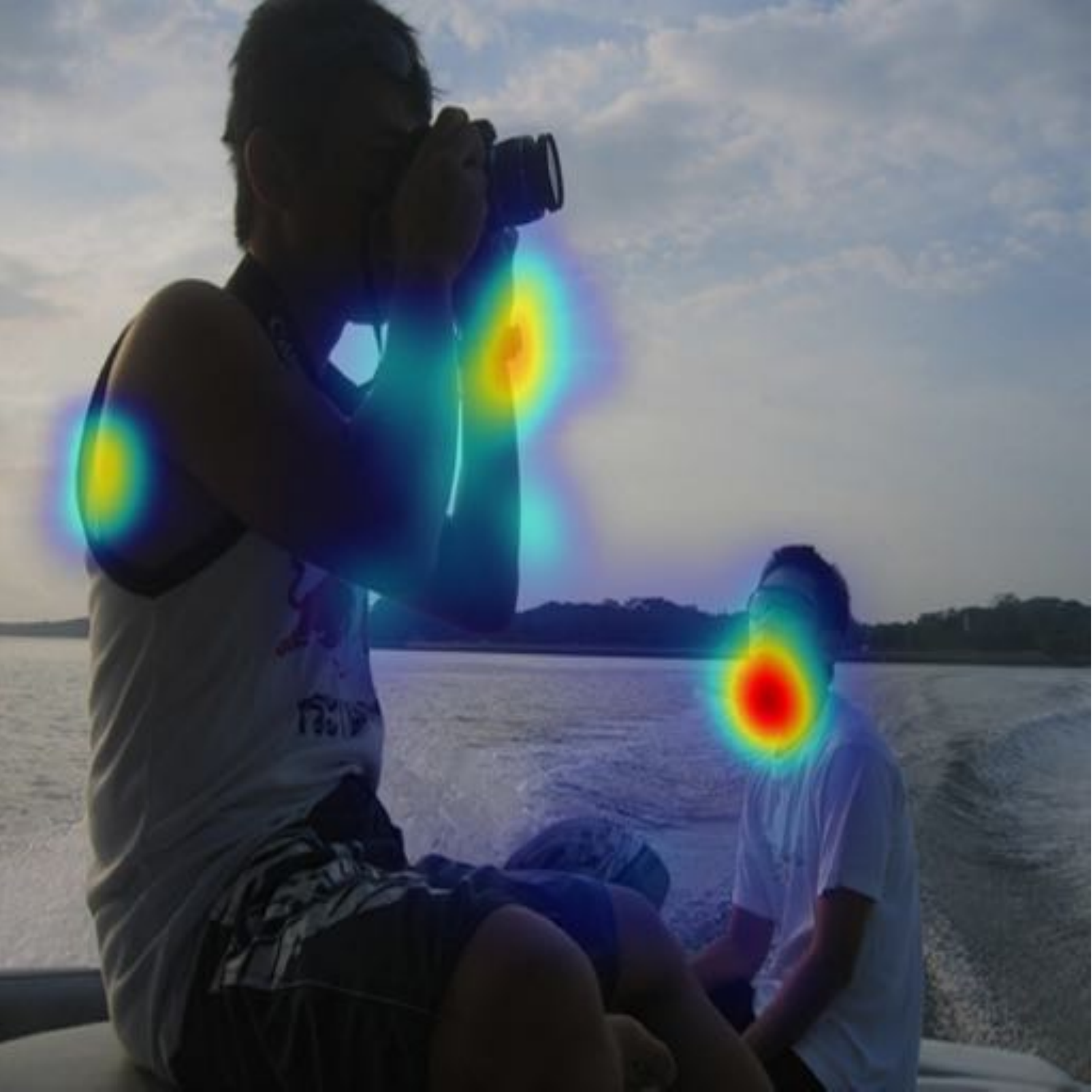} %
        
    \end{subfigure}
 \centering
    \begin{subfigure}{.160\textwidth}
       \centering
       \includegraphics[width=.95\linewidth]{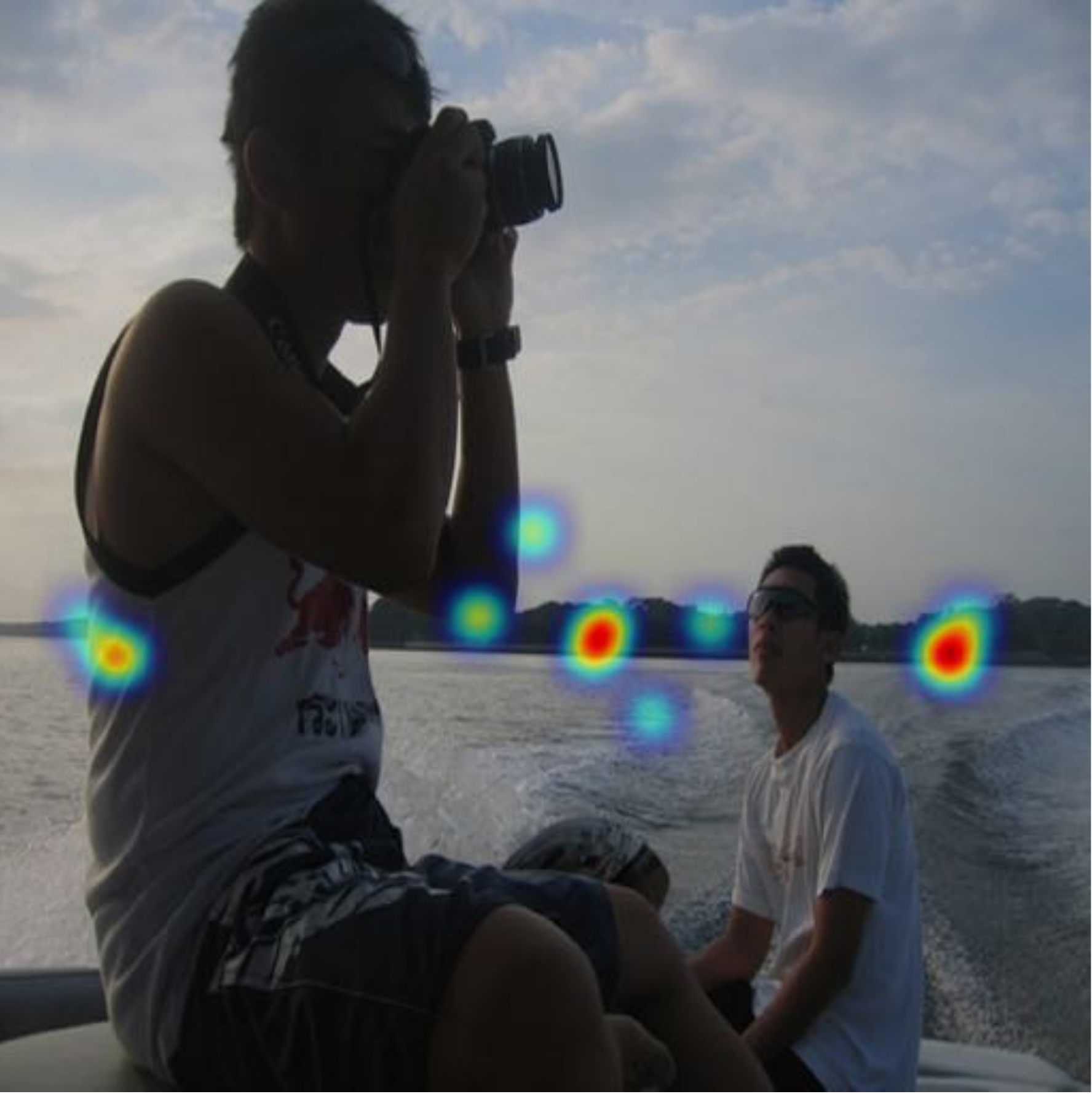} %
    
    \end{subfigure}
 \centering
    \begin{subfigure}{.160\textwidth}
       \centering
         \includegraphics[width=.95\linewidth]{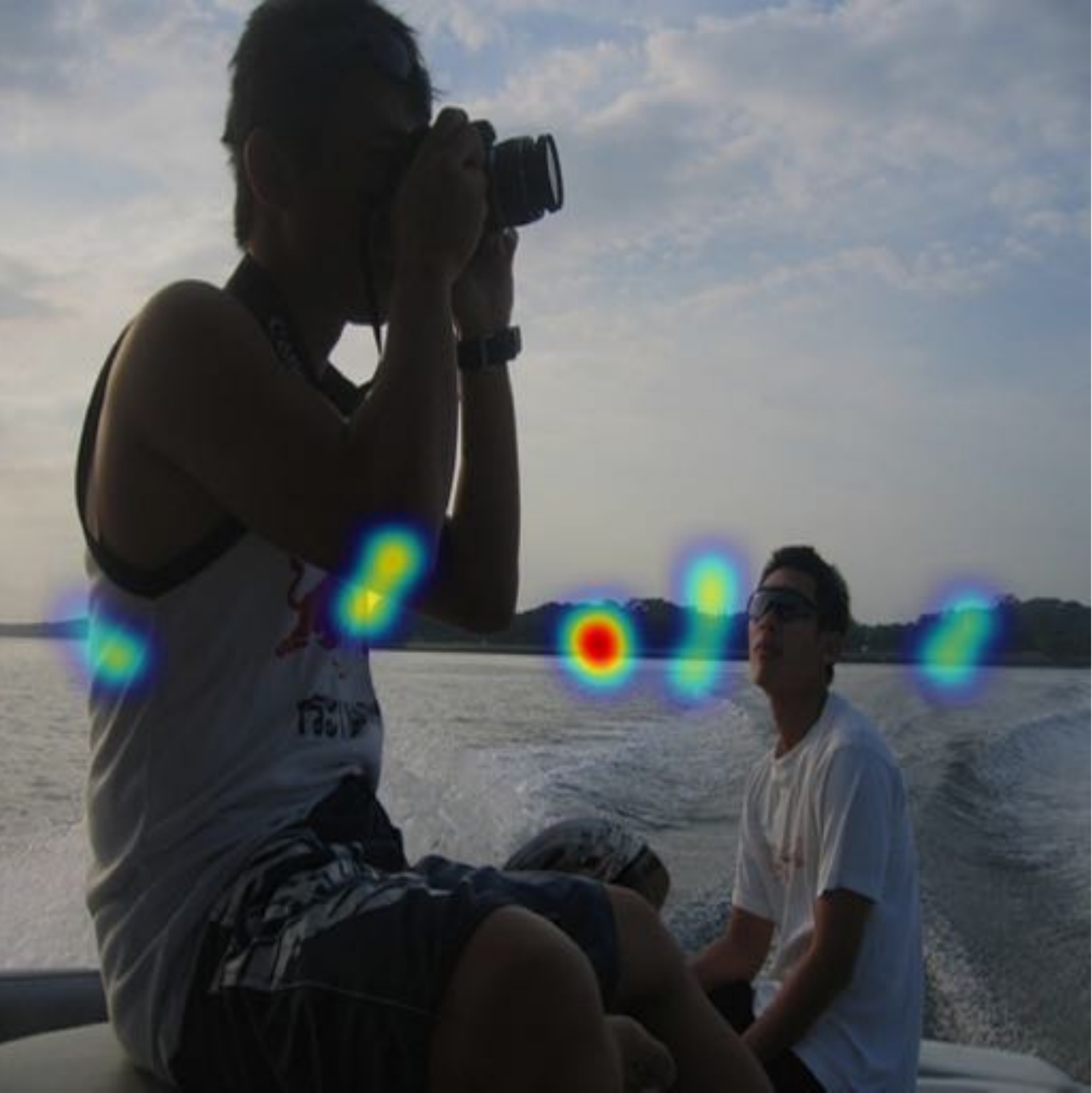} %
        
    \end{subfigure}
 \centering
    \begin{subfigure}{.160\textwidth}
       \centering
        \includegraphics[width=.95\linewidth]{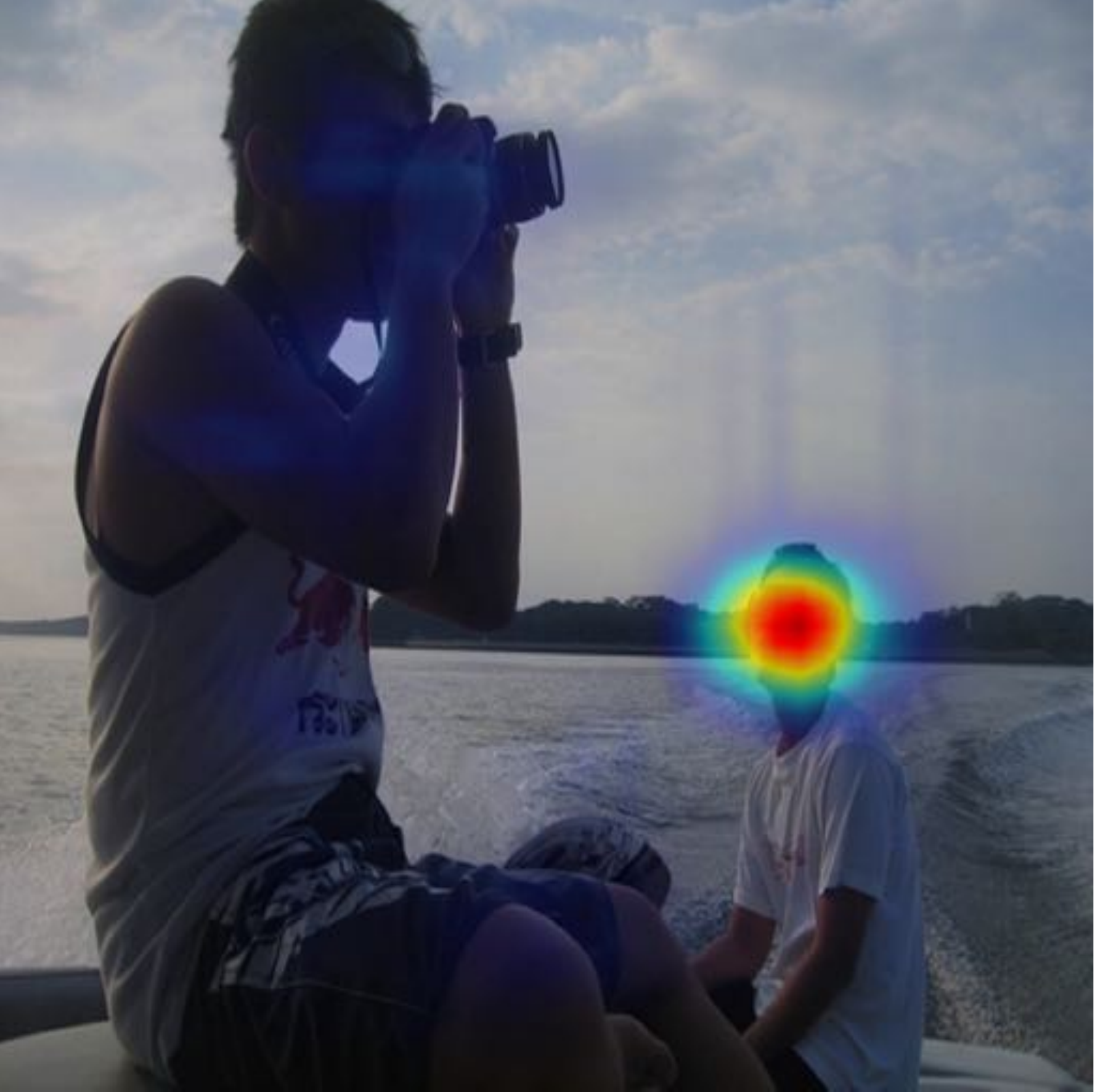} %
        
    \end{subfigure}

     
    \centering
    \begin{subfigure}{.160\textwidth}
       \centering
        \includegraphics[width=.95\linewidth]{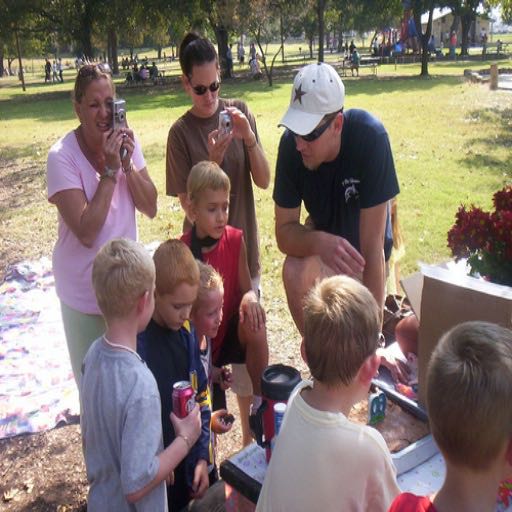} %
        
    \end{subfigure}
     \begin{subfigure}{.160\textwidth}
       \centering
        \includegraphics[width=.95\linewidth]{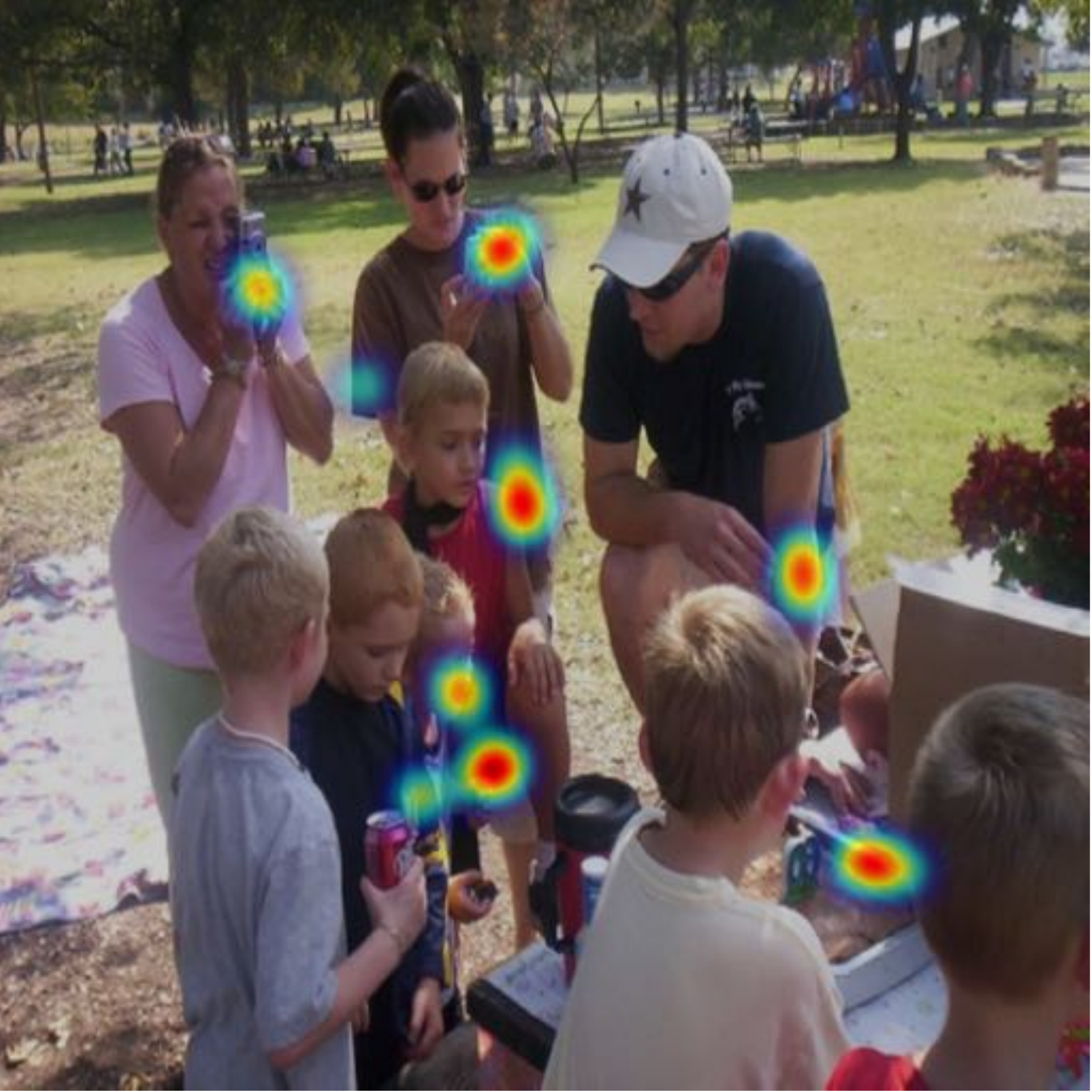} %
       
    \end{subfigure}
     \centering
    \begin{subfigure}{.160\textwidth}
       \centering
        \includegraphics[width=.95\linewidth]{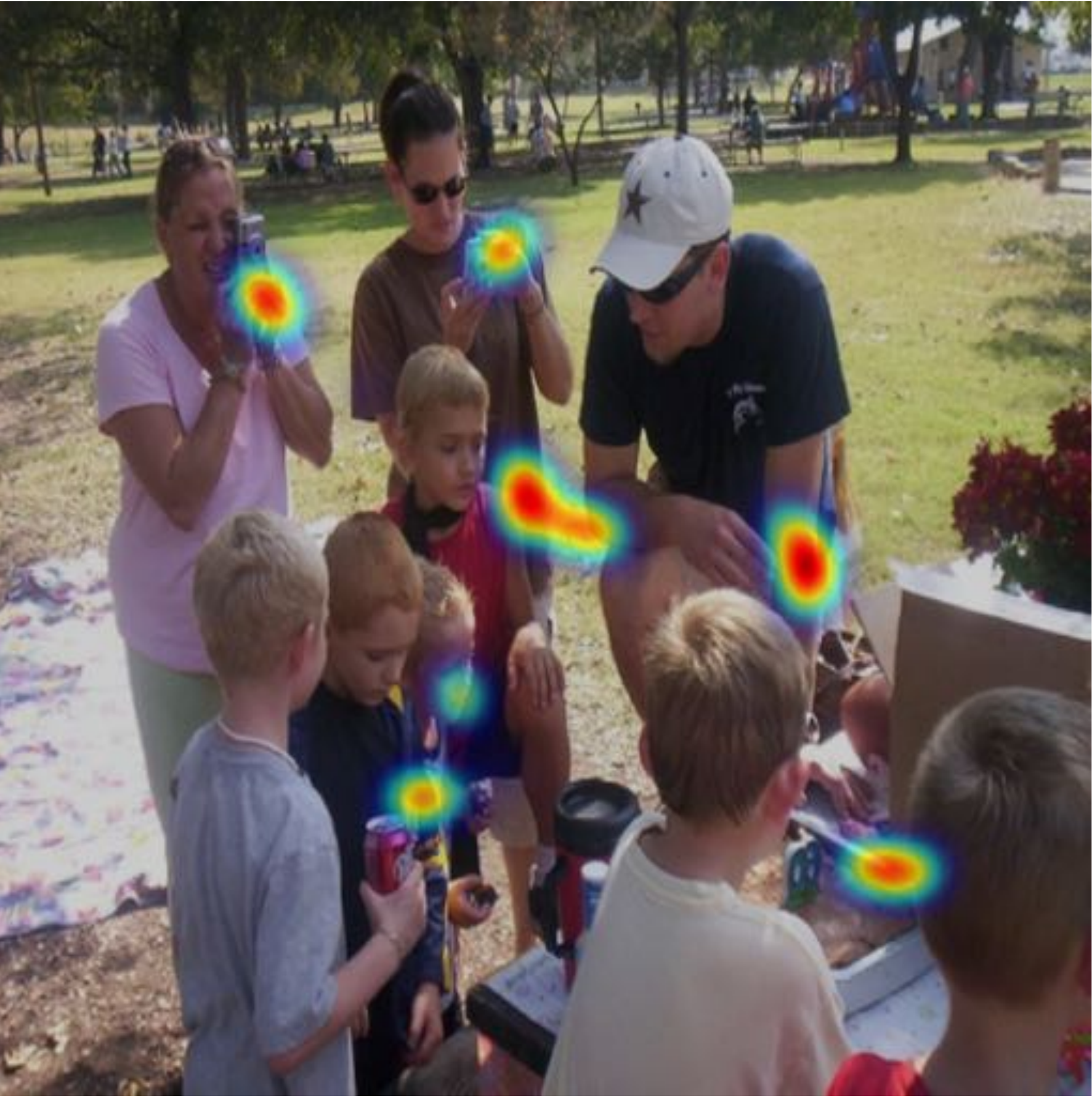} %
        
    \end{subfigure}
 \centering
    \begin{subfigure}{.160\textwidth}
       \centering
       \includegraphics[width=.95\linewidth]{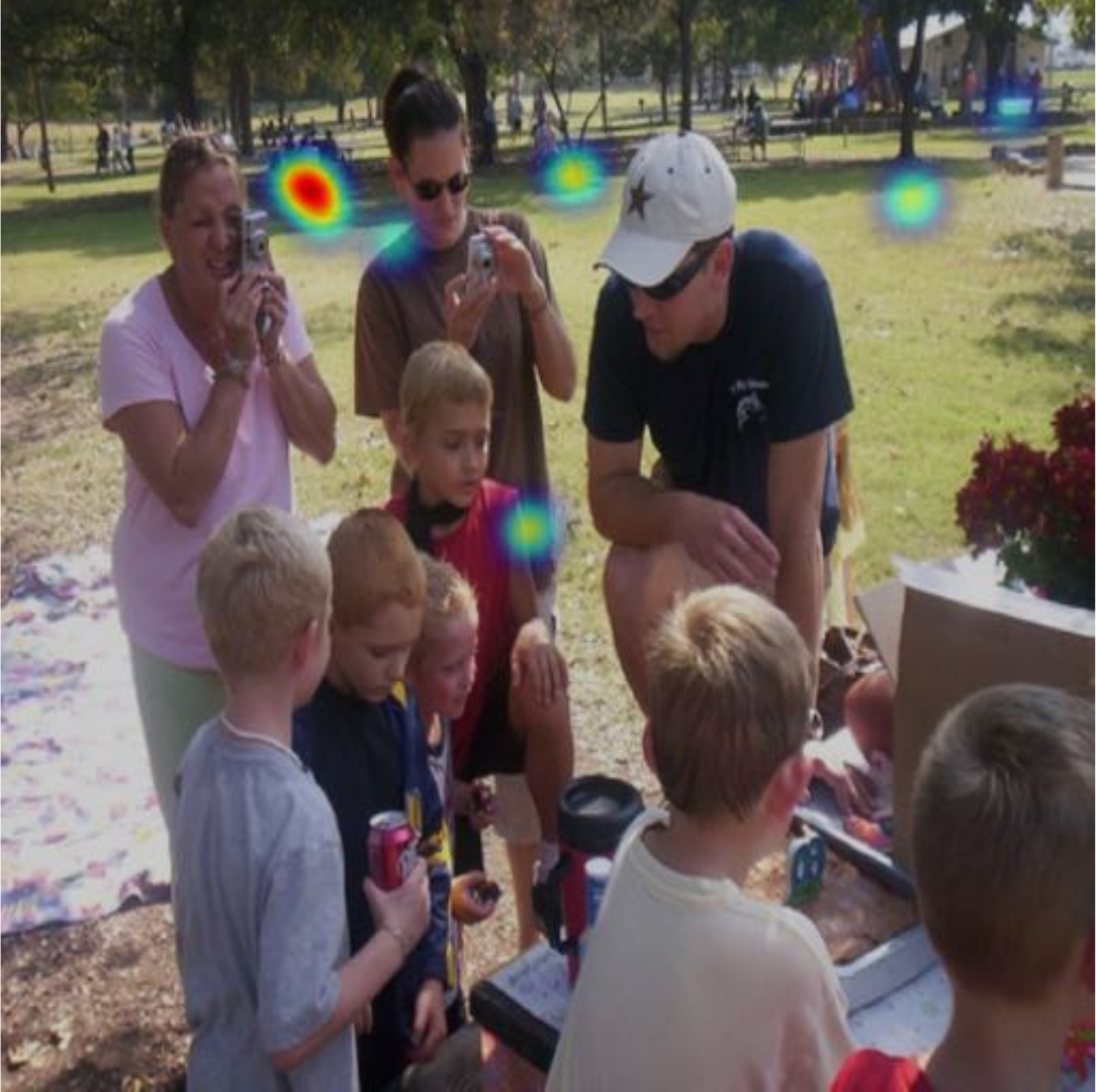} %
    
    \end{subfigure}
 \centering
    \begin{subfigure}{.160\textwidth}
       \centering
         \includegraphics[width=.95\linewidth]{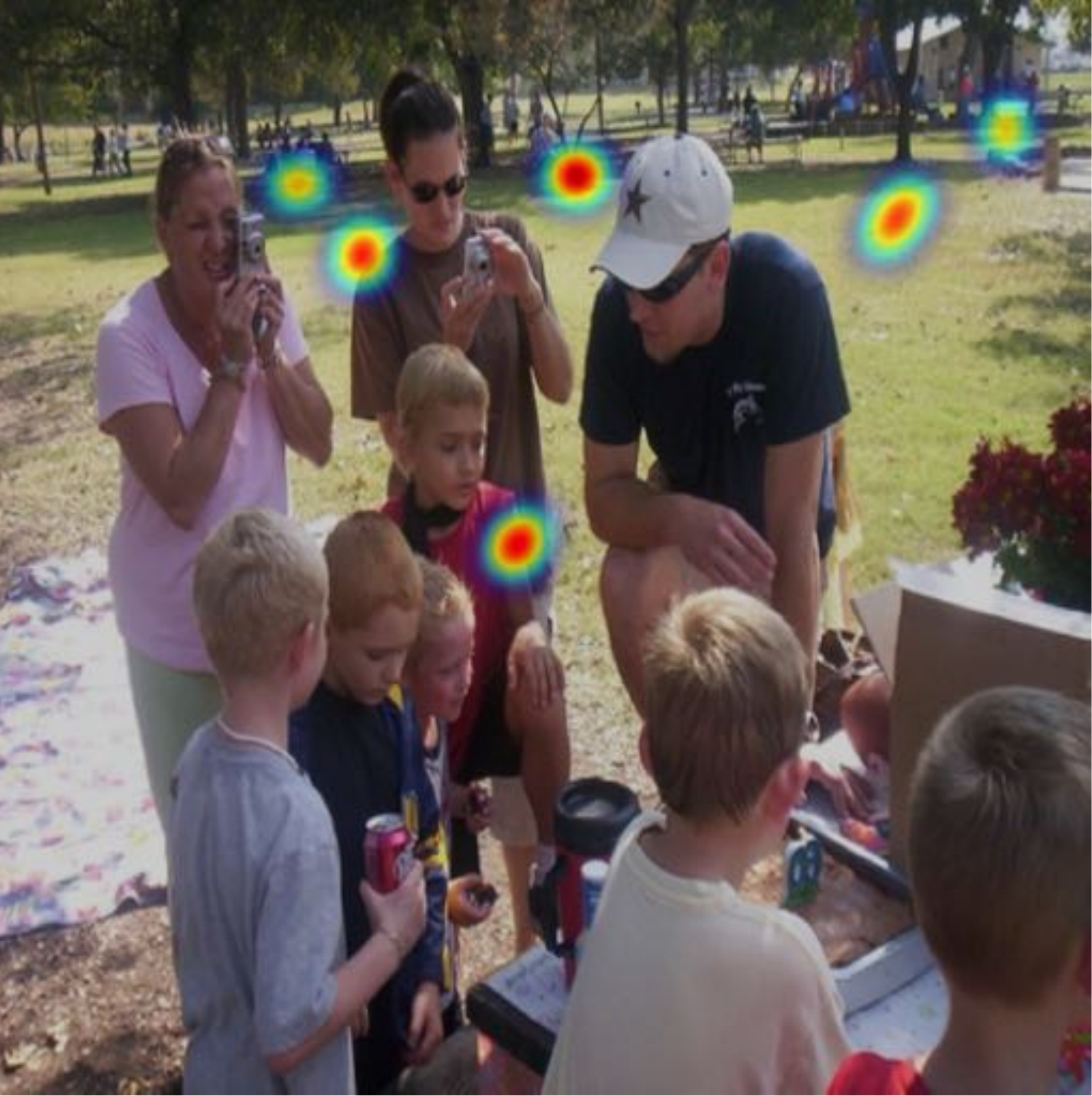} %
        
    \end{subfigure}
 \centering
    \begin{subfigure}{.160\textwidth}
       \centering
        \includegraphics[width=.95\linewidth]{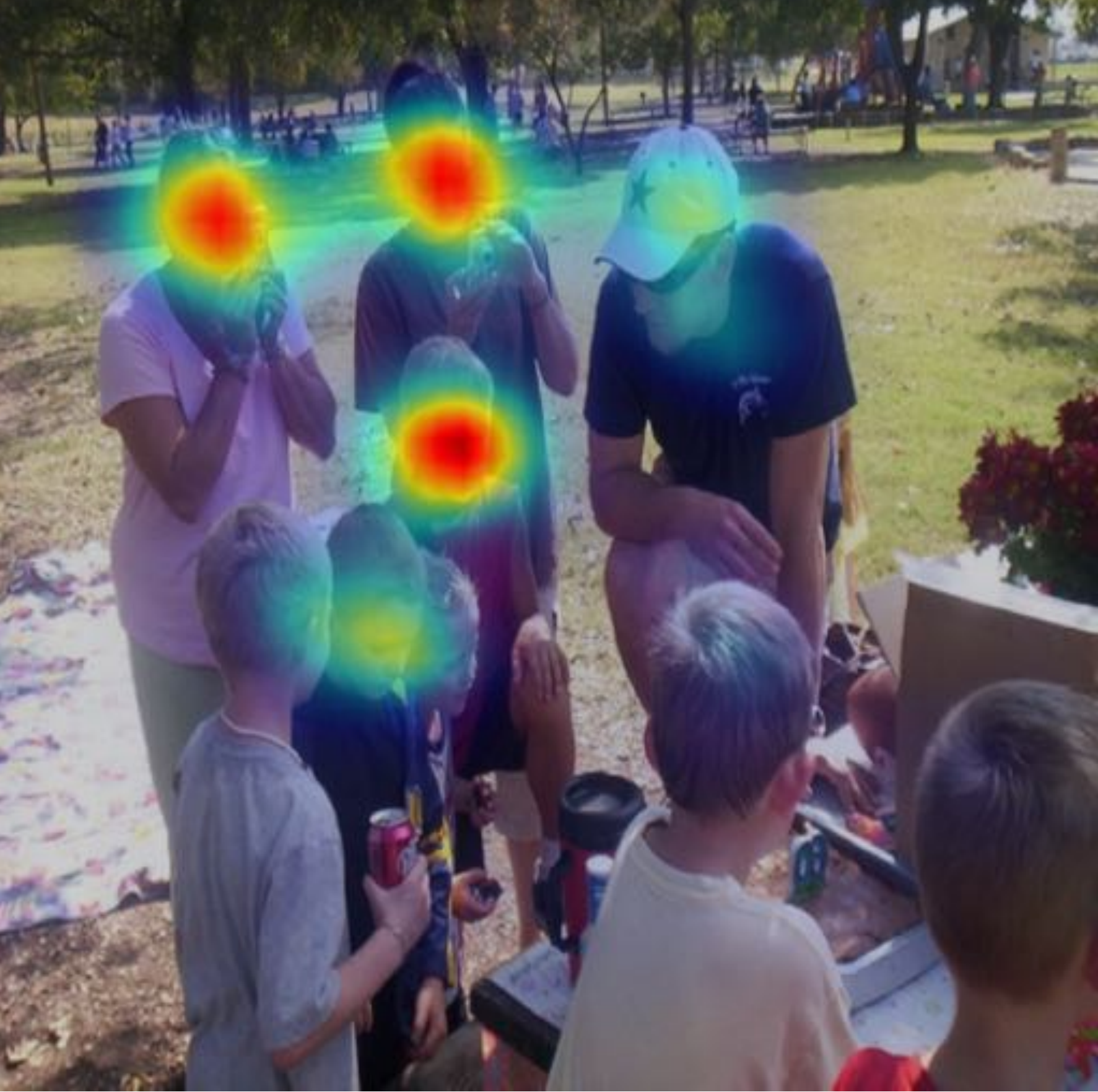} %
        
    \end{subfigure}

    \centering
    \begin{subfigure}{.160\textwidth}
       \centering
        \includegraphics[width=.95\linewidth]{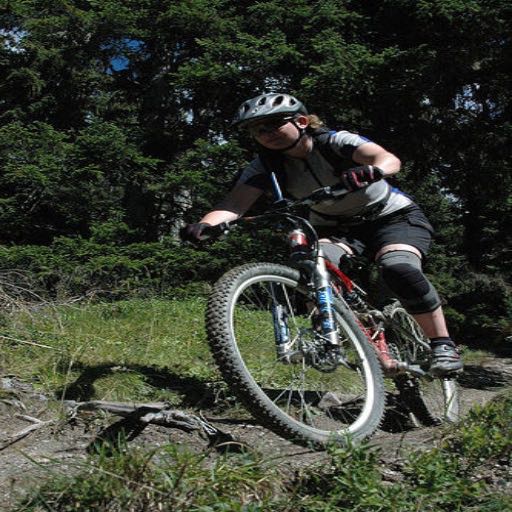} %
        
    \end{subfigure}
     \begin{subfigure}{.160\textwidth}
       \centering
        \includegraphics[width=.95\linewidth]{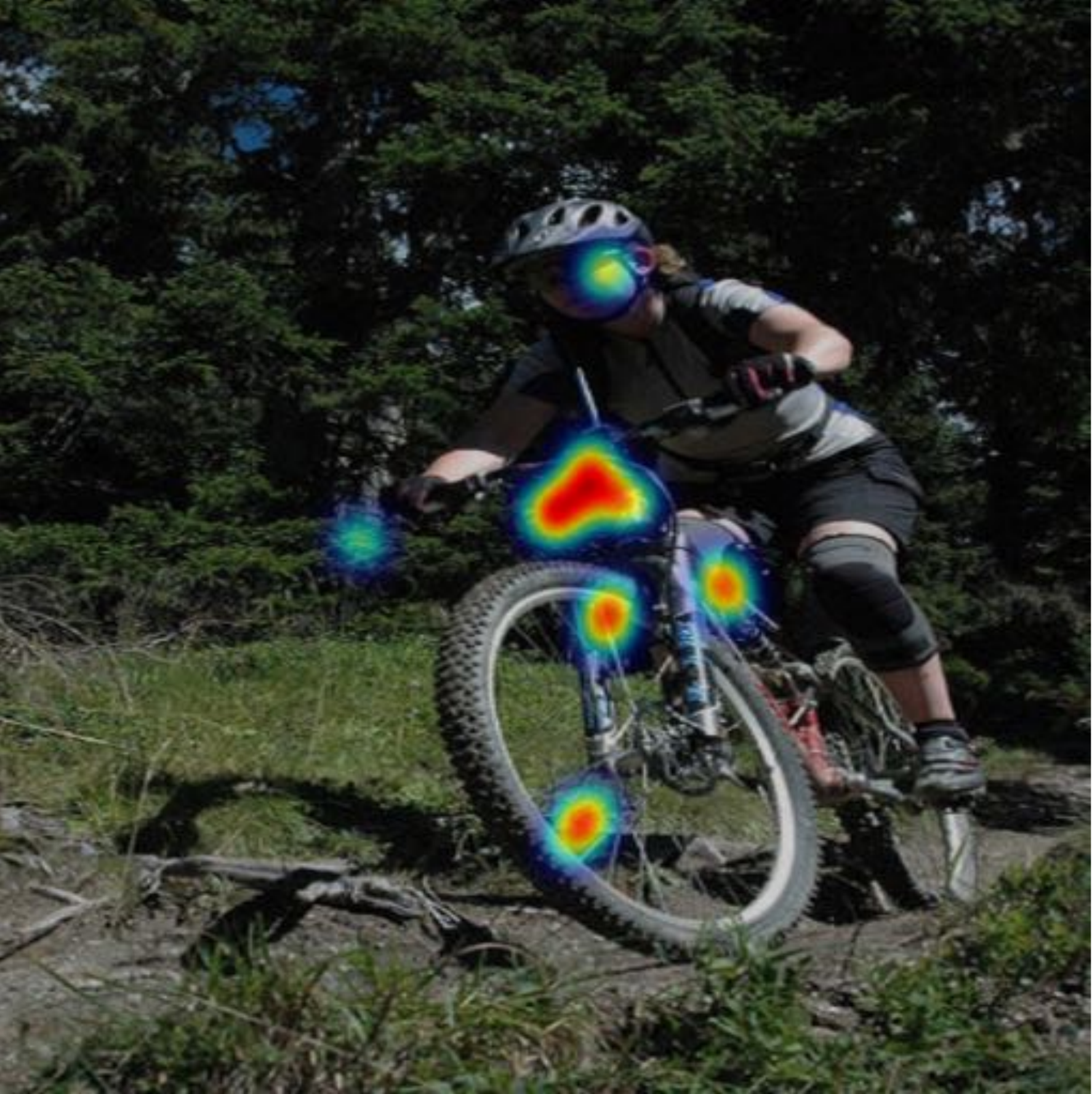} %
       
    \end{subfigure}
     \centering
    \begin{subfigure}{.160\textwidth}
       \centering
        \includegraphics[width=.95\linewidth]{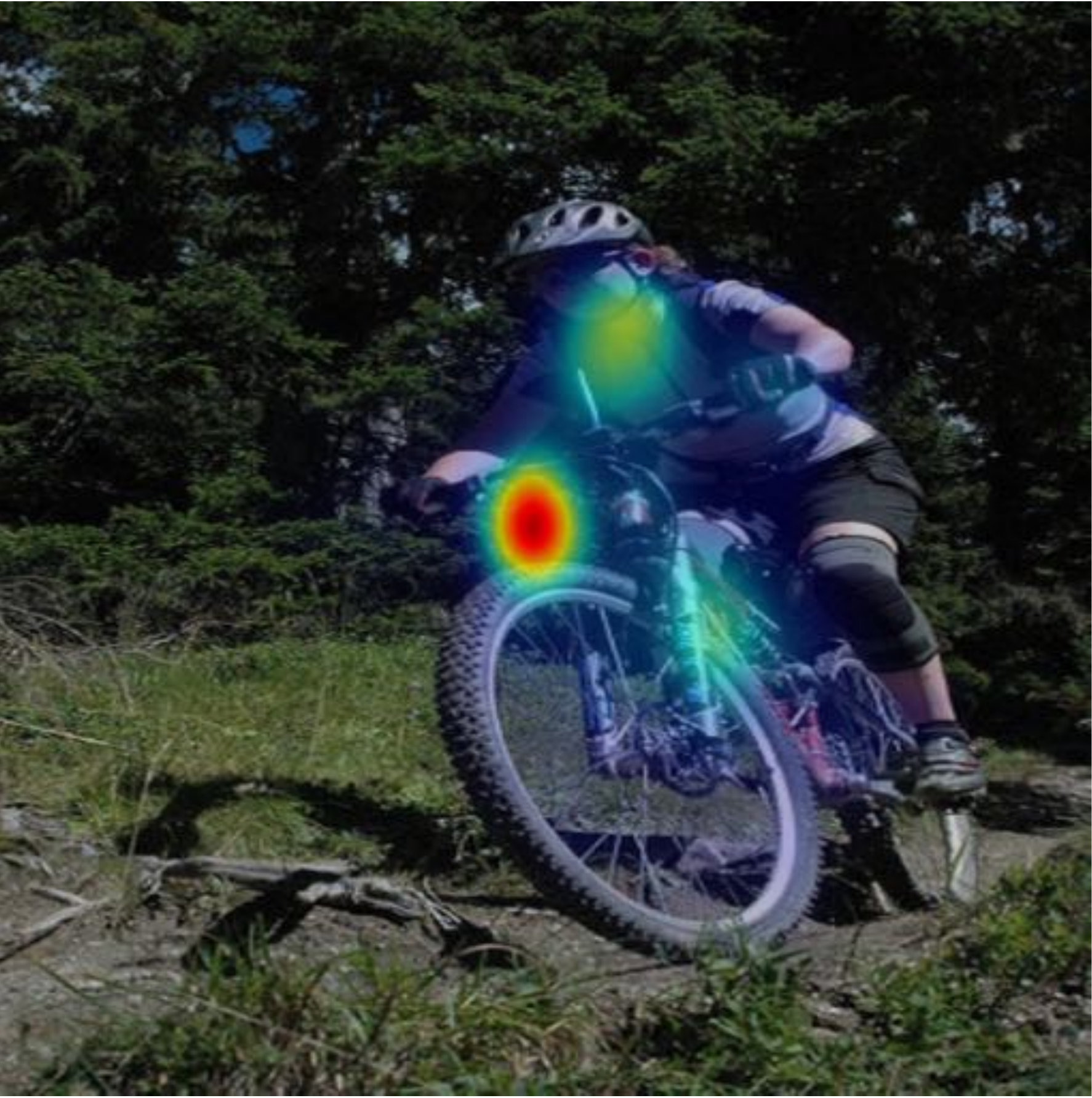} %
        
    \end{subfigure}
 \centering
    \begin{subfigure}{.160\textwidth}
       \centering
       \includegraphics[width=.95\linewidth]{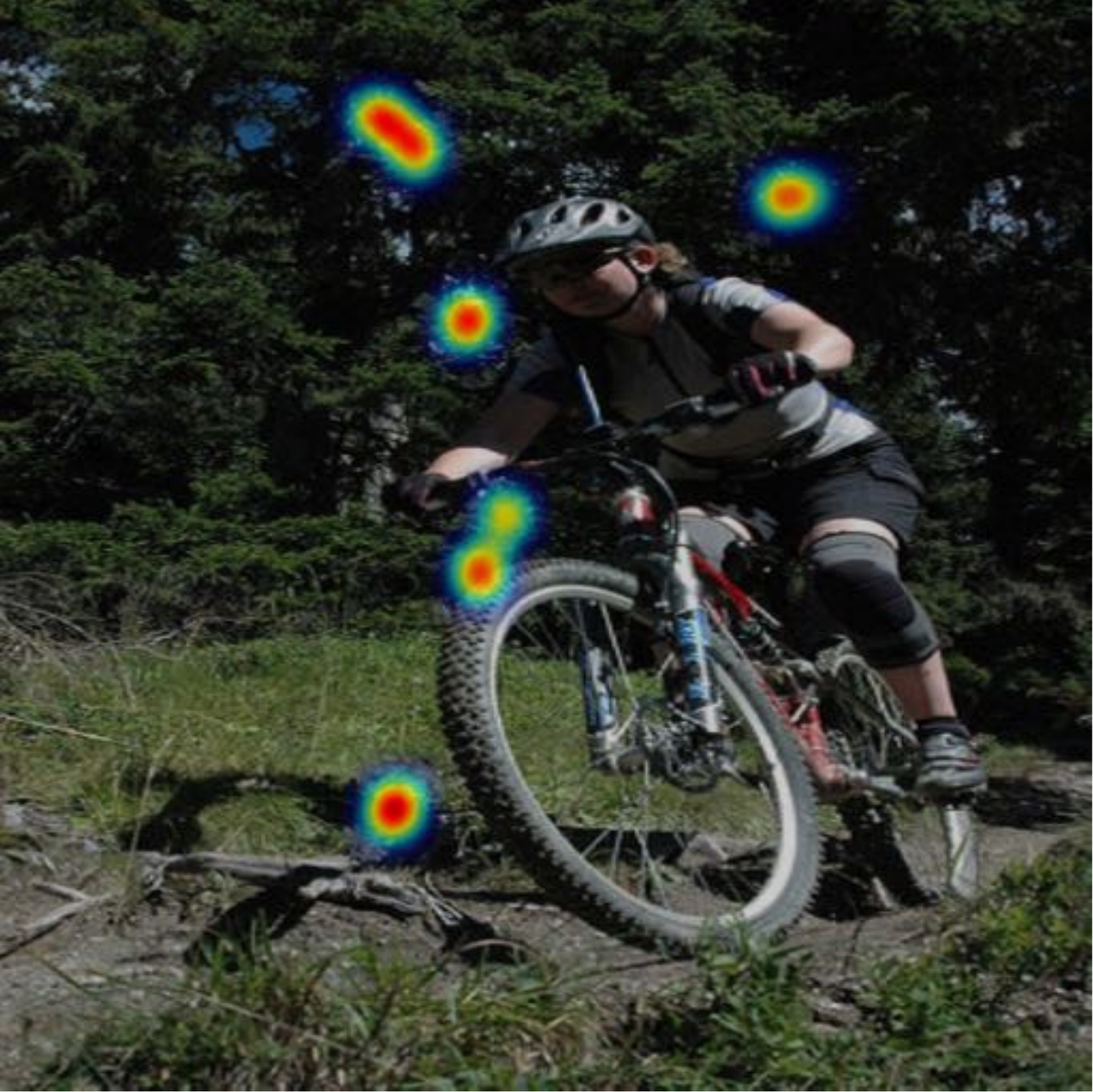} %
    
    \end{subfigure}
 \centering
    \begin{subfigure}{.160\textwidth}
       \centering
         \includegraphics[width=.95\linewidth]{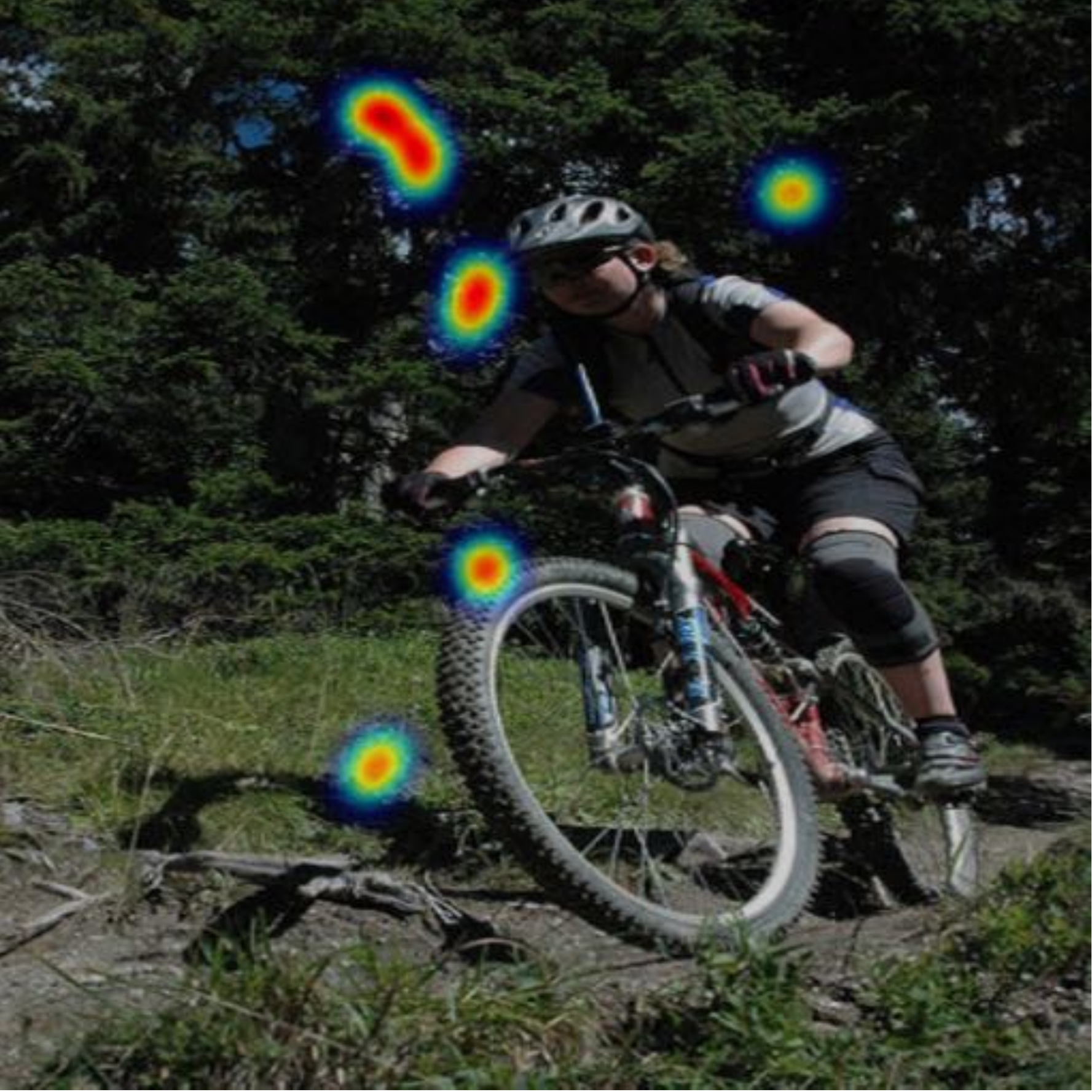} %
        
    \end{subfigure}
 \centering
    \begin{subfigure}{.160\textwidth}
       \centering
        \includegraphics[width=.95\linewidth]{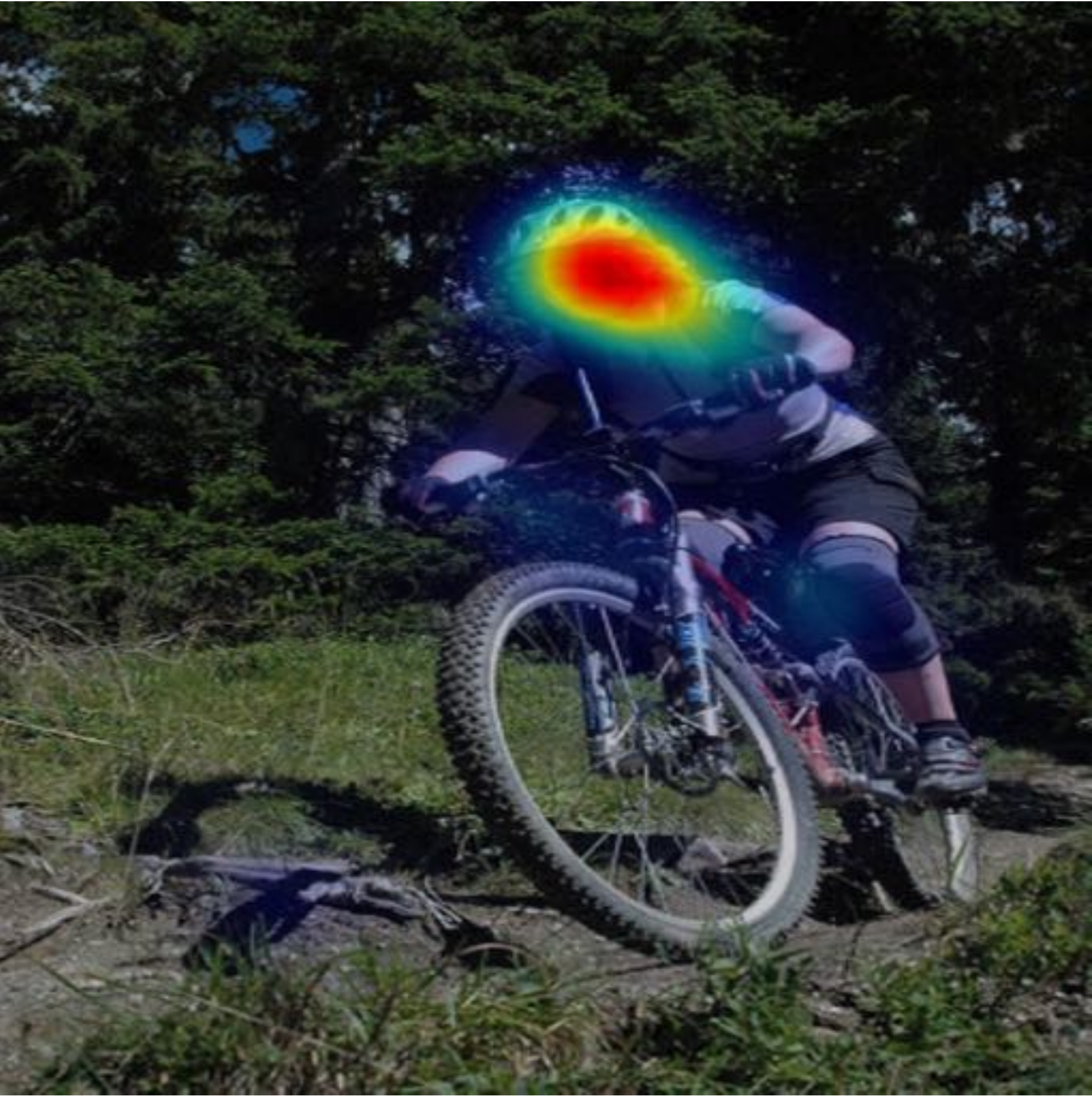} %
        
    \end{subfigure}

        \centering
    \begin{subfigure}{.160\textwidth}
       \centering
        \includegraphics[width=.95\linewidth]{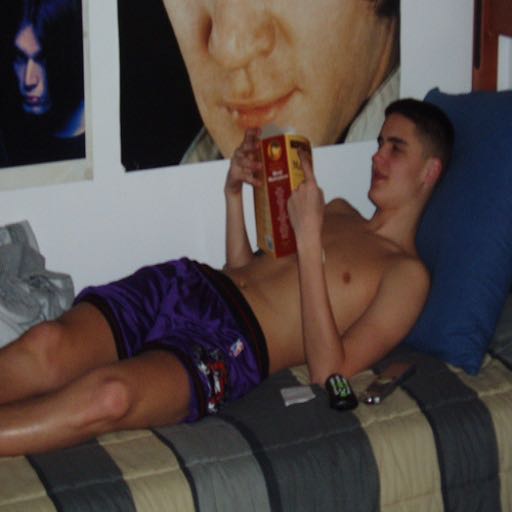} %
        \caption{Image}
    \end{subfigure}
     \begin{subfigure}{.160\textwidth}
       \centering
        \includegraphics[width=.95\linewidth]{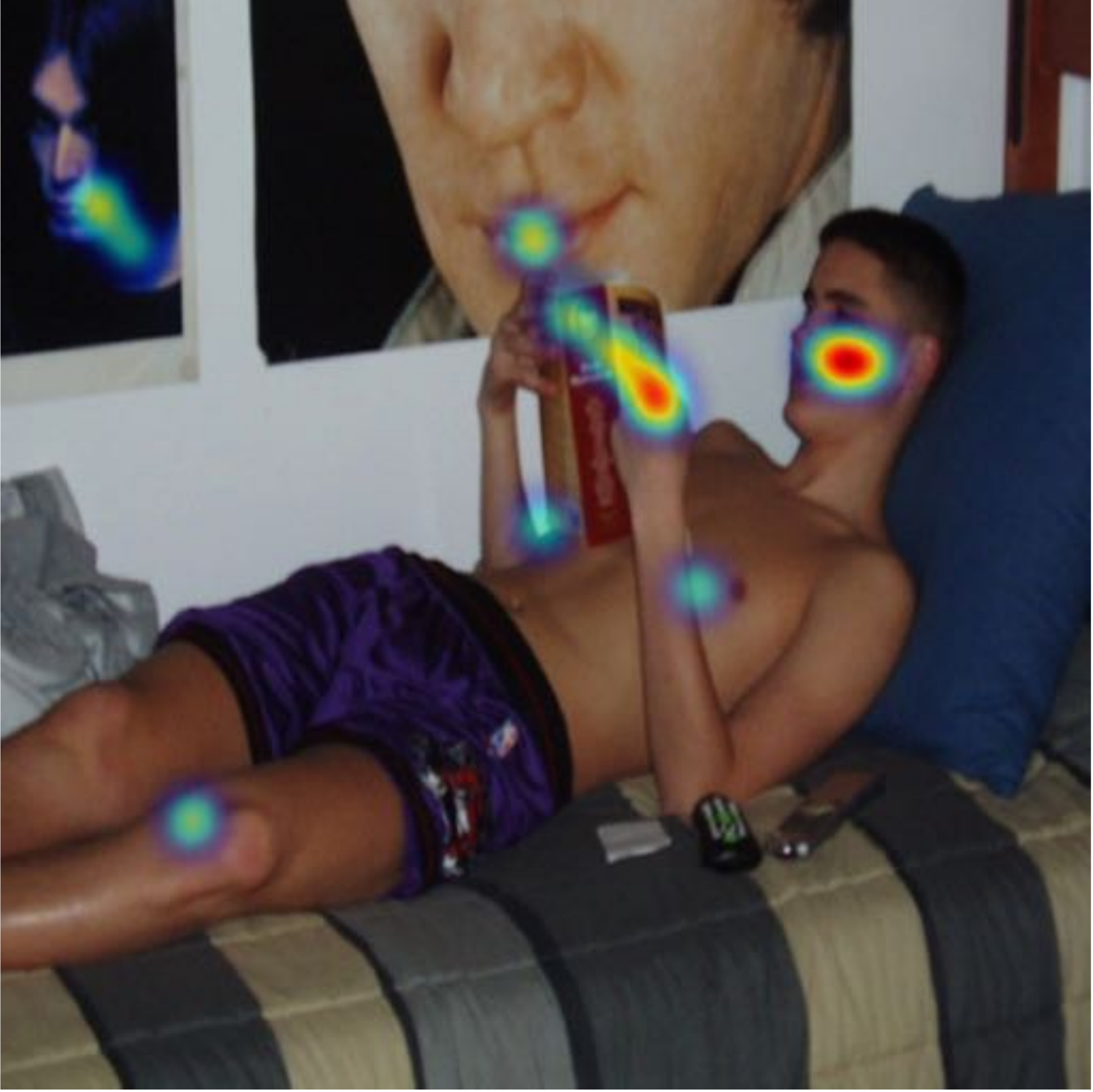} %
        \caption{Action rec GT}
    \end{subfigure}
     \centering
    \begin{subfigure}{.160\textwidth}
       \centering
        \includegraphics[width=.95\linewidth]{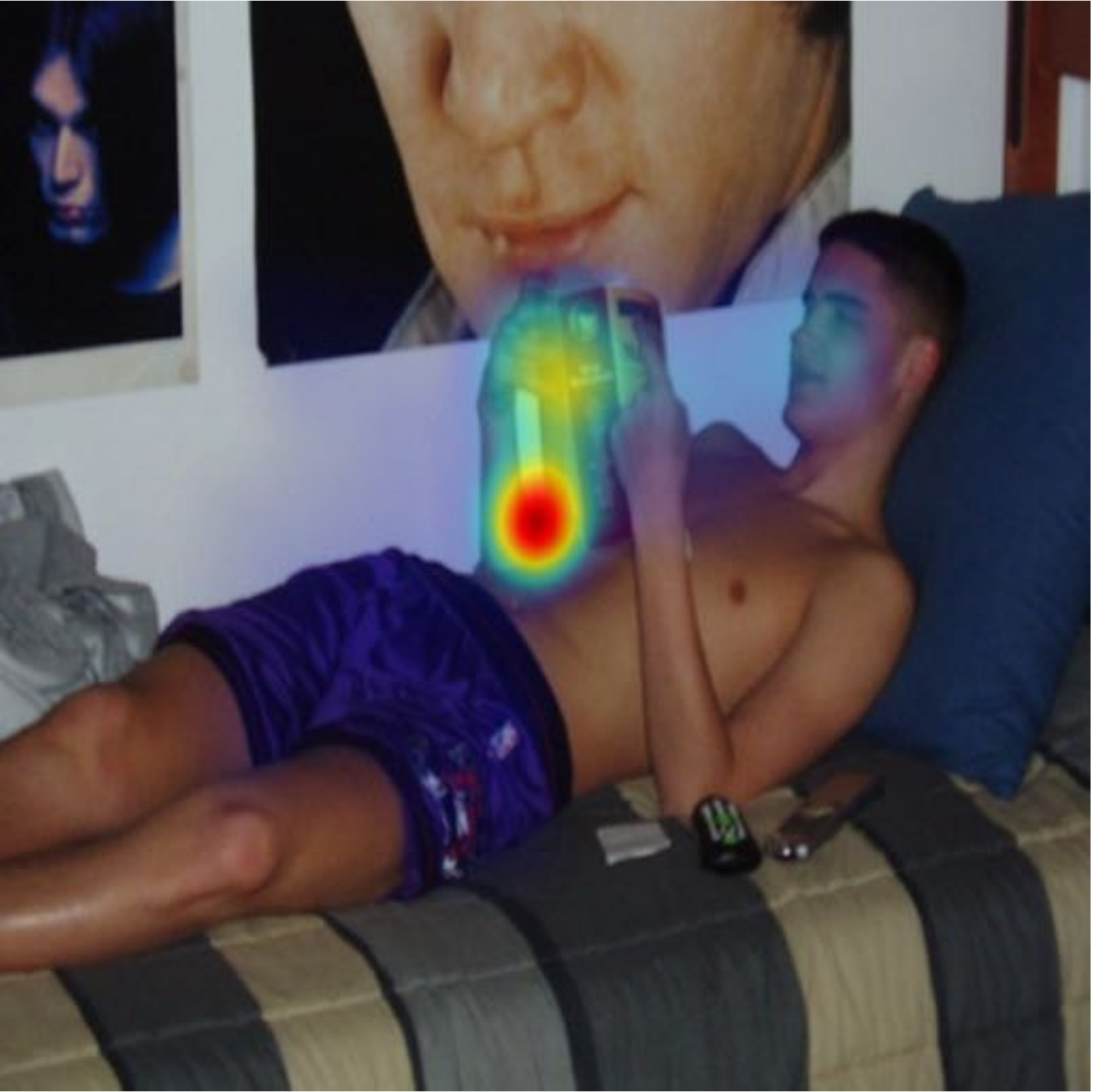} %
        \caption{Our action rec}
    \end{subfigure}
 \centering
    \begin{subfigure}{.160\textwidth}
       \centering
        \includegraphics[width=.95\linewidth]{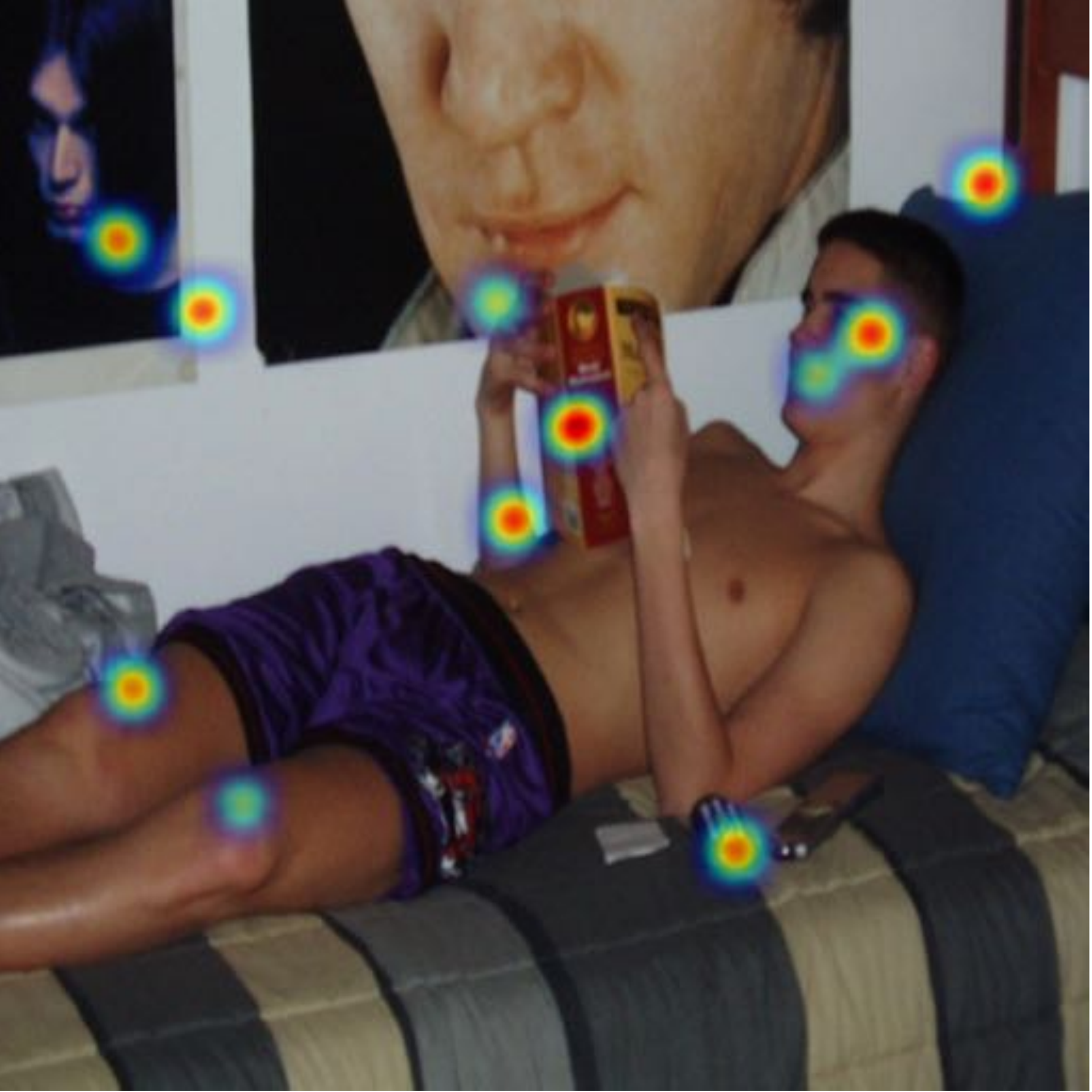} %
        \caption{Context rec GT}
    \end{subfigure}
 \centering
    \begin{subfigure}{.160\textwidth}
       \centering
        \includegraphics[width=.95\linewidth]{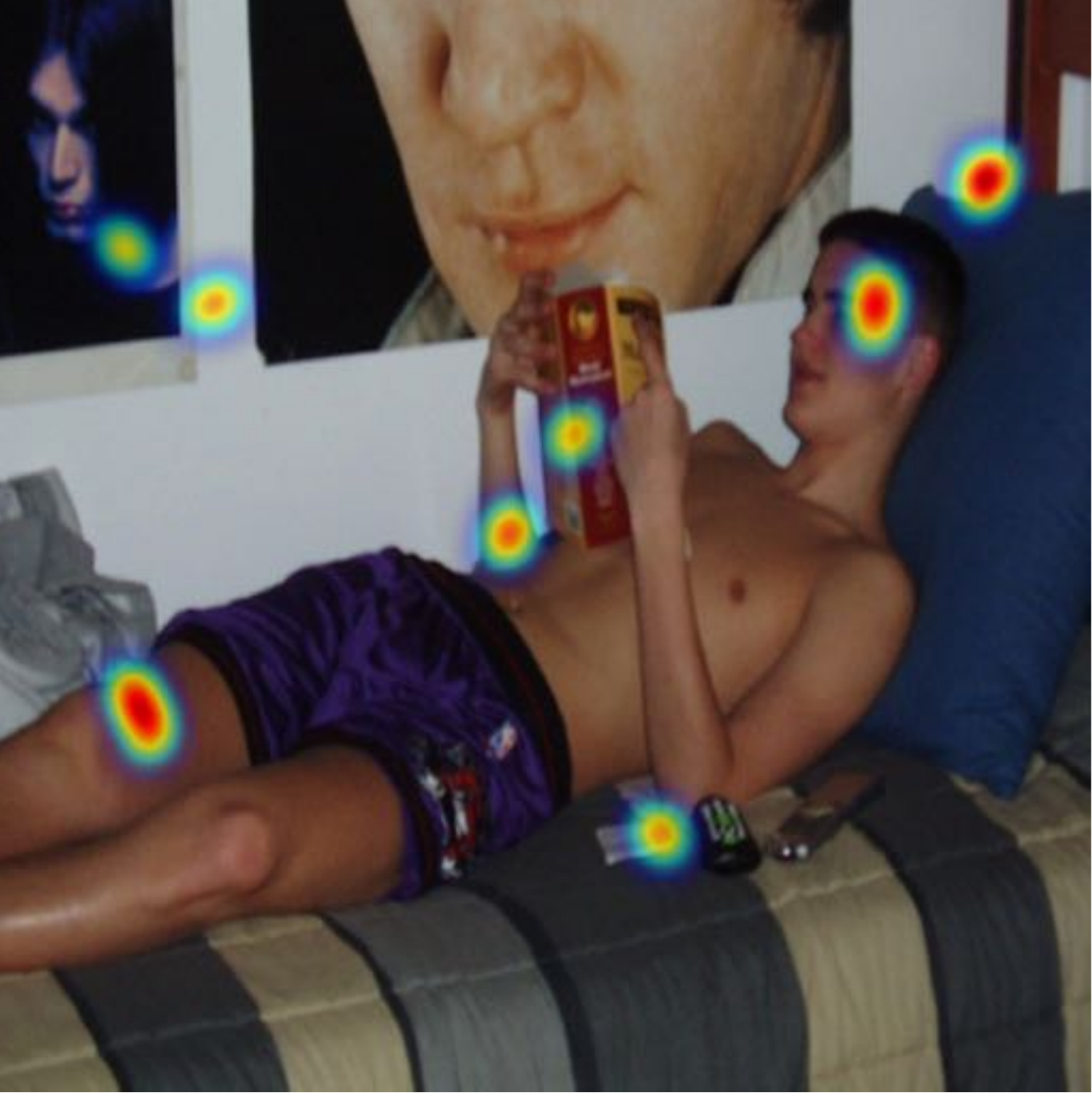} %
        \caption{Our context rec}
    \end{subfigure}
 \centering
    \begin{subfigure}{.160\textwidth}
       \centering
        \includegraphics[width=.95\linewidth]{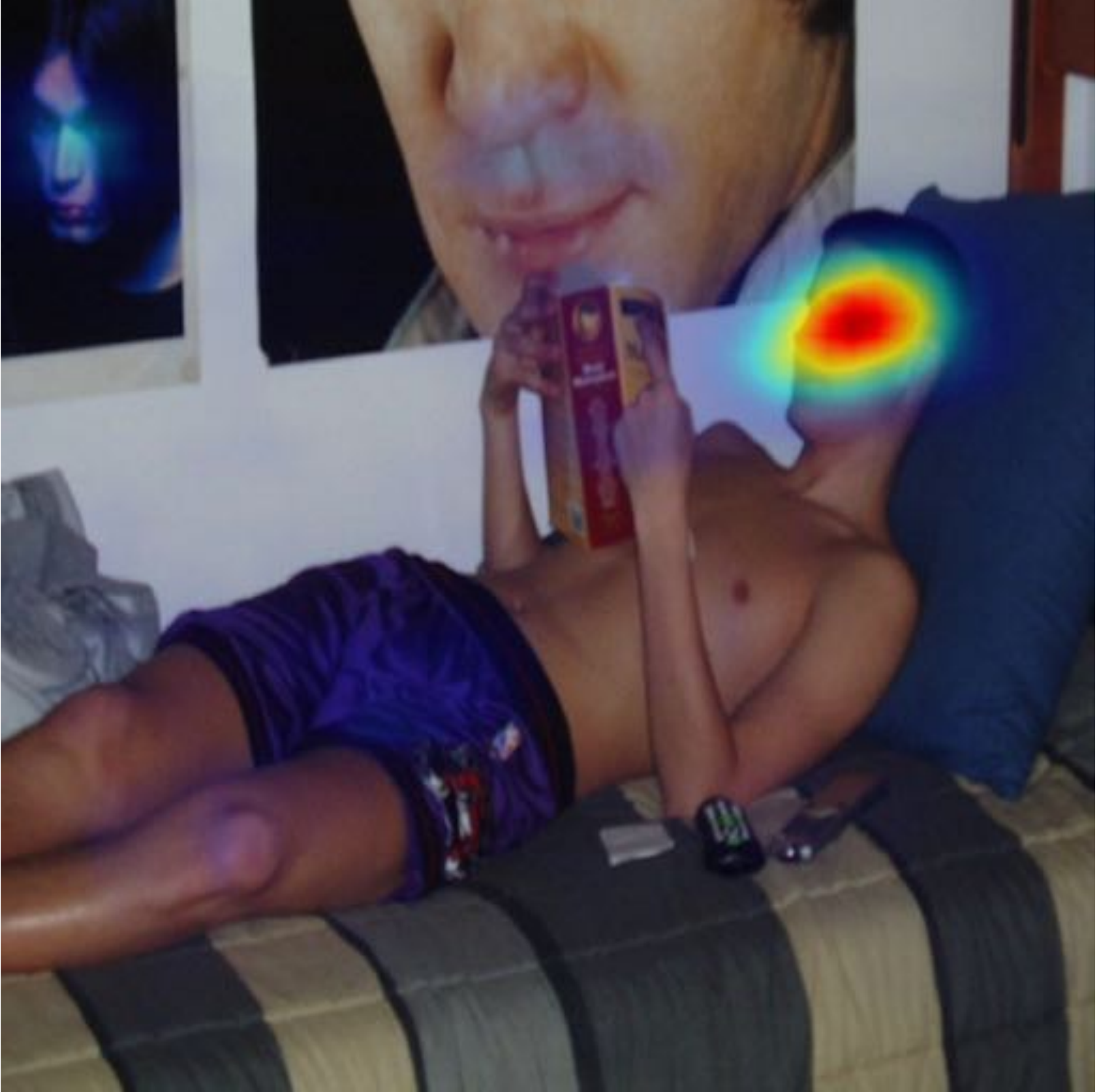} %
        \caption{ML-net}
    \end{subfigure}

     \caption{Qualitative results for VOCA-2012 dataset and comparisons to the state-of-the-art.}
			\vspace{-3mm}
        \label{fig:fig_action_context_rec}
\end{figure*}

To provide qualitative insight, some predicted maps along with ground truth and baseline ML-net \cite{deep-ml} predictions are given in  Fig. \ref{fig:fig_action_context_rec}. In the first column we show the input image, and columns ``Action rec GT'' and ``Context rec GT'' depict the ground truth saliency maps for the respective tasks. In columns  ``Our action rec'' and ``Our context rec'' we show the respective predictions from our model, and finally the  column `ml-Net' contains the prediction from the ML-net \cite{deep-ml} baseline.   Observing columns ``Action rec GT '' and ``Context rec GT'' one can clearly see how the tasks differ based on the different saliency patterns. Yet, the proposed model is able to capture these different semantics within a single network which is trained together for all the tasks. As shown in Fig. \ref{fig:fig_action_context_rec}, it has efficiently identified the image saliency from low level features as well as task dependent saliency factors from high level cues such as trees, furniture and roads. Furthermore, the single learning process and the incorporation of a memory architecture renders the plausibility of retaining the semantical relationships among different tasks and how users adapt to those. 

Tab. \ref{tab:tab_2} shows the performance of the proposed model with 5 baselines for the MIT-PD test set. The first baseline ``Scene Context'' \cite{modellin-search} utilises colour and orientation features where as ``Combined'' \cite{modellin-search} incorporates both scene context features and higher level features from a person detector \cite{dalal2006human}. Even with such explicit modelling of the task, this baseline fails to generate accurate predictions suggesting the subjective nature of the task specific viewing. With the aid of the associative memory of the proposed model we successfully capture those underlying factors.

\begin{table}[!h]
  \centering
  \begin{adjustbox}{width=.98\linewidth,center}
  \begin{tabular}{|c|c|c|}
 \hline
    & \multicolumn{2}{|c|}{\textbf{Scene Type}} \\
        \cline{2-3}
     & \textbf{Target Present}  & \textbf{Target absent}\\
           \cline{2-3}
   \textbf{Saliency Models} & \textbf{AUC}  & \textbf{AUC}  \\
    \hline
     \hline
   Scene Context \cite{modellin-search} & 0.844  & 0.845 \\
    \hline
   Combined \cite{modellin-search} & 0.896 & 0.877 \\
     \hline
  ML-net \cite{deep-ml}& 0.901 & 0.881\\
    \hline
    SalGAN \cite{pan2017salgan}&0.910 & 0.887\\
      \hline
    cGAN \cite{pix2pix} & 0.923 & 0.899  \\
    \hline
   \textbf{MC-GAN (proposed)} &\textbf{0.942}  & \textbf{0.903} \\
    \hline
    \hline
  Human \cite{modellin-search} & 0.955 & 0.930 \\
 \hline
  \end{tabular}
  \end{adjustbox}
  \caption{Experimental evaluation on MIT-PD test set}
			\vspace{-3mm}
  \label{tab:tab_2}
\end{table}

In Tab. \ref{tab:tab_3} and Tab. \ref{tab:tab_4} we present the evaluations of NSS, CC and SM metrics. In order to evaluate ML-net, SalGAN and cGAN models we utilise the implementation of the algorithm released by the authors. When comparing the results between the ML-net \cite{deep-ml}, cGAN \cite{pix2pix} and SalGAN \cite{pan2017salgan} models and proposed MC-GAN model a considerable gain in performance is observed in all the metrics considered, which emphasises a greater agreement between predicted and ground truth saliency maps. We were unable to compare other baselines using these metrics due to the unavailability of public implementations. 

\begin{table}[t]
  \centering
  \begin{adjustbox}{width=.98\linewidth,center}
  \begin{tabular}{|c|c|c|c|c|c|c|}
 \hline
    & \multicolumn{6}{|c|}{\textbf{Task}} \\
        \cline{2-7}
     & \multicolumn{3}{|c|}{\textbf{Target Present}}  & \multicolumn{3}{|c|}{\textbf{Target absent}}\\
           \cline{2-7}
   \textbf{Saliency Models} & \textbf{NSS} & \textbf{CC} & \textbf{SM} & \textbf{NSS} & \textbf{CC} & \textbf{SM}  \\
    \hline
     \hline
     ML-Net \cite{deep-ml} & 1.41 & 0.55  &0.41 &1.22 & 0.43 & 0.38 \\
      \hline
      SalGAN \cite{pan2017salgan}& 1.41 & 0.53  &0.44 &1.20 & 0.42 & 0.35 \\ 
      \hline
   cGAN \cite{pix2pix} & 1.67 & 0.51  &0.59 & 2.02 & 0.41 & 0.52  \\
    \hline
   \textbf{MC-GAN (proposed)} &\textbf{2.17} & \textbf{0.76 } &\textbf{0.71} & \textbf{2.34 } &\textbf{0.75} & \textbf{0.78 }  \\
    \hline
  \end{tabular}
  \end{adjustbox}
  \caption{Comparison between ML-Net, SalGAN, cGAN and  MC-GAN (proposed) on MIT-PD test set}
			\vspace{-3mm}
  \label{tab:tab_4}
\end{table}

\begin{figure}[!htb]
   \centering
    \begin{subfigure}{.24\linewidth}
       \centering
        \includegraphics[width=.95\linewidth]{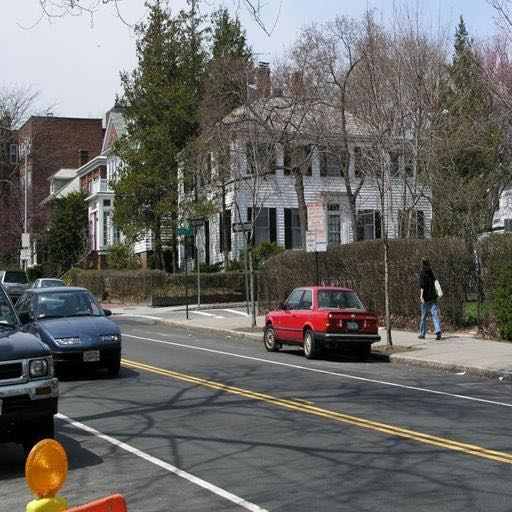} %
        
    \end{subfigure}
     \begin{subfigure}{.24\linewidth}
       \centering
        \includegraphics[width=.95\linewidth]{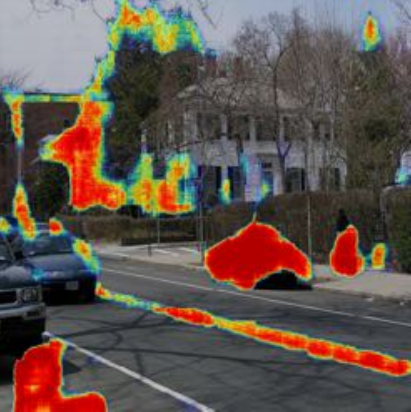} %
           
    \end{subfigure}
 \centering
    \begin{subfigure}{.24\linewidth}
       \centering
        \includegraphics[width=.95\linewidth]{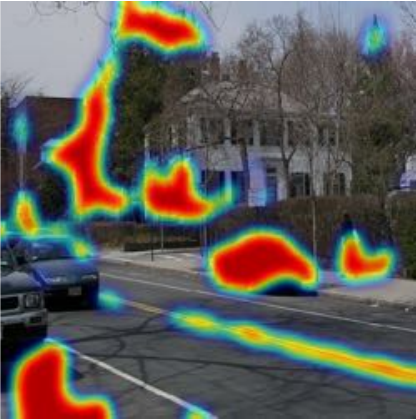} %
        
    \end{subfigure}
 \centering
    \begin{subfigure}{.24\linewidth}
       \centering
        \includegraphics[width=.95\linewidth]{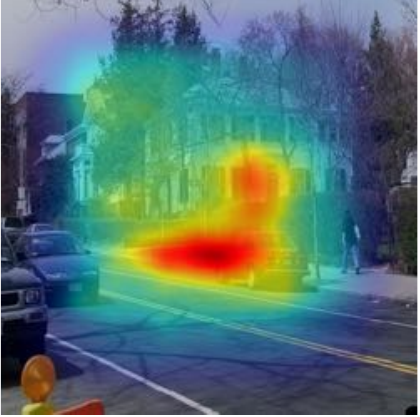} %
        
    \end{subfigure}
   
   \centering
    \begin{subfigure}{.24\linewidth}
       \centering
        \includegraphics[width=.95\linewidth]{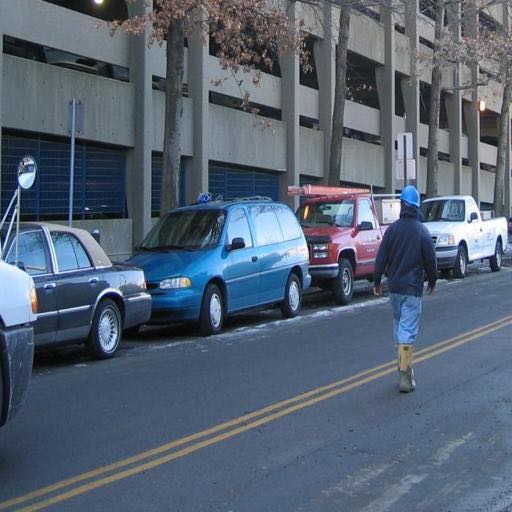} %
        
    \end{subfigure}
      \centering
    \begin{subfigure}{.24\linewidth}
       \centering
        \includegraphics[width=.95\linewidth]{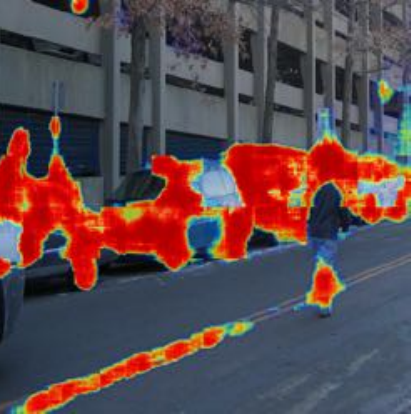} %
    
    \end{subfigure}
 \centering
    \begin{subfigure}{.24\linewidth}
       \centering
        \includegraphics[width=.95\linewidth]{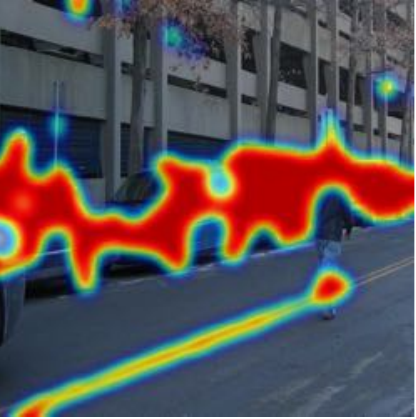} %
        
    \end{subfigure}
 \centering
    \begin{subfigure}{.24\linewidth}
       \centering
        \includegraphics[width=.95\linewidth]{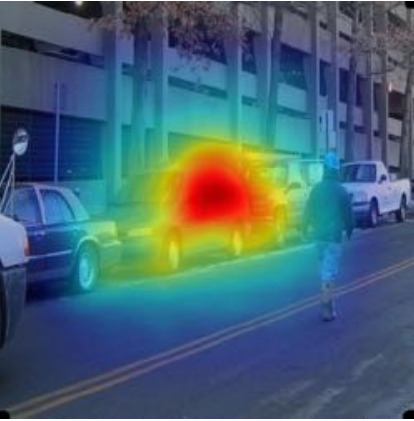} %
        
    \end{subfigure}
%
%
%
%
 
\centering
    \begin{subfigure}{.24\linewidth}
       \centering
        \includegraphics[width=.95\linewidth]{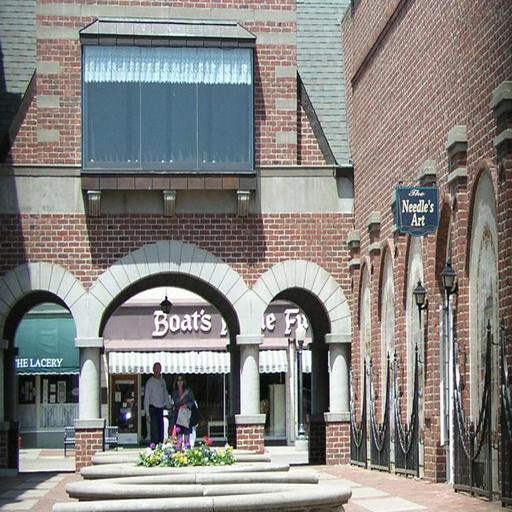} %
        
    \end{subfigure}
     \centering
    \begin{subfigure}{.24\linewidth}
       \centering
        \includegraphics[width=.95\linewidth]{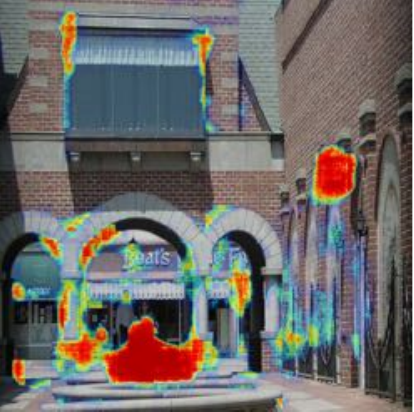} %
    
    \end{subfigure}
 \centering
    \begin{subfigure}{.24\linewidth}
       \centering
        \includegraphics[width=.95\linewidth]{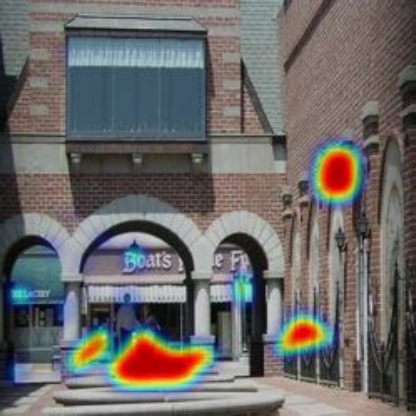} %
        
    \end{subfigure}
 \centering
    \begin{subfigure}{.24\linewidth}
       \centering
        \includegraphics[width=.95\linewidth]{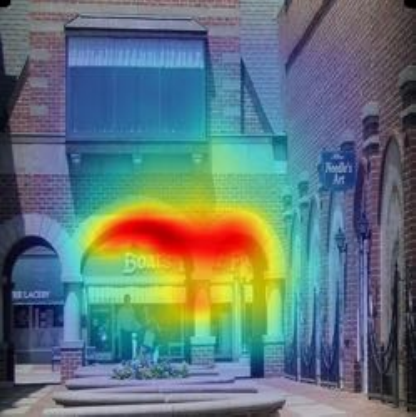} %
        
    \end{subfigure}
	\centering
    \begin{subfigure}{.24\linewidth}
       \centering
        \includegraphics[width=.95\linewidth]{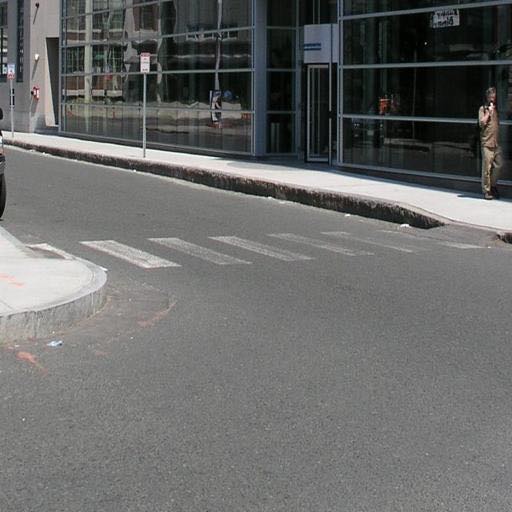} %
        
    \end{subfigure}
     \centering
    \begin{subfigure}{.24\linewidth}
       \centering
        \includegraphics[width=.95\linewidth]{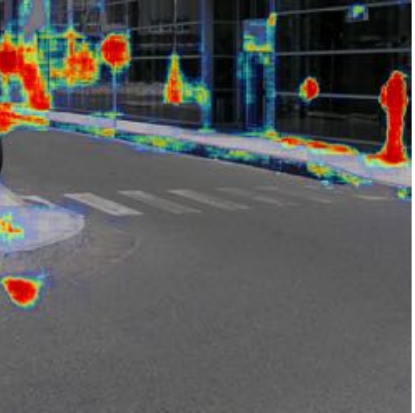} %
    
    \end{subfigure}
 \centering
    \begin{subfigure}{.24\linewidth}
       \centering
        \includegraphics[width=.95\linewidth]{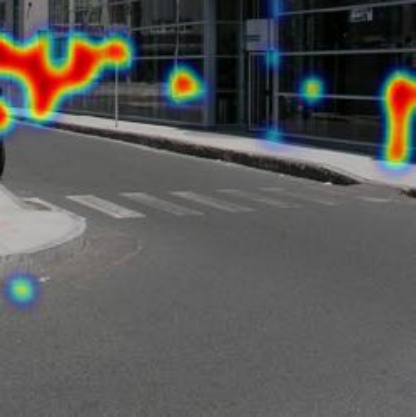} %
        
    \end{subfigure}
 \centering
    \begin{subfigure}{.24\linewidth}
       \centering
        \includegraphics[width=.95\linewidth]{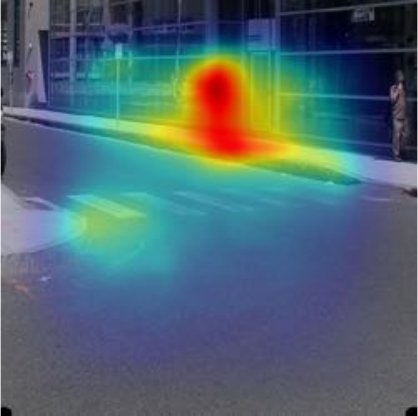} %
        
    \end{subfigure}    
     
   \centering
    \begin{subfigure}{.24\linewidth}
       \centering
        \includegraphics[width=.95\linewidth]{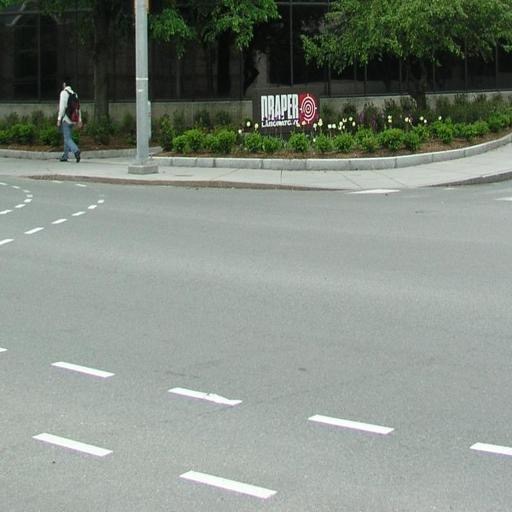} %
        
    \end{subfigure}
     \centering
    \begin{subfigure}{.24\linewidth}
       \centering
        \includegraphics[width=.95\linewidth]{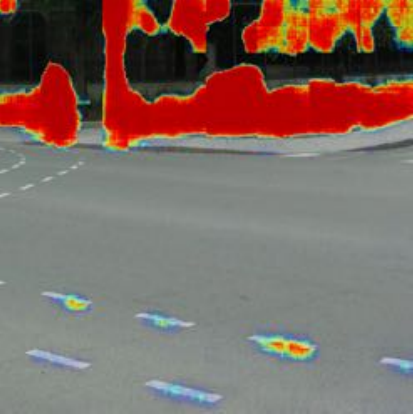} %
    
    \end{subfigure}
 \centering
    \begin{subfigure}{.24\linewidth}
       \centering
        \includegraphics[width=.95\linewidth]{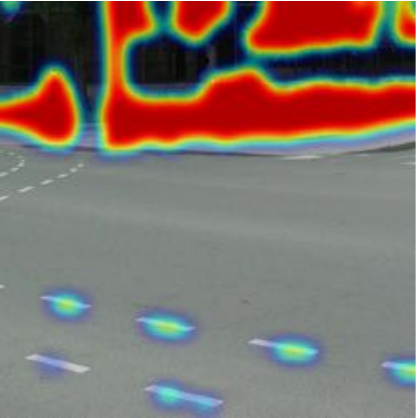} %
        
    \end{subfigure}
 \centering
    \begin{subfigure}{.24\linewidth}
       \centering
        \includegraphics[width=.95\linewidth]{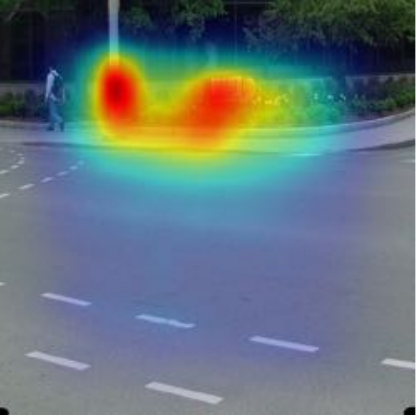} %
        
    \end{subfigure}
 
%
%
%
        \centering
    \begin{subfigure}{.24\linewidth}
       \centering
        \includegraphics[width=.95\linewidth]{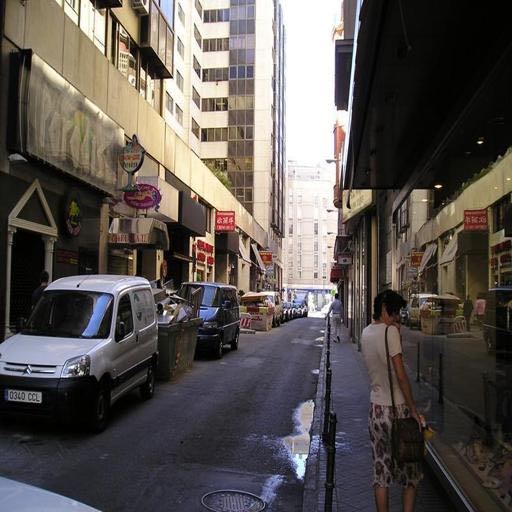} %
        \caption{Image}
    \end{subfigure}
     \centering
    \begin{subfigure}{.24\linewidth}
       \centering
        \includegraphics[width=.95\linewidth]{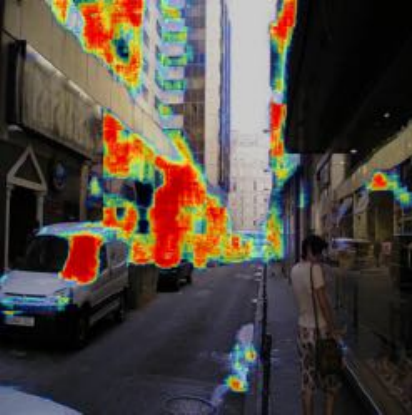} %
        \caption{GT}
    \end{subfigure}
 \centering
    \begin{subfigure}{.24\linewidth}
       \centering
        \includegraphics[width=.95\linewidth]{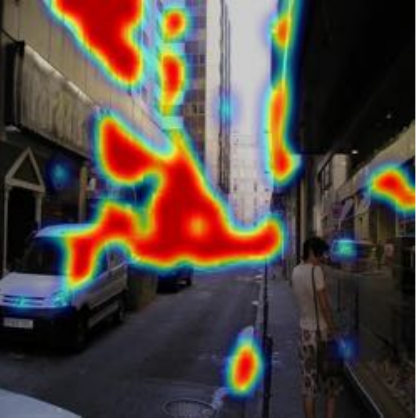} %
        \caption{Our}
    \end{subfigure}
 \centering
    \begin{subfigure}{.24\linewidth}
       \centering
        \includegraphics[width=.95\linewidth]{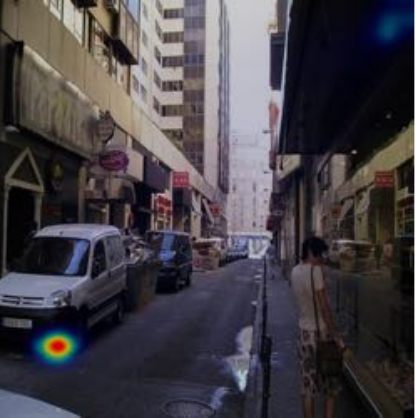} %
        \caption{ML-net}
    \end{subfigure}

     \caption{Qualitative results for MIT-PD dataset and comparisons to the state-of-the-art.}
        \label{fig:fig_context_rec}
\end{figure}

The qualitative results obtained from the proposed model along with the ML-net \cite{deep-ml} network on a few examples from the MIT-PD dataset are shown in Fig. \ref{fig:fig_context_rec}. We would like to emphasise the usage of a subject's prior knowledge in the task of searching for people in the urban scene. The subjects selectively attend the areas such as high rise buildings (see rows 2, 6) and pedestrian walkways (see rows 1, 3-6), which are more likely to contain humans, which our model has effectively captured. With the lack of capacity to model such user preferences, the baseline ML-Net model generates centre biased saliency without effectively understanding the subject's strategy. 

\subsection{Task Specific Generator Activations}
\begin{figure}[htb]
   \centering
    \begin{subfigure}{.190\linewidth}
       \centering
        \includegraphics[width=.95\linewidth, ,height= .82\linewidth]{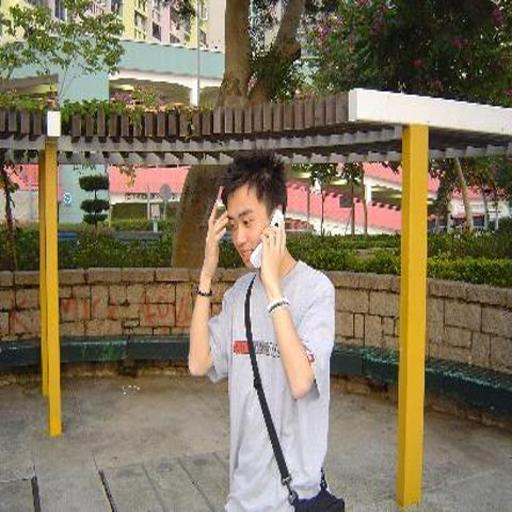} %
         \caption{Input}
          \end{subfigure}
     \begin{subfigure}{.190\linewidth}
       \centering
        \includegraphics[width=.95\linewidth]{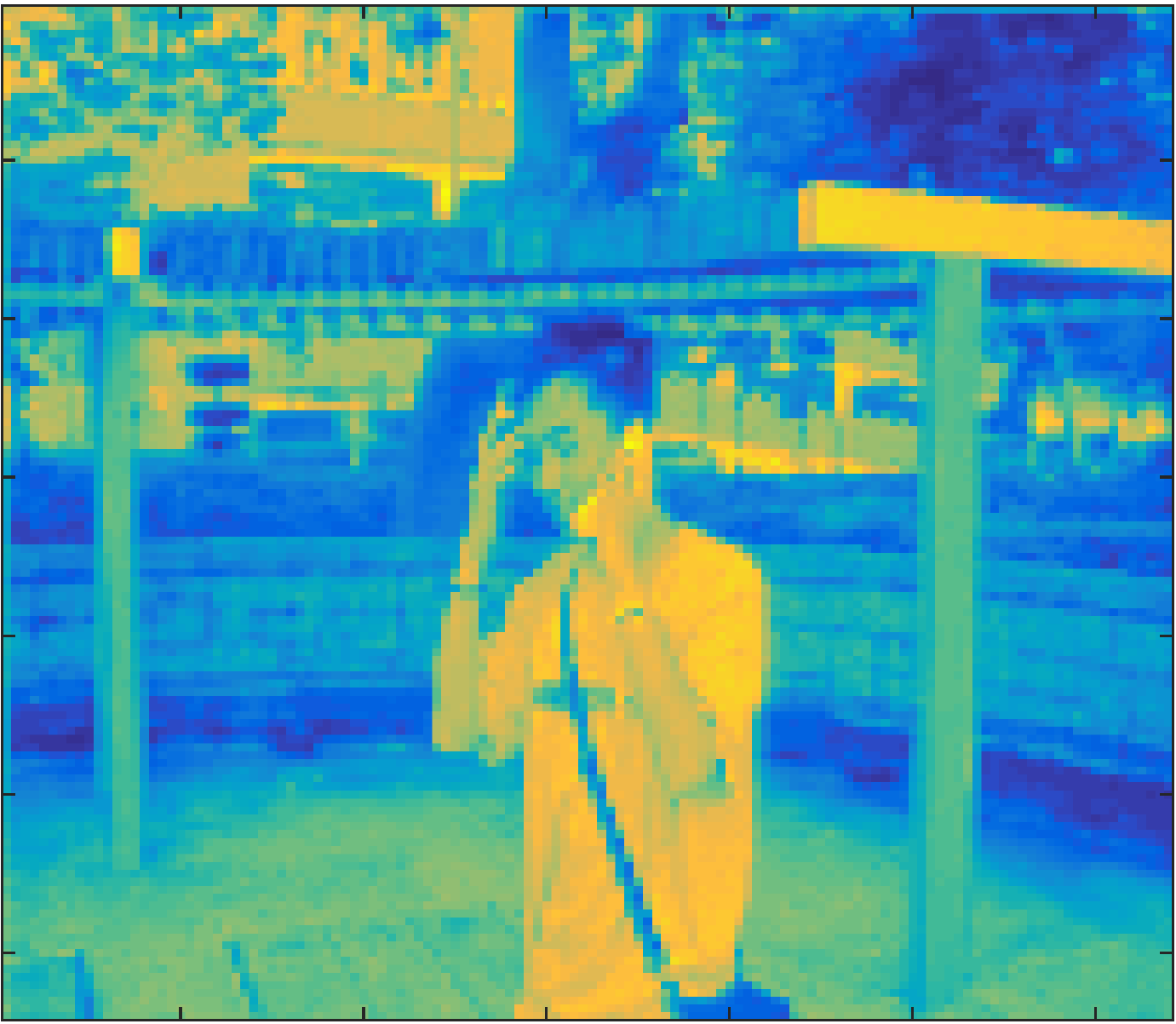} %
        \caption{l-2 AR}  
         \end{subfigure}
 \centering
    \begin{subfigure}{.190\linewidth}
       \centering
        \includegraphics[width=.95\linewidth]{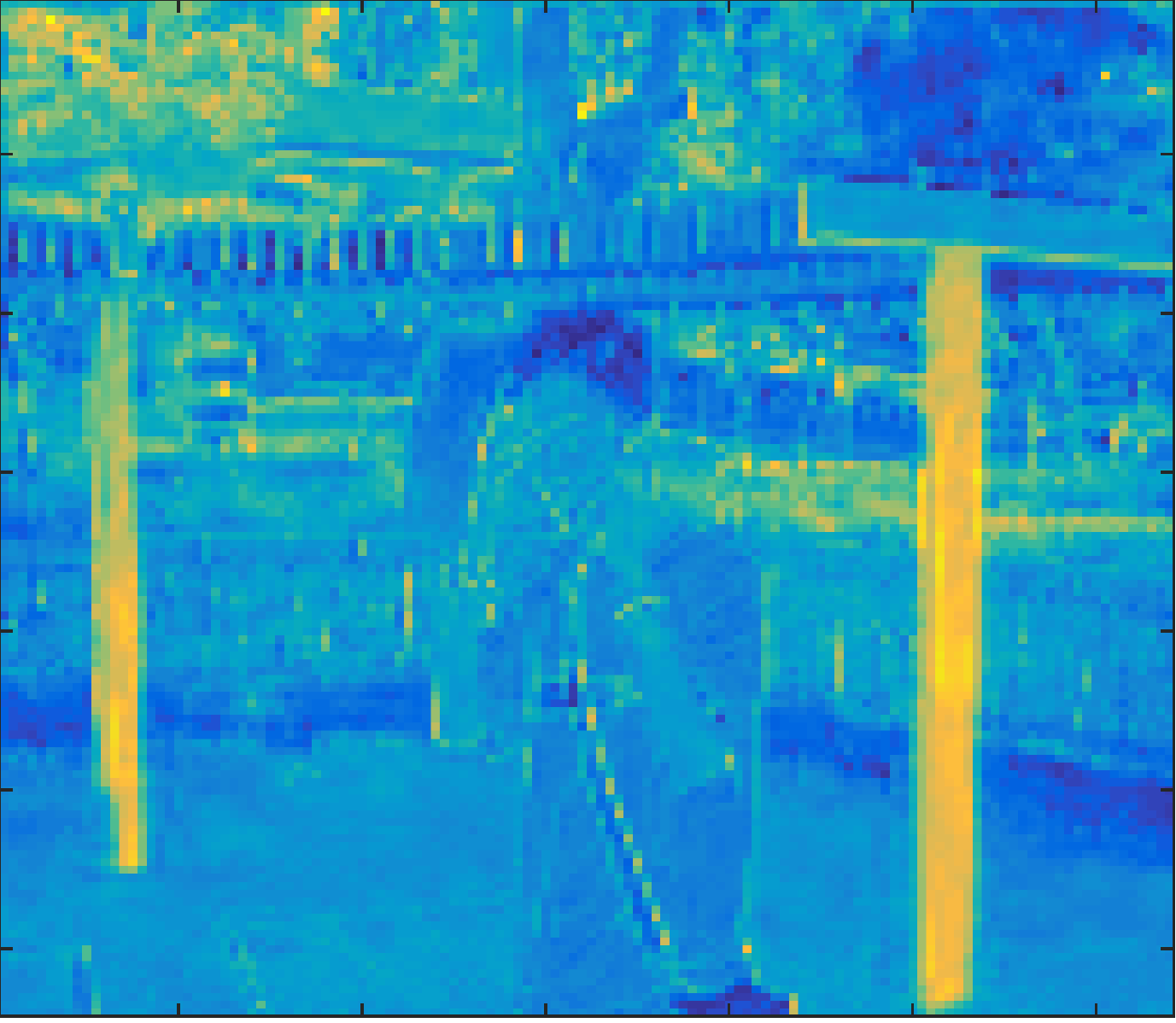} %
         \caption{l-2 CR}
    \end{subfigure}
 \centering
    \begin{subfigure}{.190\linewidth}
       \centering
        \includegraphics[width=.95\linewidth ]{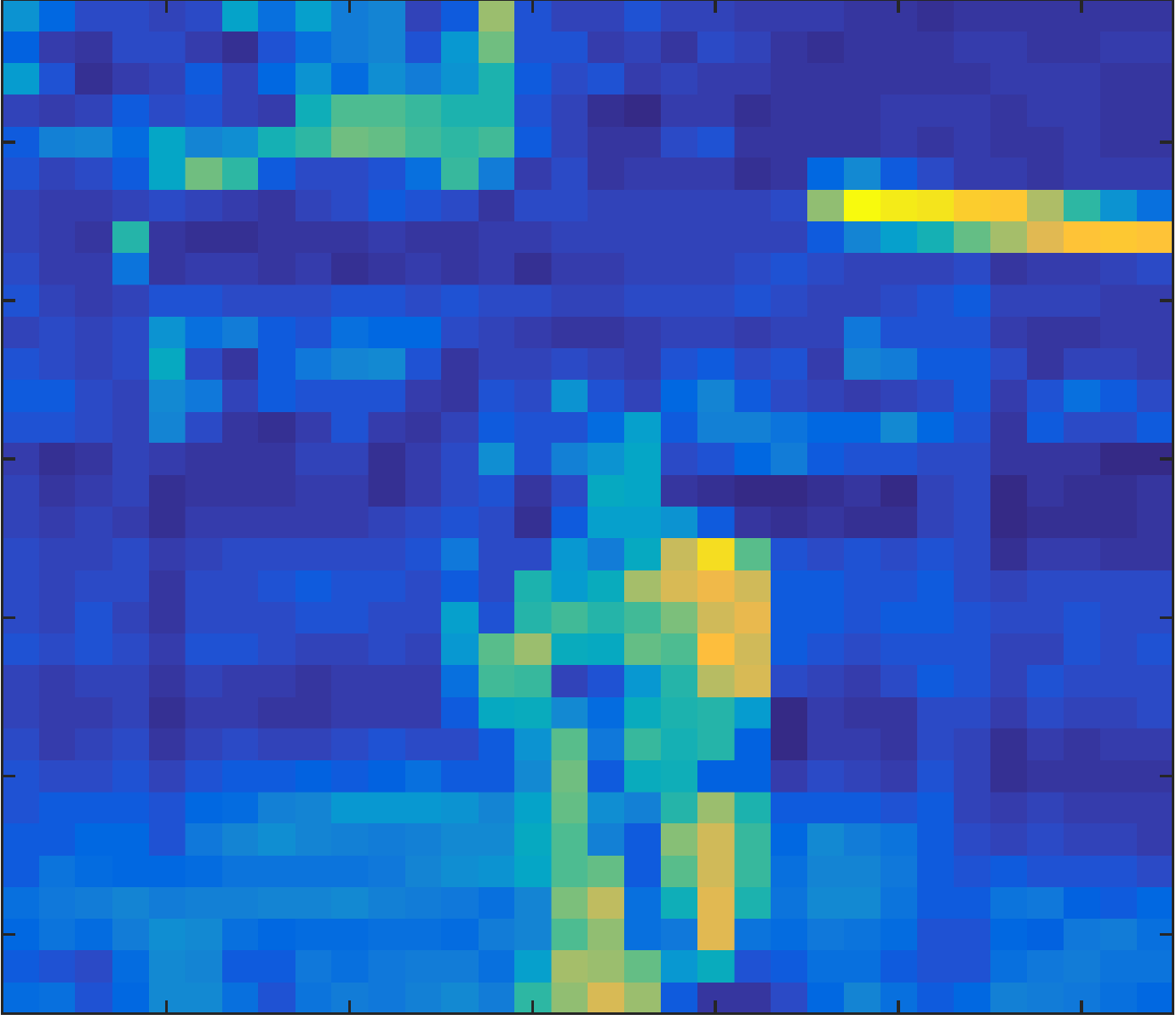} %
        \caption{l-8 AR}
        \end{subfigure}
    \centering
    \begin{subfigure}{.190\linewidth}
       \centering
        \includegraphics[width=.95\linewidth]{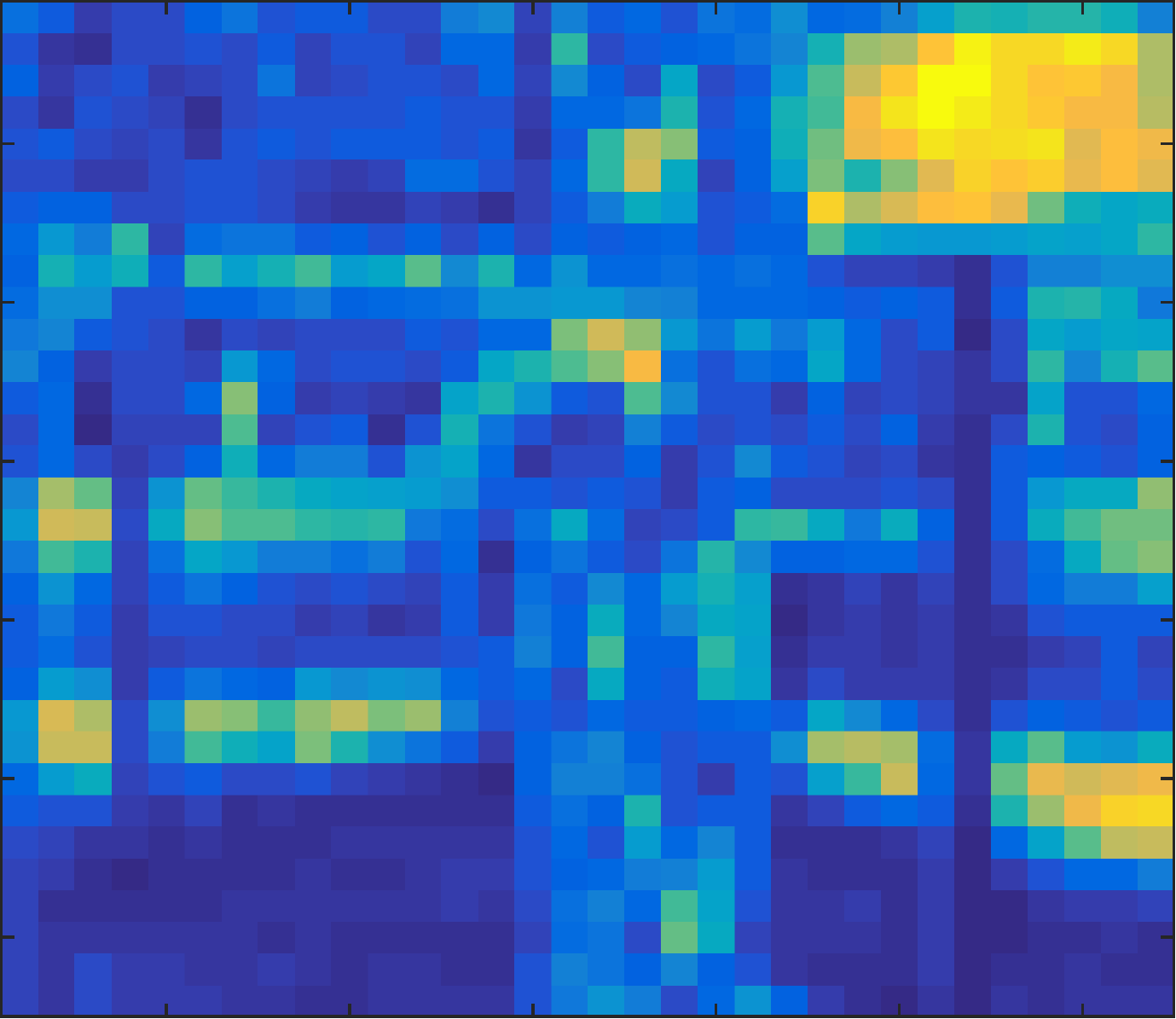} %
         \caption{l-8 CR}
          \end{subfigure}
    
      \centering
    \begin{subfigure}{.190\linewidth}
       \centering
        \includegraphics[width=.95\linewidth, ,height= .82\linewidth]{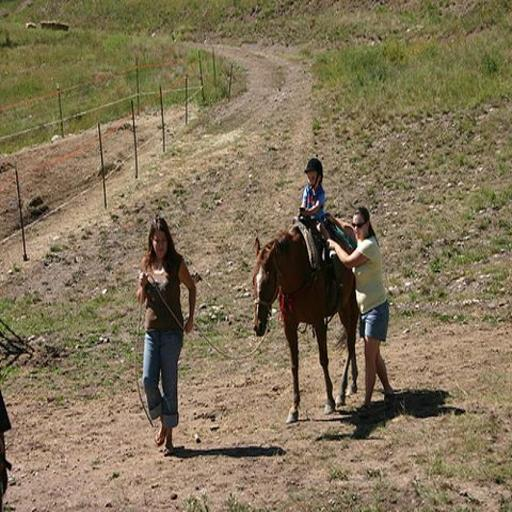} %
        \caption{Input}
    \end{subfigure}
     \begin{subfigure}{.190\linewidth}
       \centering
        \includegraphics[width=.95\linewidth]{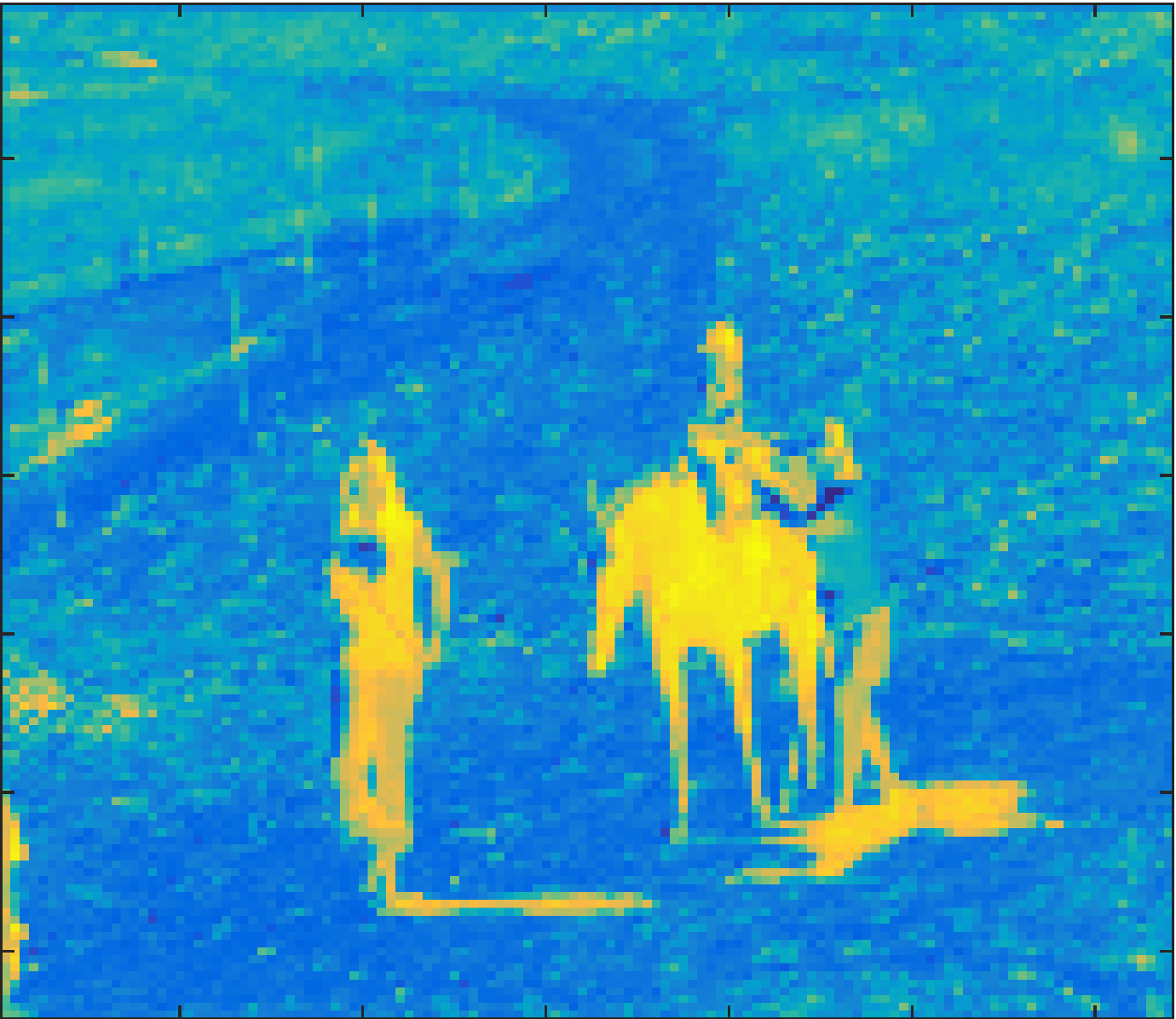} %
         \caption{l-2 AR}  
    \end{subfigure}
 \centering
    \begin{subfigure}{.190\linewidth}
       \centering
        \includegraphics[width=.95\linewidth, ,height= .82\linewidth]{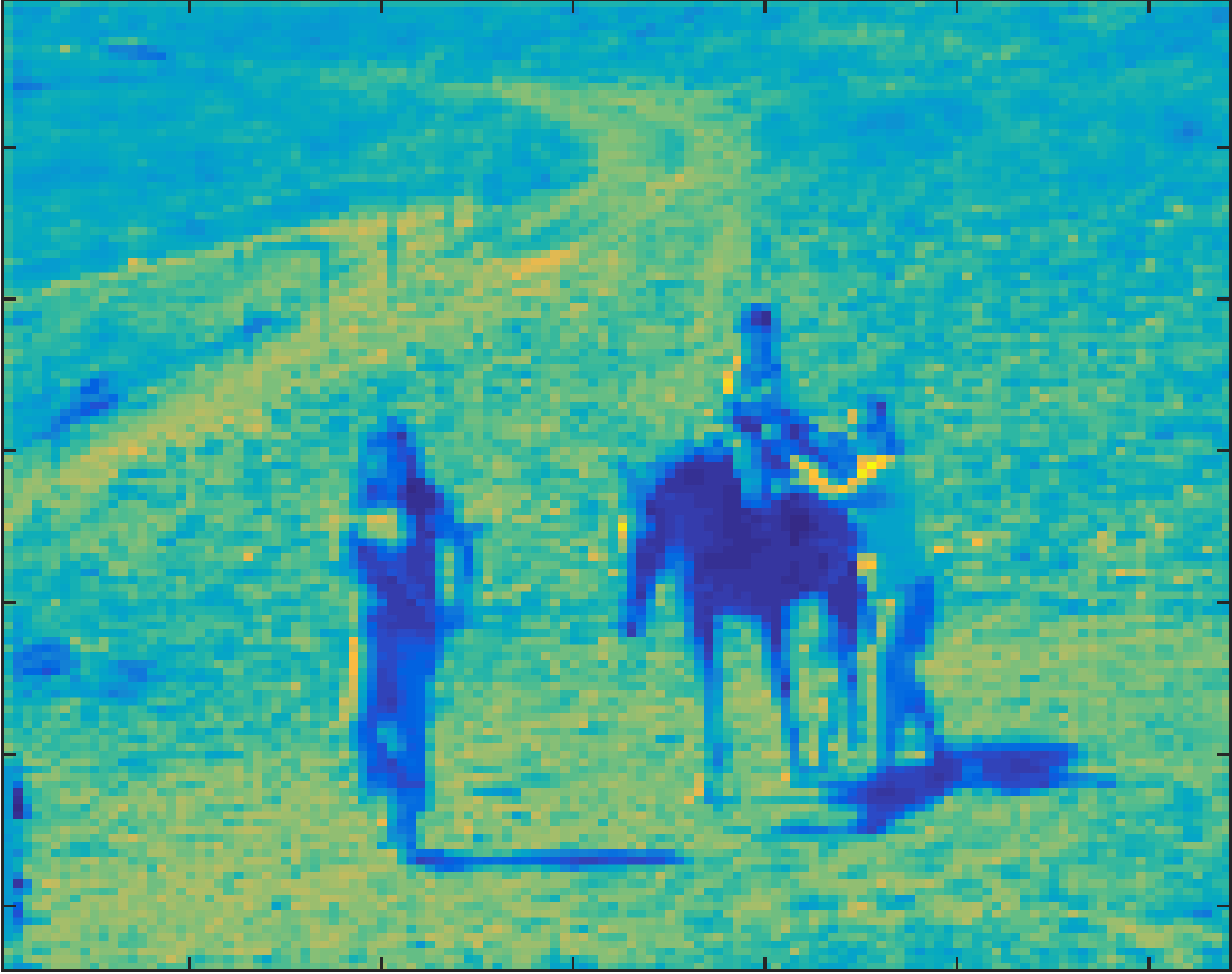} %
         \caption{l-2 CR}
    \end{subfigure}
 \centering
    \begin{subfigure}{.190\linewidth}
       \centering
        \includegraphics[width=.95\linewidth, ,height= .82\linewidth]{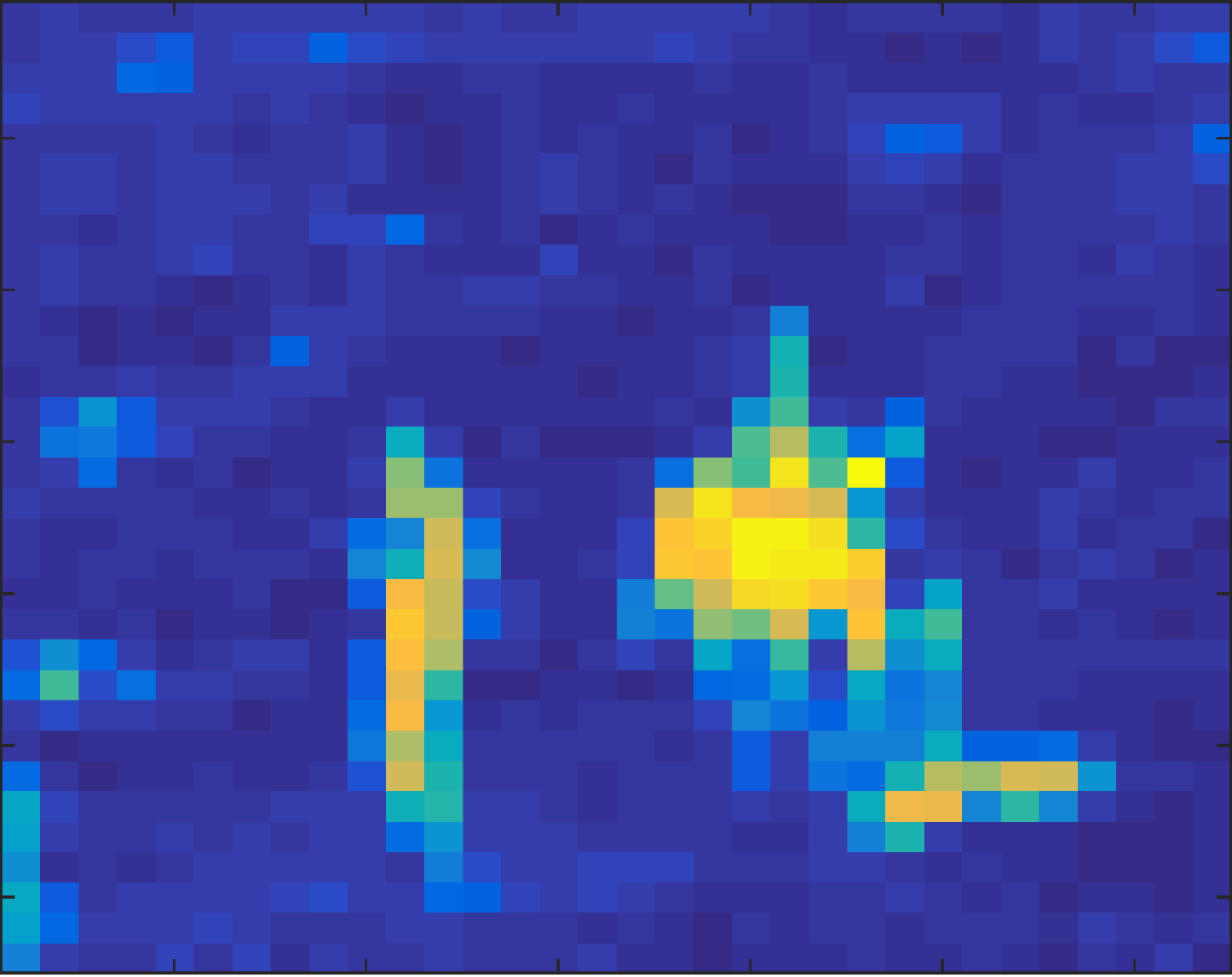} %
        \caption{l-8 AR}
    \end{subfigure}
    \centering
    \begin{subfigure}{.190\linewidth}
       \centering
        \includegraphics[width=.95\linewidth, ,height= .82\linewidth]{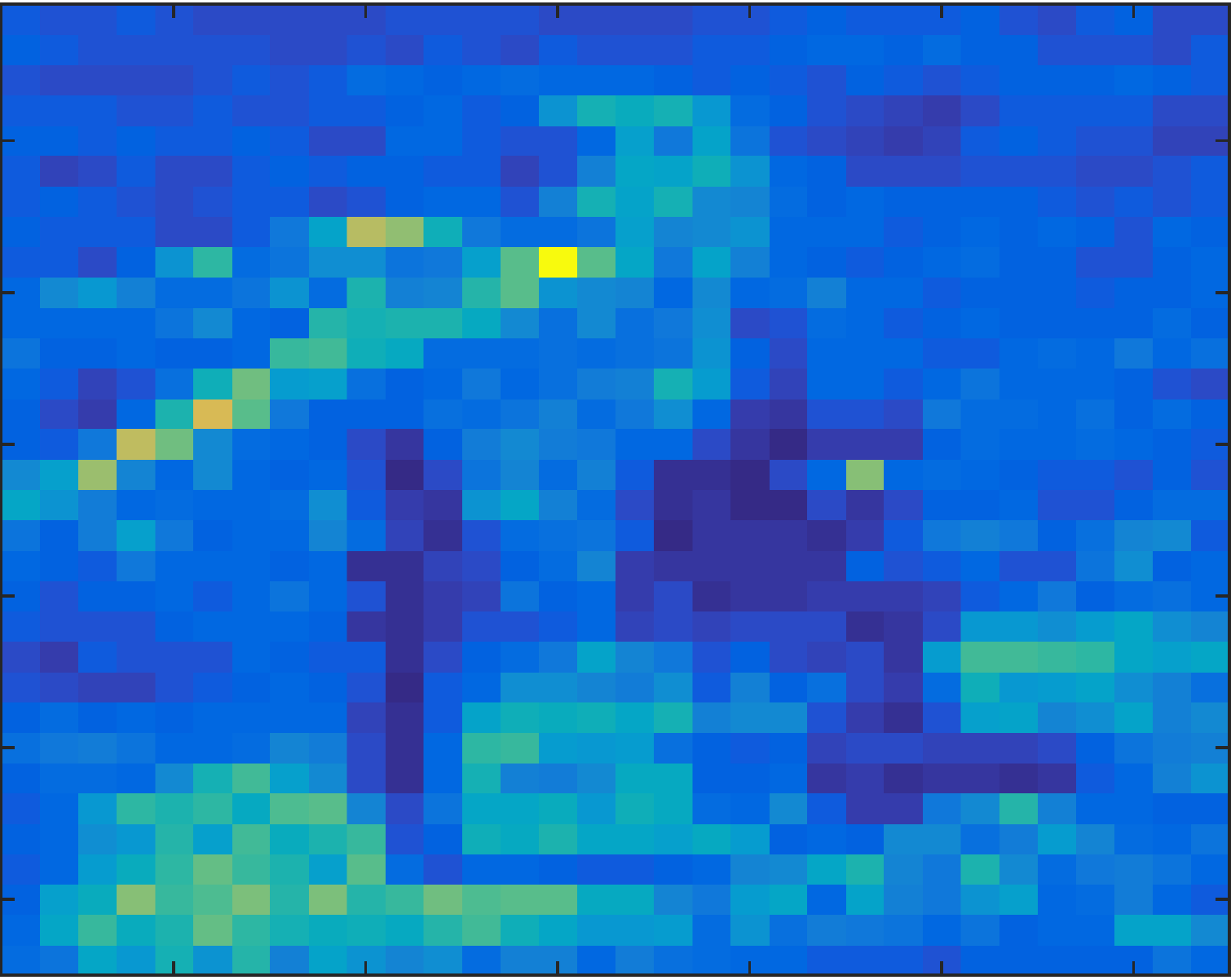} %
        \caption{l-8 CR}
    \end{subfigure}
     \caption{Visualisation of the generator activations from 2nd (l-2) and 2nd last (l-8) convolution layers for action recognition (AR) and context recognition (CR) tasks. The importance varies from blue to yellow where blue represents the areas of least importance and yellow represents areas of more importance.}
			\vspace{-3mm}
\label{fig:generator_activations}
    \end{figure}

In Fig. \ref{fig:generator_activations} we visualise the activations from the 2nd (conv-l-2) and 2nd last (conv-l-8) convolution layers of the generator. The task specific learning of the proposed conditional GAN architecture is clearly evident in the activations. For instance, when the task at hand is to recognise actions the generator activations are highly concentrated around the foreground of the image (see (b), (g)), while for context recognition the model has learned that the areas of interest are in the background of the image (see (c), (h)). These task specific salient features are combined and compressed hierarchically and in latter layers (i.e conv-l-8), the networks has learned the most specific areas to focus when generating the output saliency map.

\vspace{-2mm}
\section{Conclusion}

This work introduces a novel human saliency estimation architecture which combines task and user specific information together in a generative adversarial pipeline. We show the importance of fully capturing the context information which incorporates the task information, subject behavioural goals and image context. The resultant frame work offers several advantages compared to task specific handcrafted features, enabling direct transferability among different tasks. Qualitative and quantitative experimental evaluations on two public datasets demonstrates superior performance with respect to the current state-of-the-art. 

\vspace{-2mm}
\footnotesize{
\subsubsection*{Acknowledgement}
\vspace{-2mm}
This research was supported by an Australian Research Council's Linkage grant (LP140100221). The authors also thank QUT High Performance Computing (HPC) for providing the computational resources for this research.
}

\bibliographystyle{IEEEtran}
\bibliography{my_ref}

\end{document}